\documentclass[10pt,twocolumn,letterpaper]{article}

\usepackage[pagenumbers]{cvpr} % To force page numbers, e.g. for an arXiv version
\usepackage{graphicx}
\usepackage{amsmath}
\usepackage{amssymb}
\usepackage{booktabs}
\definecolor{cvprblue}{rgb}{0.21,0.49,0.74}
\usepackage[pagebackref,breaklinks,colorlinks,allcolors=cvprblue]{hyperref}
\usepackage[dvipsnames]{xcolor}
\usepackage{soul}
\usepackage{url}
\usepackage[utf8]{inputenc}
\usepackage{graphicx}
\usepackage{amsmath}
\usepackage{booktabs}
\usepackage{epsfig}
\usepackage{amssymb}
\usepackage{hhline}
\usepackage{multirow}
\usepackage{array}
\usepackage{wrapfig}
\usepackage{tabu}
\usepackage{floatrow, comment, enumerate}
\usepackage{color}
\usepackage{algorithmic}
\usepackage[linesnumbered,ruled,vlined]{algorithm2e}
\usepackage{subcaption}
\usepackage[utf8]{inputenc}
\usepackage{paralist}

\DeclareUnicodeCharacter{2212}{-}
\newcommand{\best}[1]{{\bf{#1}}}
\newcommand{\second}[1]{\underline{#1}}
\let\oldnl\nl% Store \nl in \oldnl
\newcommand{\nonl}{\renewcommand{\nl}{\let\nl\oldnl}}% Remove line number for one line (algorithm2e)

\newcommand{\red}[1]{{\textbf{\color{red}#1}}}
\newcommand{\blue}[1]{{\textbf{\color{blue}#1}}}

\newcommand{\HUGE}{\@setfontsize\Huge{50}{60}}
\def\paperID{9720} % *** Enter the Paper ID here
\def\confName{CVPR}
\def\confYear{2025}

\title{Delve into Visual Contrastive Decoding for Hallucination Mitigation of \\ Large Vision-Language Models}

\author{%
  Yi-Lun Lee\textsuperscript{$\dagger$} \quad Yi-Hsuan Tsai\textsuperscript{$\ddagger$} \quad Wei-Chen Chiu\textsuperscript{$\dagger$}  \\
  \textsuperscript{$\dagger$}National Yang Ming
Chiao Tung University \quad \textsuperscript{$\ddagger$}Atmanity\\  
  \small{\texttt{\{yllee10727, walon\}@cs.nctu.edu.tw}}, \;  
  \small{\texttt{yhtsai@atmanity.io}}
}

\begin{document}
\maketitle
\begin{abstract}
While large vision-language models (LVLMs) have shown impressive capabilities in generating plausible responses correlated with input visual contents, they still suffer from hallucinations, where the generated text inaccurately reflects visual contents.
To address this, recent approaches apply contrastive decoding to calibrate the model's response via contrasting output distributions with original and visually distorted samples, demonstrating promising hallucination mitigation in a training-free manner.
However, the potential of changing information in visual inputs is not well-explored, so a deeper investigation into the behaviors of visual contrastive decoding is of great interest.
In this paper, we first explore various methods for contrastive decoding to change visual contents, including image downsampling and editing. Downsampling images reduces the detailed textual information while editing yields new contents in images, providing new aspects as visual contrastive samples.
% with queried or hallucinated objects yields new images with additional contents, both of which provide new aspects to serve as contrastive visual samples.
%
To further study benefits by using different contrastive samples, we analyze probability-level metrics, including entropy and distribution distance.
Interestingly, the effect of these samples in mitigating hallucinations varies a lot across LVLMs and benchmarks.
Based on our analysis, we propose a simple yet effective method to combine contrastive samples, offering a practical solution for applying contrastive decoding across various scenarios.
Extensive experiments are conducted to validate the proposed fusion method among different benchmarks.
Code is available at \href{https://github.com/YiLunLee/VCD_Analysis}{https://github.com/YiLunLee/VCD\_Analysis}.
\end{abstract}    
\vspace{-2mm}
\section{Introduction}
\label{sec:intro}
\begin{figure}[!t]
    \centering
    \includegraphics[width=0.9\linewidth]{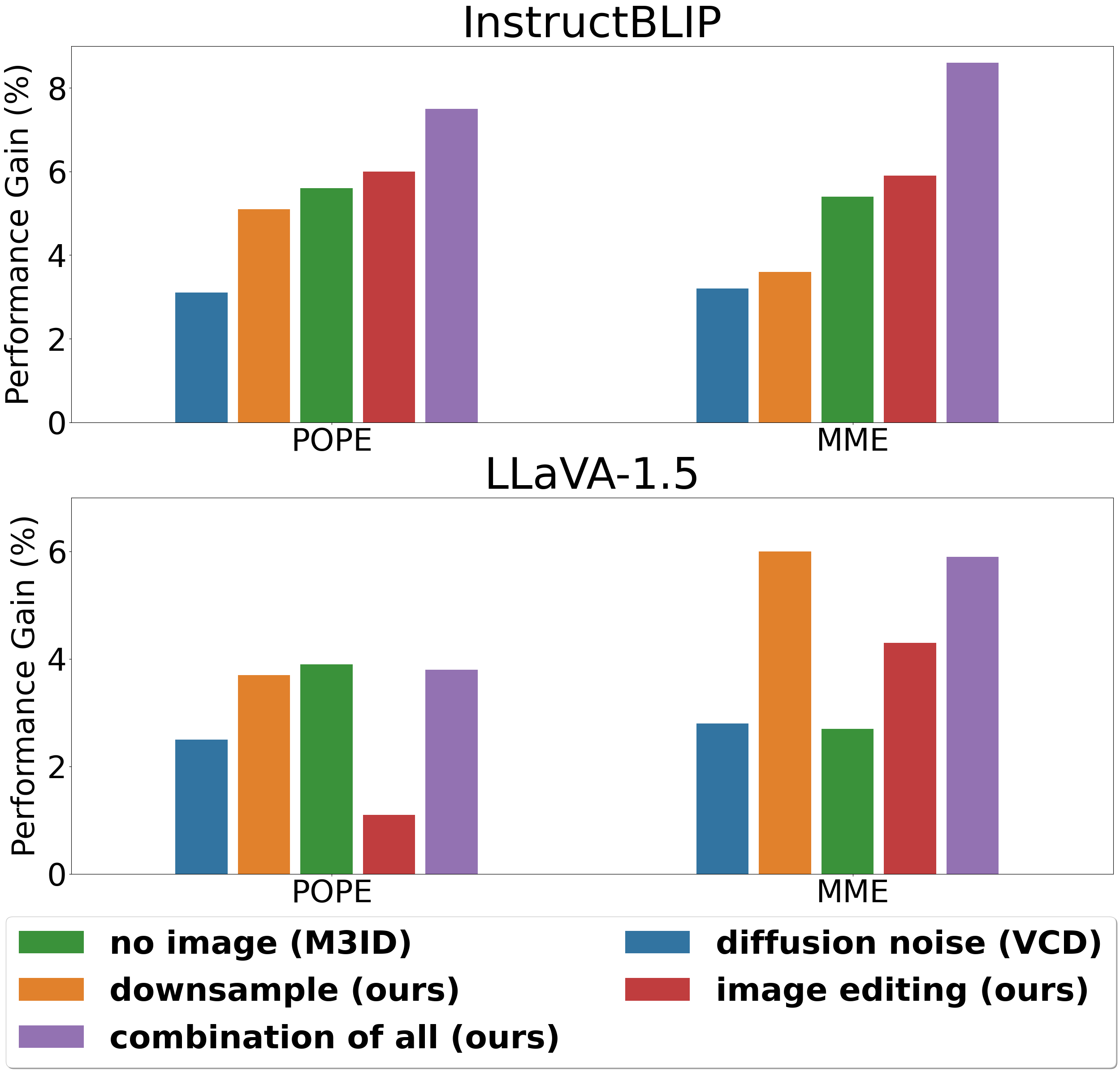}
    \caption{Performance gain brought by visually changed samples across different LVLMs and benchmarks (POPE~\cite{li2023pope} and MME~\cite{fu2023mme}). We find that contrastive decoding with different samples reaches varying performance gains across benchmarks and LVLMs, motivating us to develop a solution to leverage the advantage of each CD method in a combined manner.
    }
    \label{fig:teasor}
    \vspace{-2mm}
\end{figure}

% introduce LVLMs superiority and hallucination
Recent large vision-language models (LVLMs)~\cite{instructblip,llava15,bai2023qwen} have shown strong capabilities of yielding detailed and plausible responses correlated to the visual inputs. However, stemming from LLMs, LVLMs also suffer from ``hallucination'', where the generated responses are fluent and semantically coherent but may conflict with input visual contents, leading to a risk of producing fabricated information.
To mitigate hallucinations in LVLMs, various approaches have been explored, including learning from human feedback or preference~\cite{sun2023rlhf,zhao2023hadpo,ouali2024clipdpo}, fine-tuning LVLMs with newly proposed hallucination-calibration datasets~\cite{gunjal2024detecting, liu2024lrvinstruct}, training a post-hoc calibrator to rectify hallucinated responses~\cite{zhou2023lure,whitehead2024pre}, and designing novel decoding strategies~\cite{leng2024vcd,favero2024m3id,wang2024icd}.
% - Learning from human feedback or preference (RLHF, DPO)
% - finetune with newly proposed hallucination-calibration datasets (Instruct tuning)
% - training an additional hallucination revised (Postdoc revisor)
% - decoding strategies (VCD)
However, most of them require collecting new datasets and fine-tuning LVLMs, which can be a huge burden with manual efforts and computational resources, as well as less flexible to deploy in specific domains.

% introduce the Contrastive decoding and its problems
In contrast, training-free decoding strategies such as contrastive decoding (CD)~\cite{li2023cd,leng2024vcd,favero2024m3id,chen2024halc,wang2024icd}, which adjusts predicted logits via subtracting the logits from contrastive samples (e.g., distorted visual inputs), serves as a simple yet effective approach to mitigate hallucinations in LVLMs without additional datasets and fine-tuning.
The insight is that the visual input with distortion exacerbates hallucinations happening in the response, and thus, CD subtracts the hallucinated contents from the original distribution to mitigate hallucinations.
% Hallucination in LVLMs is caused by insufficient captured visual information and the dominant pre-trained LLM bias, encouraging the LVLMs to concentrate on the text contents without acquiring visual information.
For example, VCD~\cite{leng2024vcd} adds Gaussian noise to the input image in a diffusion manner, which increases the uncertainty of LVLMs' responses and thus strengthens the probability of producing hallucinations as contrastive samples. M3ID~\cite{favero2024m3id} directly removes the visual input, forcing the LVLMs to contrast with the bias from pre-trained LLMs. ICD~\cite{wang2024icd} introduces deceptive instruction for the visual-text interface (e.g., Q-Former~\cite{instructblip}), exacerbating the misalignment between visual and text embedding.
% Despite the remarkable performance,  
% However, the dimensions of visually changed samples are not well explored; there are other ways to reduce information, and in contrast, adding content to images is also not discussed.
% VCD -> increase uncertainty
% M3ID -> contrast with LLM bias
% ICD -> destroy the visual and textual alignment 
% downsample -> reduce the high frequency of image contents
% editing -> insert additional content into images

% explore visually changed samples
Inspired by the success of contrastive decoding with visually changed samples, we further explore the other two ways of distorting visual inputs to serve as contrastive samples, including image downsampling and editing. The former reduces high-frequency details from images, while the latter inserts specific hallucination contents into the input image.
% Downsampling input images reduces the high-frequency details, increasing the probability of producing hallucinations. In contrast, image editing with hallucinating objects inserts the specific hallucination contents into the input images to encourage LVLMs to induce hallucination, serving a novel perspective for contrastive decoding.
We hypothesize that different visually changed samples bring various outcomes of mitigating hallucinations. To verify this, we provide the performance gain of different CD methods on various benchmarks and LVLMs in Figure~\ref{fig:teasor}.
% why need more concrete analysis
We find that the efficacy of different CDs varies across LVLMs and tasks. For example, contrastive decoding with image editing improves the InstructBLIP~\cite{instructblip} on the POPE~\cite{li2023pope} and MME~\cite{fu2023mme} benchmarks the most but has a minor improvement for LLaVA-1.5~\cite{llava15} on the POPE benchmark.
The observation indicates that different samples exhibit significantly varied suitability across settings, which is practically challenging to determine which contrastive samples to use when applying to novel datasets and tasks.
This encourages us to delve into the impact of using various CDs and thus develop a simple solution to leverage the advantage of each CD method.

% point out the core insight of ours paper
% Given these visually changed samples, the . It is still not clear how this contrastive sampling influences the final response. To

% In this paper, we first introduce two other ways to change image contents as visual contrastive samples: 1) downsampling, that reduces the contextual details of images, and 2) image editing, that adds specific contents to images, which can be queried objects or hallucination contents.
To investigate the effect of each visually changed sample, we conduct thorough analysis from the perspective of the probability distribution of generated words, including entropy and predicted distribution distance. We further delve into models' behavior of various CD methods on different benchmarks with different pre-trained LVLMs. Interestingly, we observe that different CDs exhibit significantly varied suitability across LVLMs and datasets. In general, visually changed samples with higher entropy of next-token prediction act as more effective contrastive samples for mitigating hallucination.

Based on these findings, it can be a practical challenge to determine the suitable type of CDs to mitigate hallucinations when deploying the LVLMs in a new dataset or downstream task. To this end, we propose a simple yet effective fusion method to apply all different types of CDs according to their predictions' behavior. Specifically, we leverage entropy as guidance to automatically determine the scale of influence of contrastive decoding. 
Extensive experiments on different benchmarks verify the efficacy of the proposed fusion method in different situations, demonstrating that applying various visually changed samples with an entropy-weighted combination serves as a general solution to mitigate hallucinations in real-world scenarios across datasets and LVLMs.
Our main contributions are as follows:
\begin{compactitem}
    \item We explore various visual CDs to produce visually changed samples, such as image downsampling and editing, and investigate the impact of these samples for mitigating hallucinations in LVLMs.
    \item We conduct a thorough analysis of various CDs using probability-level metrics on LVLM's response, demonstrating that each contrastive sample has its distinct benefit across tasks and datasets.
    \item Based on our analysis, we further propose a simple yet effective fusion method to leverage different contrastive samples, serving as a more practical and general solution to mitigate hallucinations with various LVLMs on different tasks and datasets.
\end{compactitem}

\section{Related Work}
\subsection{Hallucination Mitigation in LVLMs}
With advancements in LLMs, various LVLMs incorporate visual encoders like CLIP~\cite{radford2021clip} with LLMs (e.g., LLaMA~\cite{touvron2023llama}, Vincuna~\cite{zheng2023vicuna}, or Qwen~\cite{qwen}) to yield reasonable and detailed responses related to visual contents, including but not limited to InstructBLIP~\cite{instructblip}, LLaVA-1.5~\cite{llava15}, Qwen-VL~\cite{bai2023qwen}, MiniGPT-4~\cite{zhu2023minigpt}, and MPLG-Owl2~\cite{ye2024mplug2}. 
Despite the promising performance, these LVLMs still suffer from hallucinations, where the generated responses are inaccurately grounded with image contents.
To address this, several approaches are proposed from different aspects: \textbf{learning with human feedback and preference} methods~\cite{sun2023rlhf,zhao2023hadpo,ouali2024clipdpo} either conduct reinforcement learning with the human feedback~\cite{sun2023rlhf} or leverage Direct Preference Optimization (DPO) framework~\cite{rafailov2024dpo} to fine-tune LVLMs with DPO datasets generated by GPT-4v~\cite{zhao2023hadpo} or CLIP Ranking~\cite{ouali2024clipdpo};
\textbf{finetuning} methods~\cite{liu2024lrvinstruct,gunjal2024detecting} first introduce newly collected hallucination-calibration datasets (e.g., LRV-Instruct~\cite{liu2024lrvinstruct} and M-HalDetect~\cite{gunjal2024detecting}) and finetune LVLMs with them to prevent hallucinations;
\textbf{post-hoc} methods~\cite{zhou2023lure,yin2023woodpecker} includes post-processing calibration with external models~\cite{yin2023woodpecker} or training post-hoc revisors to correct hallucinated responses~\cite{zhou2023lure}.

However, most of the above-mentioned approaches have practical drawbacks: they require significant manual effort to create hallucination-specific datasets and involve fine-tuning LVLMs or training post-hoc revisors.
In contrast, some studies focus on \textbf{adjusting decoding strategies}, offering a more practical solution without additional training or data. 
OPERA~\cite{huang2024opera} investigates the over-trust issue in LVLMs and penalizes tokens prone to hallucinations.
% CGD~\cite{deng2024seeing} addresses the sentence-level hallucinations by leveraging CLIP~\cite{radford2021clip} to estimate the semantic difference and select the sentence grounded with image contents the most.
CGD~\cite{deng2024seeing} leverages CLIP~\cite{radford2021clip} to select candidate sentences grounded with image contents.
As a mainstream of the decoding process, contrastive decoding (CD)~\cite{li2023cd,leng2024vcd,favero2024m3id,wang2024icd,chen2024halc,zhu2024ibd} mitigates hallucinations by adjusting logits based on contrastive samples.
%subtracts the original logits outputs by the logits of hallucination contents resulted from contrastive samples.
In this paper, we mainly focus on exploring contrastive decoding approaches.

\subsection{Contrastive Decoding}
Contrastive decoding~\cite{li2023cd} is oriented from mitigating the hallucination of LLMs, leveraging the contrast between expert and amateur models to amplify good expert behavior and diminish undesired amateur behavior.
To extend contrastive decoding for LVLMs, several approaches~\cite{leng2024vcd,favero2024m3id,wang2024icd,chen2024halc,zhu2024ibd} are proposed with different designs of contrastive samples prone to producing hallucinations. As a pioneer, VCD~\cite{leng2024vcd} increases the uncertainty of visual inputs by adding Gaussian noise in a diffusion manner, producing more hallucinated contents.
M3ID~\cite{favero2024m3id} directly contrasts the predictions from the original visual inputs with those of pre-trained LLMs without input images, which induce hallucinations from language priors.
IBD~\cite{zhu2024ibd} finetunes LVLM to be image-biased LVLM via amplification of attention on visual tokens and then contrast with conventional LVLM, which is likely text-biased to mitigate the hallucination.
ICD~\cite{wang2024icd} adjusts the instruction for the Q-Former~\cite{instructblip} with deceptive prompts to destroy the alignment between visual and text embeddings, leading to more likelihoods of producing hallucinations.
HALC~\cite{chen2024halc} selects different field of views (FOVs) based on decoding probabilities and performs contrastive decoding of FOV pairs to find the optimal outputs matching the visual input.

Despite the impressive performance, the potential of using visually changed samples in LVLMs remains underexplored. %we argue that the potential of leveraging visually changed/distorted samples has not been fully explored, both in breadth and depth. 
To this end, we first explore different manipulations of image contents for creating visually changed samples and investigate the prediction-level metrics on LVLMs' response and revision behaviors of contrastive decoding. Moreover, based on these observations, we propose an effective fusion method to leverage the strengths of different visually changed samples.

% \begin{itemize}
%     \item Hallucination for Multimodal LLMs

%     \item Contrastive Decoding
% \end{itemize}
\section{Visual Contrastive Decoding}
\begin{figure*}[!h]
    \centering
    \includegraphics[width=1\linewidth]{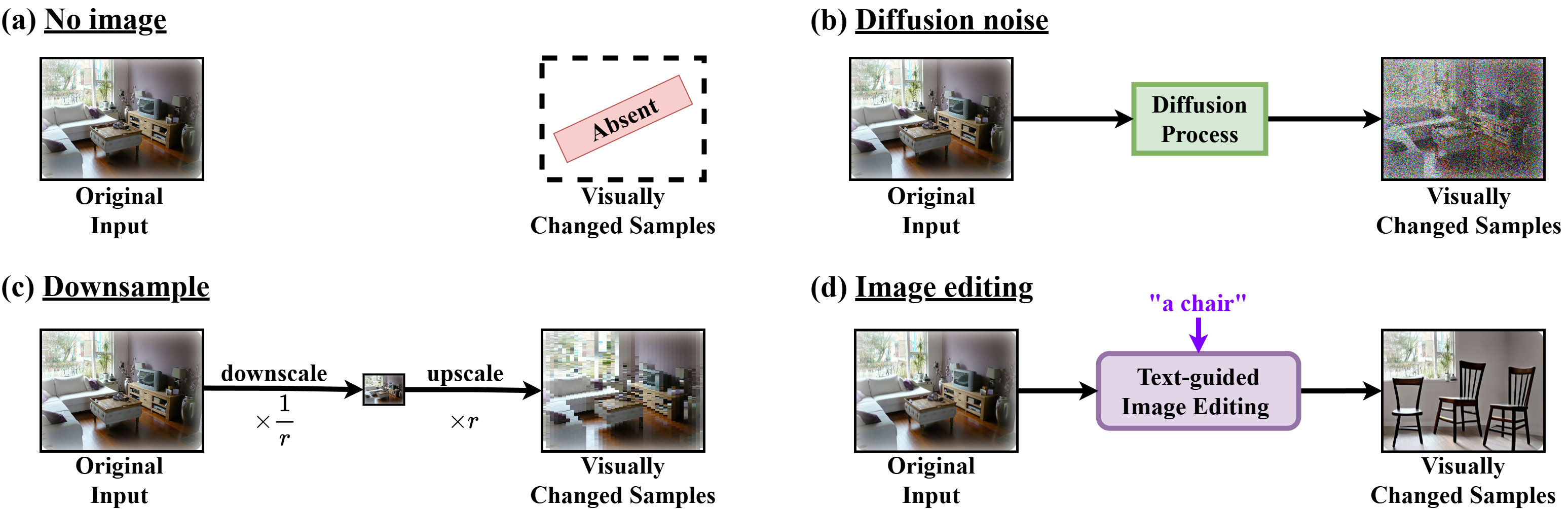}
    \caption{Illustration of various visually changed samples for different contrastive decoding strategies.}
    \label{fig:model}
    % \vspace{-3mm}
\end{figure*}
% \begin{figure*}[!t]
%     \centering
%     \LARGE
%     \resizebox{1.0\linewidth}!{
%         \begin{tabular}{ccc}
%         % \multirow{3}{*}{(a)} 
%         \begin{tabular}{c}\includegraphics[width=\textwidth]{figures/probdist/all_probdist_pope_a.png}\end{tabular}
%         &\begin{tabular}{c}\includegraphics[width=\textwidth]{figures/probdist/all_probdist_mme.png}\end{tabular}
%         &\begin{tabular}{c}\includegraphics[width=\textwidth]{figures/probdist/all_probdist_mme.png}\end{tabular}\\
        
%         (a) Diffusion noise (VCD~\cite{leng2024vcd}) & (b) Downsample (ours) & (c) image editing (ours)
%         \end{tabular}
%     }
%     \caption{The overview of different visually changed samples, including contrastive decoding pipeline and illustration of visually changed images. \textbf{[To be added if necessary]}
%     }
%     \label{fig:probdist}
% \end{figure*}

\subsection{Preliminaries}
\paragraph{Decoding process of LVLMs.}
We first introduce the generation paradigm of LVLMs, involving an LVLM parameterized by $\theta$, visual input $v$, and textual query $x$. Then LVLM considers the contextual information from the visual input $v$ to yield relevant responses $y$ to the textuary query $x$, where the responses are generated auto-regressively via sampling from the distribution conditioned on visual input $v$ and textual query $x$. The auto-regressive generation process can be formulated as follows:
\begin{equation}
\label{equ:decoding}
\begin{aligned}
y_t \sim p_\theta(y_t|v, x, y_{<t})=\text{softmax}(logit_\theta(y_t|v,x,y_{<t})).
\end{aligned}
\end{equation}
Here, $y_t$ denotes the token at time step $t$, and $y_{<t}$ represents the sequence of generated tokens up to the time step $(t-1)$. 
However, during the decoding process, the hallucination response may occur in several situations: 1) the visual contextual information is captured unsuccessfully, 2) LLM priors suppress the information from image embeddings, and 3) statistics from pre-trained datasets affect the responses with the similar scene learned previously. 

\vspace{-4mm}
\paragraph{Contrastive decoding with visually changed samples.}
To mitigate hallucinations, several approaches~\cite{li2023cd,leng2024vcd,favero2024m3id,wang2024icd} are proposed to adjust the decoding process in a contrastive manner. Specifically, samples that are changed with less visual information are introduced to produce a hallucination response due to insufficient visual contents. Then, these hallucinated contents are subtracted from the original response during the contrastive decoding process to prevent LVLMs from producing hallucinations.
That is, the original logits of the next token are substrated by those of contrastive samples, where the formula can be denoted as:
\begin{equation}
\label{equ:contrastive_decoding}
\begin{aligned}
p_{cd}(y_t|v, v_{cd}, x, y_{<t})&=\text{softmax}[(1+\alpha)logit_\theta(y_t|v,x,y_{<t})\\
&-\alpha(logit_\theta(y_t|v_{cd},x,y_{<t}))],
\end{aligned}
\end{equation}
where $v_{cd}$ is the visually changed sample and $\alpha$ is the weight to control the effect of differences between the two distributions.

\subsection{Visually Changed Samples in Different Aspects}
The high-level concept behind contrastive decoding is to create visually changed (or distorted) samples serving as contrastive samples, which are more likely to produce hallucinated contents due to the reduced image information. Previous works~\cite{leng2024vcd, favero2024m3id} have explored two ways, adding diffusion noise to the image or simply removing the image. Inspired by these methods, we further explore two distinct types: image downsampling and editing. The illustration of visually changed samples is shown in Figure~\ref{fig:model}.

\vspace{-4mm}
\paragraph{Diffusion noise.}
As a pioneer in applying contrastive decoding to multimodal scenarios, VCD~\cite{leng2024vcd} proposes applying a diffusion process following~\cite{ho2020denoising} to add Gaussian noise $\mathbf{N}$ onto the original input image $v_o$. During the diffusion process, the diffusion function $q$ incrementally adds a small amount of Gaussian noise for $N$ steps, producing distorted image sequences $\{v_1, v_2, ..., v_N\}$. The diffusion process can be formed:
\begin{equation}
\label{equ:diffusion_noise}
% \large
\begin{aligned}
&q(v_n|v_{n-1})= \mathbf{N}(v_n; \sqrt{1-\gamma v_{n-1}},\gamma \mathbf{I}),  \\
&q(v_N|v_o)= \prod_{n=1}^{N}q(v_n|v_{n-1}),
\end{aligned}
\end{equation}
where $N$ is the total steps of adding noise and $\gamma$ controls the amount of noise added in each step.
As the noise step $n$ increases, the original image $v_o$ gradually loses the visual contextual information, thus increasing the uncertainty of the next token prediction and amplifying the probability of hallucination. After $N$ diffusion steps, we obtain the visually changed samples with diffusion noise $v_{cd} = v_N$.

\vspace{-4mm}
\paragraph{No image.}
As mentioned before, one of the reasons for producing hallucinations is the dominant LLM priors learned from the pre-training phase. To mitigate such hallucinations, M3ID~\cite{favero2024m3id} contrasts the responses of LVLMs with those of LLMs, where the input image is absent. Since no visual information is provided, the LLMs follow the learned priors to generate a related response to the textual query. The contrastive decoding process is as follows:
\begin{equation}
\label{equ:decoding_no_image}
\begin{aligned}
p_{cd}(y_t|v, v_{cd}, x, y_{<t})&=\text{softmax}[(1+\alpha)(logit_\theta(y_t|v,x,y_{<t})\\
&-\alpha(logit_{\theta_{LLM}}(y_t|x,y_{<t})],
\end{aligned}
\end{equation}
where $\theta_{LLM}$ denotes parameters of the pre-trained LLM.

\vspace{-4mm}
\paragraph{Image downsampling.}
In addition to the above two methods, we propose downsampling images to a lower resolution, where high-frequency details are reduced, to serve as another type of visually changed sample. Given an image size, we first downscale each side of the image by ratio $\frac{1}{r}$ and then directly upscale the output by ratio $r$ to meet the original size. The downsampling process can be denoted as:
\begin{equation}
\label{equ:downsample}
% \large
\begin{aligned}
v_{cd} = resize(resize(v, \frac{1}{r}), r),
\end{aligned}
\end{equation}
where $resize()$ is the function to resize the image by the given ratio. In our experiments, we set $r=32$ by default.

\vspace{-4mm}
\paragraph{Image editing.}
Contrasting with the existing methods, image editing provides another aspect of creating visually changed samples. Recently, text-driven image editing models~\cite{brooks2022instructpix2pix,geng2024instructdiffusion,lin2024text} present impressive ability to edit specific contents grounded with the given text instruction or captions without losing the similarity to the original image.
We apply one of the text-driven image editing models, Instructpix2pix~\cite{brooks2022instructpix2pix}, to edit the original visual input $v$ given the editing textual instructions $t_{edit}$, and regard the resultant edited image as a visually changed sample for contrastive decoding. The editing process can be denoted as:
\begin{equation}
\label{equ:image editing}
% \large
\begin{aligned}
v_{cd} = edit(v, t_{edit}).
\end{aligned}
\end{equation}
The instructions can be obtained in different ways based on the downstream tasks. The instruction for question-answering tasks (which we mainly focus on) is the queried objects or contents from the questions.
% , while we leverage the generated descriptions as instructions to edit images for generation tasks like image captioning.
We provide more examples of instructions and their corresponding edited images in the supplementary material.

% \subsection{Plausibility Constraints}
% Based on the contrastive decoding process from \eqref{equ:contrastive_decoding}, a potential issue emerges as it penalizes the entire LVLM's responses via contrast with visually changed samples. However, either original or visually changed samples may still generate words related to linguistics standards and common sense. Inaccurate punishment of these words may cause an unpredictable and even failed response. To tackle this issue, we follow previous works~\cite{li2023cd, leng2024vcd} to introduce the plausibility constraint, which avoids the indiscriminate penalization caused by directly applying contrastive decoding to entire responses. The plausibility constraint is triggered by the confidence related to candidate tokens of original visual inputs, which can be formulated as:
% \begin{equation}
% \label{equ:plausibility_constraint}
% % \large
% \begin{aligned}
% &V_{cond}(y<t) = \{y_t \in V: \\
% &p_\theta(y_t|v, x, y_{<t}) \geq \beta~\underset{\omega}{max}~p_\theta(\omega|v, x, y_{<t})\}  \\
% &p_{cd}(y_t|v, v_{cd}, x, y_{<t}) = 0 ~~\text{if} y_t \notin V_{cond}(y<t)
% \end{aligned}
% \end{equation}

% where the probability of other candidates below the constraints is replaced with 0.
% With the plausibility constraint, contrastive decoding influences the next token predicting behaviors only when the LVLMs are not confident with their prediction.

\begin{figure*}[!t]
    \centering
    \LARGE
    \resizebox{1.0\linewidth}!{
        \begin{tabular}{cc}
        % \multirow{3}{*}{(a)} 
        \begin{tabular}{c}\includegraphics[width=\textwidth]{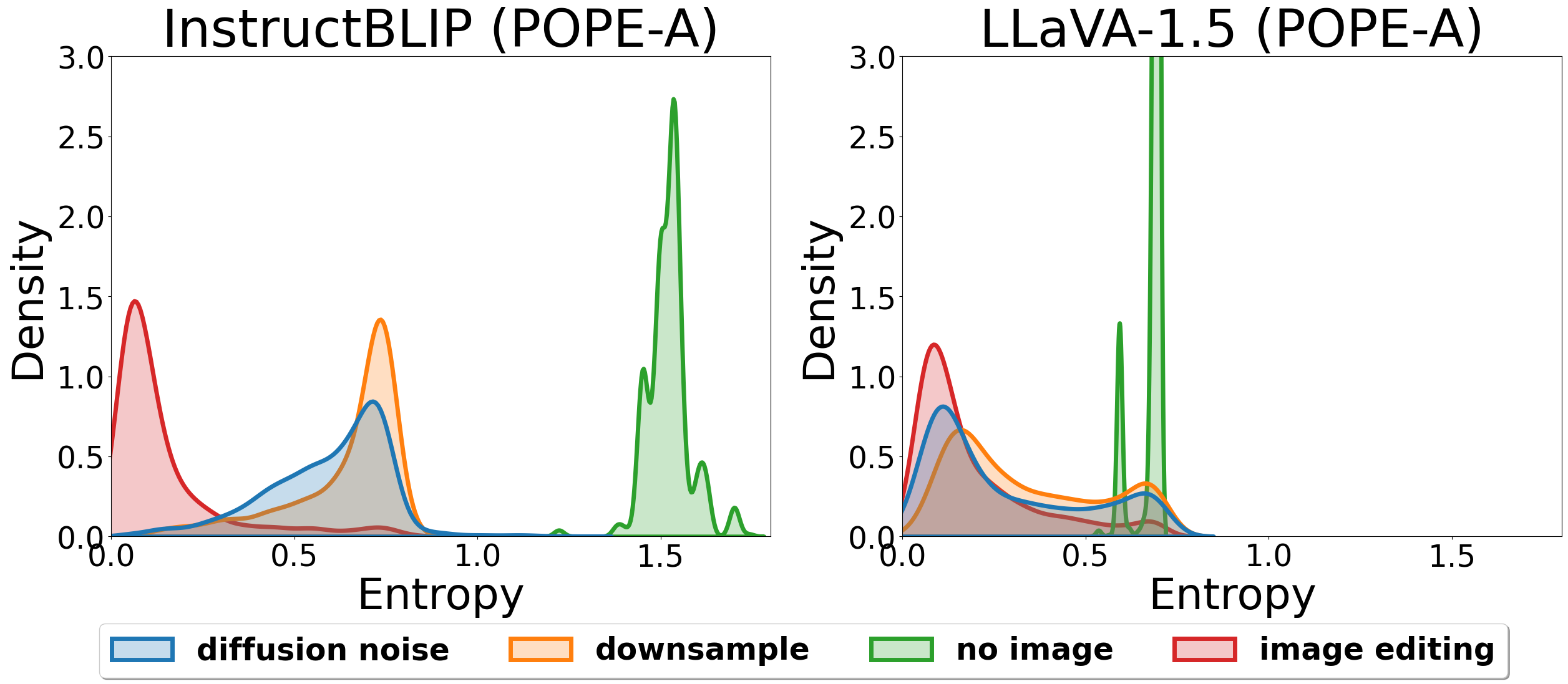}\end{tabular}
        &\begin{tabular}{c}\includegraphics[width=\textwidth]{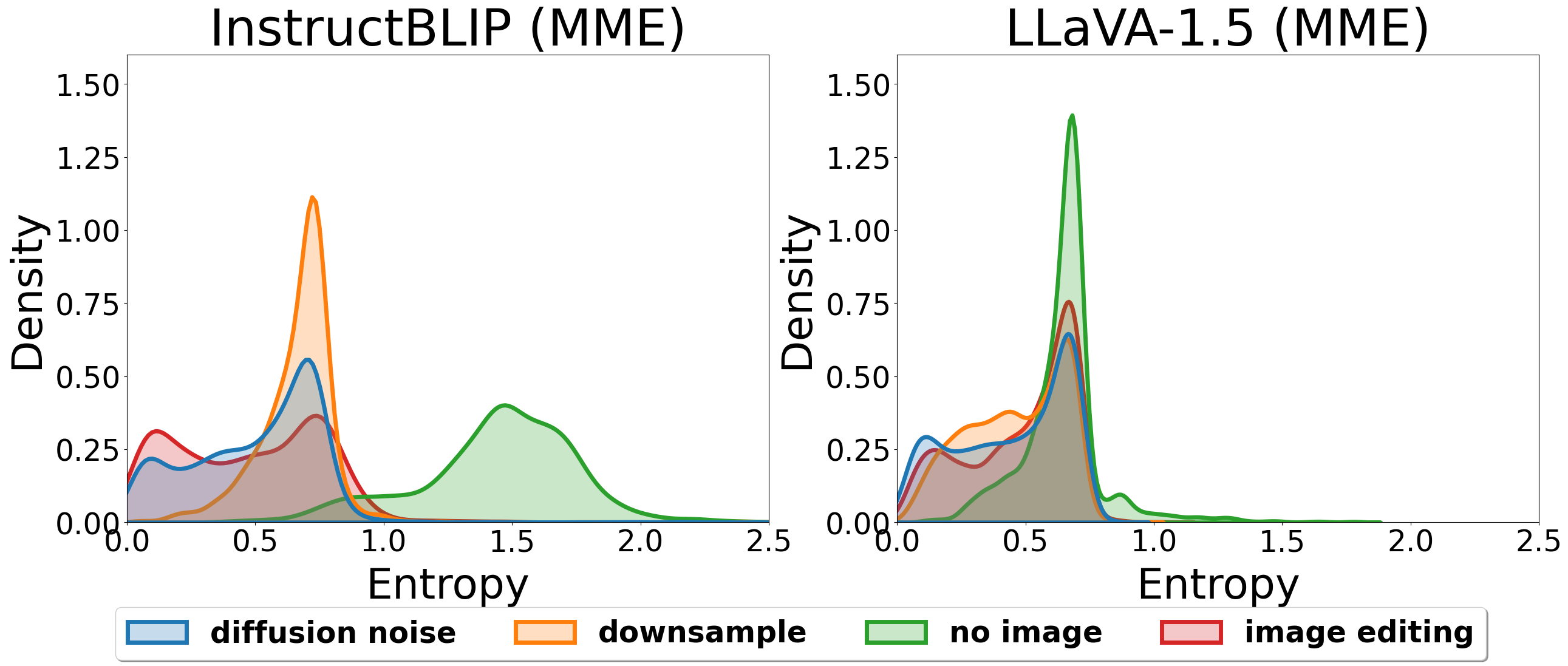}\end{tabular}\\
      
        (a) POPE-adversarial & (b) MME
        \end{tabular}
    }
    \caption{The entropy analysis on the POPE and MME benchmarks.
    }
    \label{fig:entropy}
\end{figure*}

\section{Analysis on Visual Contrastive Encoding}
Although prior studies (e.g., VCD~\cite{leng2024vcd} and M3ID~\cite{favero2024m3id}) have demonstrated impressive improvement on LVLMs via contrastive decoding with visually changed samples (e.g., adding noise and removing the image, respectively), The effectiveness of using different visual CD methods to mitigate the hallucination issue is not well explored yet. 
As mentioned in Sec.~\ref{sec:intro}, we observe that on different benchmarks (POPE~\cite{li2023pope} and MME~\cite{fu2023mme}), the performance gain brought from different visually changed samples varies across LVLMs and tasks.

We hypothesize that applying different visually changed samples yields varying influences on mitigating hallucinations. Thus, we conduct several analysis on the prediction-level statistics to understand behaviors of using each visually changed sample, including entropy/confidence, probability distribution distance, revised outcomes' behaviors, and the pair-wise relationship of rectified answers. To better investigate this without randomness, we focus on the first token of the response using the Yes-No Question task (i.e., POPE and MME), where the next token is selected with the highest probability.

\subsection{Benchmarks and LVLMs}
\paragraph{Benchmarks.}
% We choose two well-known benchmarks for studying the hallucination effect, POPE and MME. 
POPE~\cite{li2023pope}, the Polling-based Object Probing Evaluation, aims to assess the object hallucination. This benchmark consists of Yes-No question-answering pairs related to the corresponding image, querying the LVLM to answer whether a specific object exists in the image or not. The ratio of Yes and No questions is balanced (i.e., 50\% vs 50\%). Within this benchmark, there are three different difficulties of queried non-existent objects, including \textit{random}, \textit{popular}, and \textit{adversarial}. In the \textit{random} setting, the non-existent objects are selected from all the objects in datasets randomly, while the \textit{popular} setting selects non-existent objects from the high-frequency pool. For the \textit{adversarial} setting, the non-existent objects are selected from the co-occurrence objects related to the queried image scene.

MME~\cite{fu2023mme}, a more challenging benchmark for assessing hallucination from multiple perspectives. It contains 14 Yes-No question-answering tasks, including four coarse-grained perception tasks, six fine-grained perception tasks, and four cognition-focus tasks.

\vspace{-4mm}
\paragraph{LVLM backbones.}
We follow prior works~\cite{leng2024vcd} to evaluate the prediction tendency and behaviors of different visually changed samples using InstructBLIP~\cite{instructblip}, LLaVA-1.5~\cite{llava15}, and Qwen-VL~\cite{bai2023qwen}. 
Specifically, these LVLMs are comprised of three components: image encoder, visual-textual interface, and language decoder. The main difference among these three LVLMs is the design of the visual-textual interface: InstructBLIP uses the Q-Former (a transformer) to extract task-relevant image features related to the textual query; LLaVA-1.5 directly projects visual embedding into language embedding with multi-layer perception (MLP), and Qwen-VL utilizes a position-aware vision-language adapter (a cross-attention layer) to compress image features. These differences in the visual-textual interface may result in different output behaviors of visually changed samples.
To better illustrate the observation within the page limit, we discuss InstructBLIP and LLaVA-1.5 in the manuscript, leaving comparisons with all LVLMs in the supplementary materials.

% \begin{figure*}[!t]
%     \centering
%     \LARGE
%     \resizebox{1.0\linewidth}!{
%         \begin{tabular}{cc}
%         % \multirow{3}{*}{(a)} 
%         \begin{tabular}{c}\includegraphics[width=\textwidth]{figures/confidence/all_cd_conf_pope_a.png}\end{tabular}
%         &\begin{tabular}{c}\includegraphics[width=\textwidth]{figures/confidence/all_cd_conf_mme.png}\end{tabular}\\
      
%         (a) POPE-adverarial & (b) MME
%         \end{tabular}
%     }
%     \caption{The analysis of confidence analysis on POPE and MME benchmarks.
%     }
%     \label{fig:confidencew}
% \end{figure*}

\subsection{Analysis with Entropy}
\label{sec:entropy}
We investigate the behavior of each visually changed sample in terms of entropy (i.e., confidence).
It is acknowledged that entropy represents the uncertainty of the prediction. Previous work~\cite{leng2024vcd} demonstrates that visual uncertainty amplifies the language priors and statistical bias, which leads to hallucinations. The entropy can serve as an indicator to determine the extent of inducing hallucinations given different visually changed samples. To delve into more concrete observations using entropy, we visualize the entropy distribution of the first token in the Yes-No question, as shown in Figure~\ref{fig:entropy}.
Visually changed samples have distinct properties of entropy across different LVLMs and benchmarks, which reflects the design of manipulation on visual inputs.
We draw several observations:\\
\indent 1)  \textbf{No image} method only contrasts with the response from the pre-trained LLM without image information, producing more noisy and unpredictable outputs. \\
\indent 2) \textbf{Downsample} and \textbf{diffusion noise} methods have similar entropy distributions, indicating a similar effect of losing some image information to increase the uncertainty of output predictions. \\
% although the modifications on images are different methods (i.e., downsample and adding noise), the effects of losing image information to increase the uncertainty of output prediction are similar. \\
\indent 3) Different from other visually changed samples, the \textbf{image editing} method provides a more specific modification related to the textual instruction extracted from the textual query of LVLMs, where the editing goal is to insert the queried objects into the image. Hence, the entropy is very low in comparison with others on POPE, which only focuses on object existence. In contrast, the MME benchmark contains various perception and cognition tasks; the editing difficulty of the queried fine-grained contents increases, and thus, edited images may fail to be consistent with the original images. \\
% and grounded to queried contents.\\

\begin{figure*}[!t]
    \centering
    \LARGE
    \resizebox{1.0\linewidth}!{
        \begin{tabular}{cc}
        % \multirow{3}{*}{(a)} 
        \begin{tabular}{c}\includegraphics[width=\textwidth]{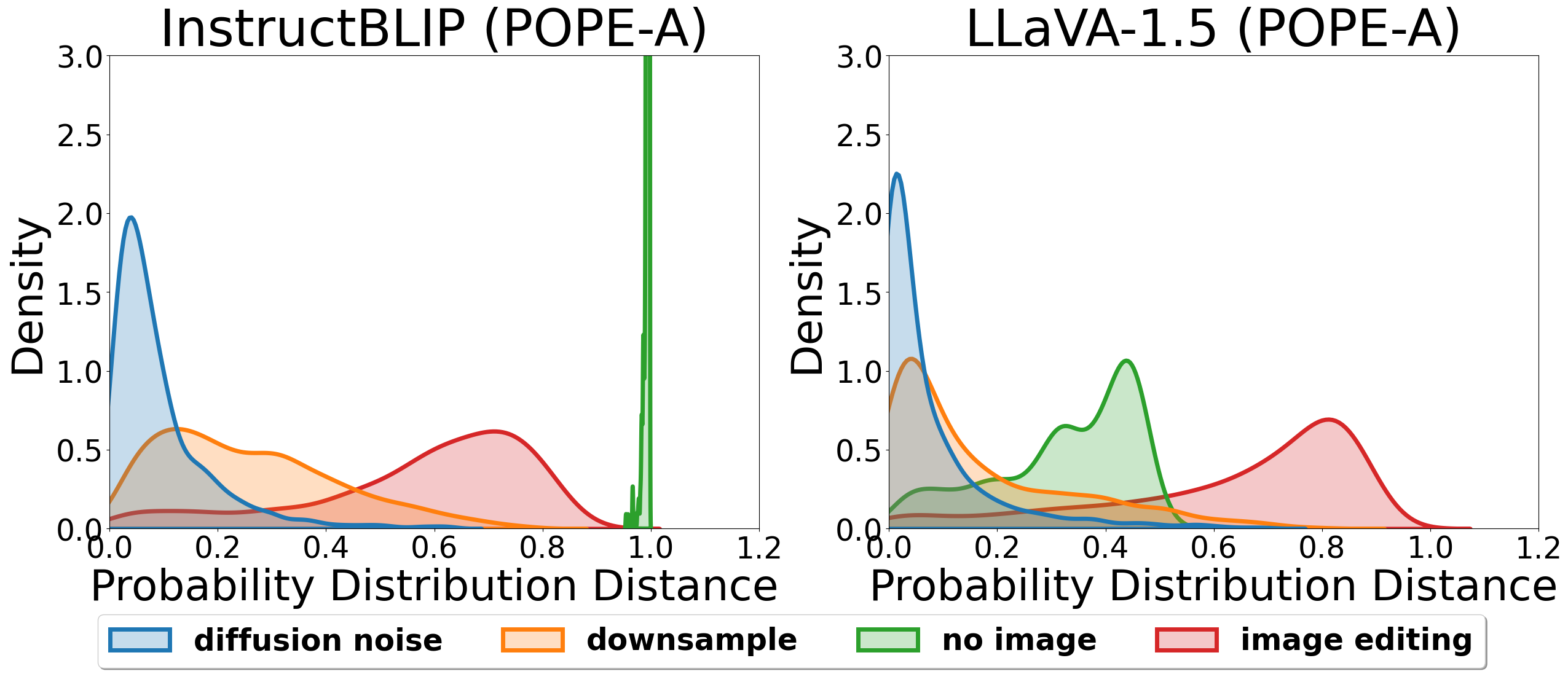}\end{tabular}
        &\begin{tabular}{c}\includegraphics[width=\textwidth]{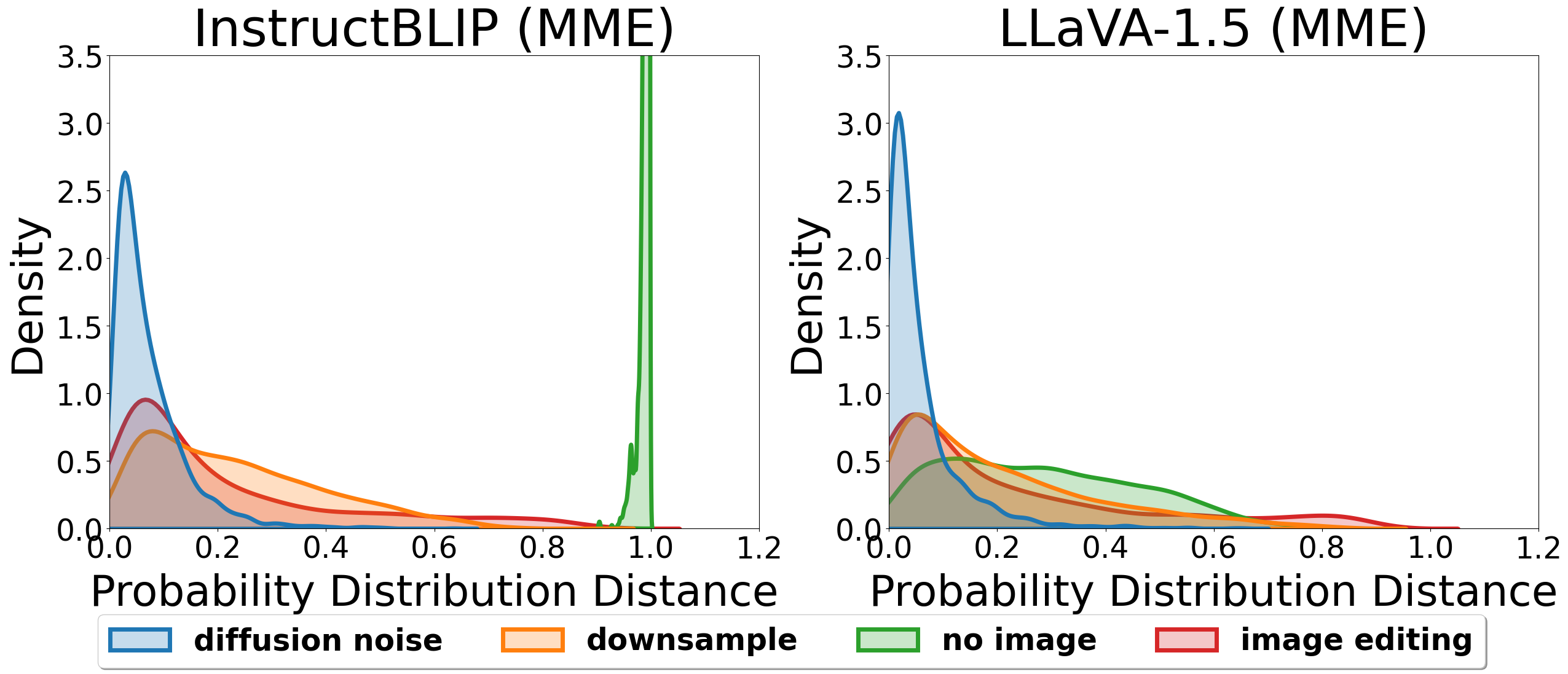}\end{tabular}\\
      
        (a) POPE-adversarial & (b) MME
        \end{tabular}
    }
    \caption{The analysis of probability distribution distance on the POPE and MME benchmarks.
    }
    \label{fig:probdist}
\end{figure*}

\begin{figure*}[!t]
    \centering
    \Huge
    \resizebox{1\linewidth}!{
        \begin{tabular}{cccc}
        % \multirow{3}{*}{(a)} 
        \begin{tabular}{c}\includegraphics[width=0.7\textwidth]{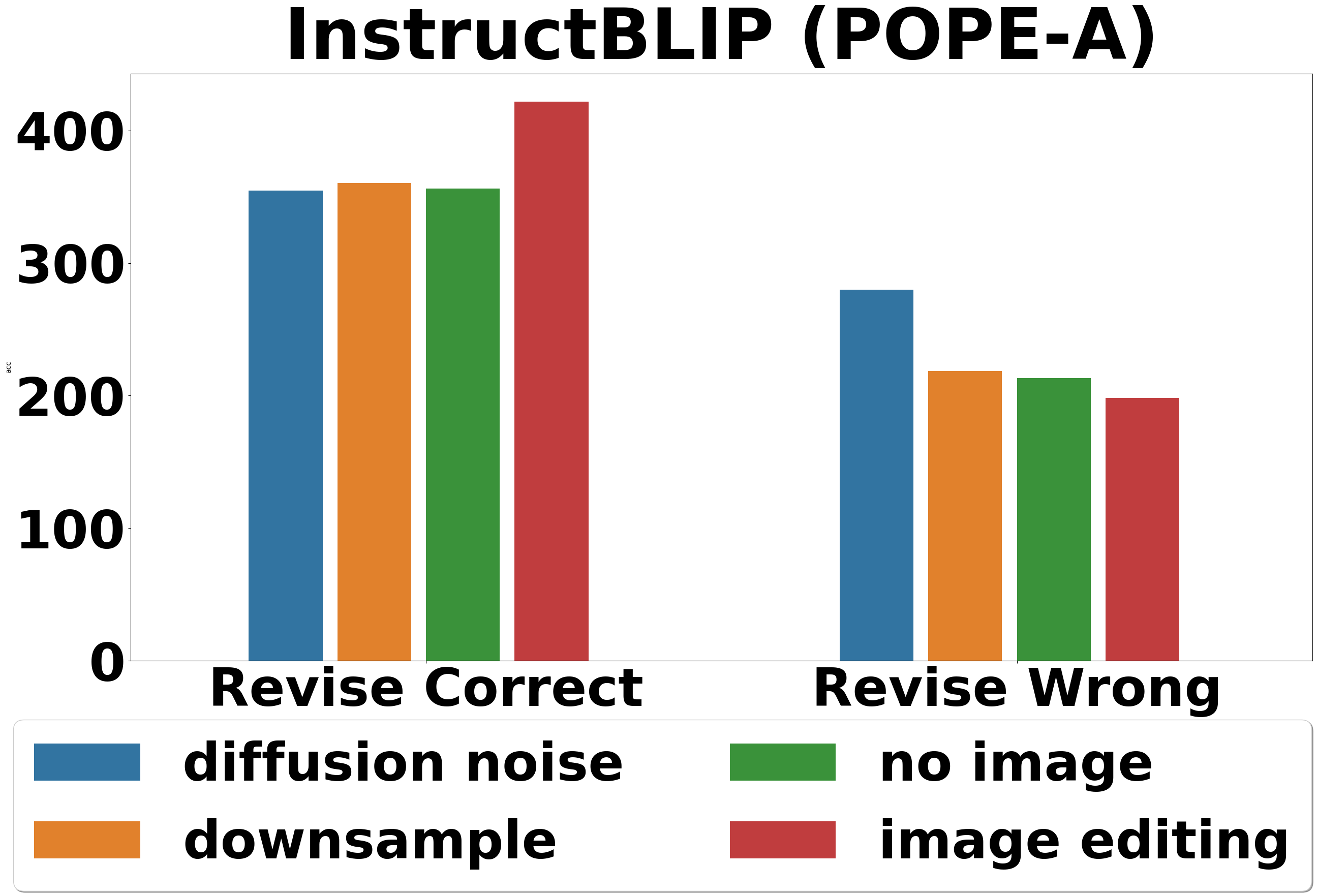}\end{tabular}
        &\begin{tabular}{c}\includegraphics[width=0.7\textwidth]{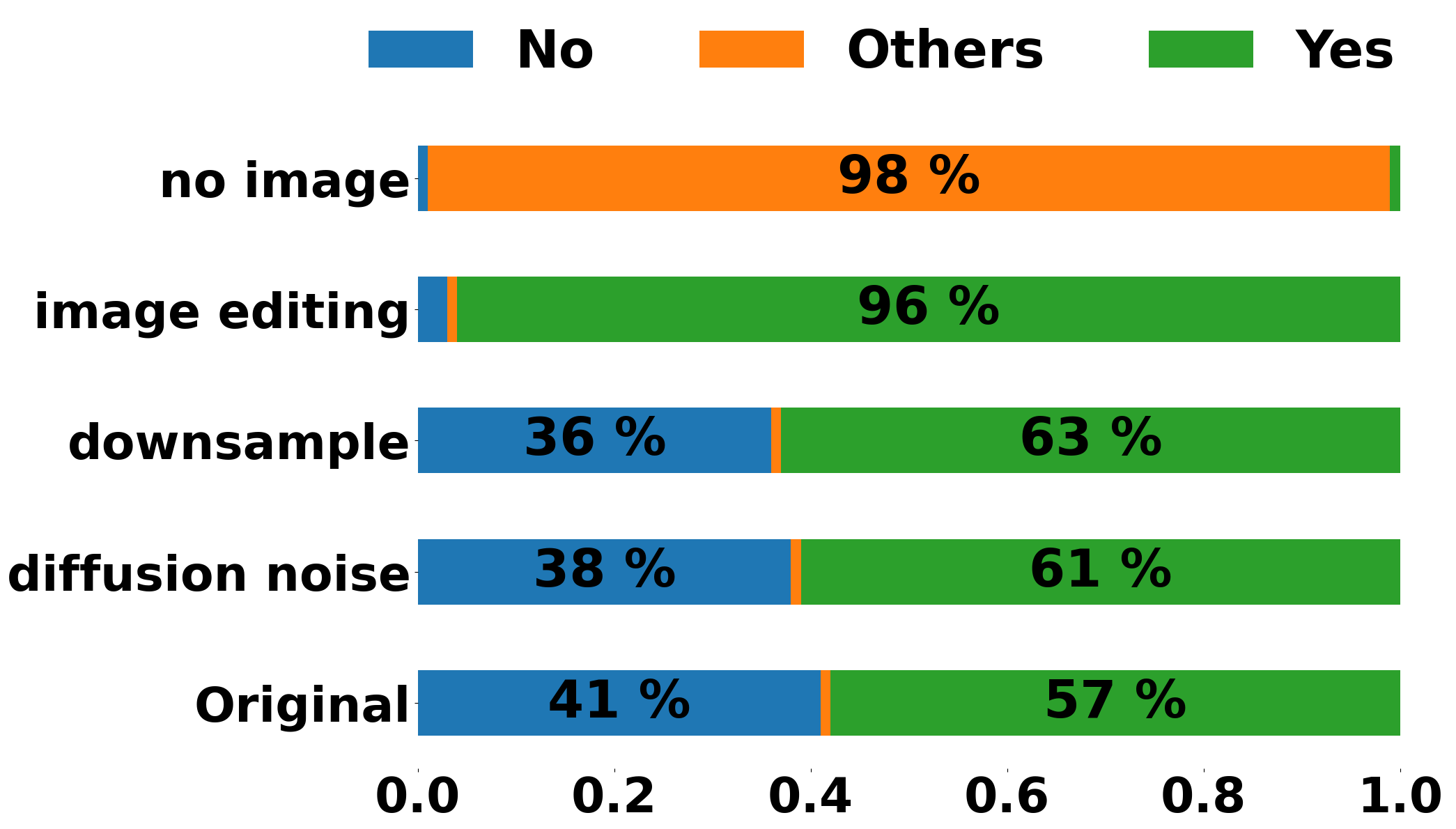}\end{tabular}
        &\begin{tabular}{c}\includegraphics[width=0.7\textwidth]{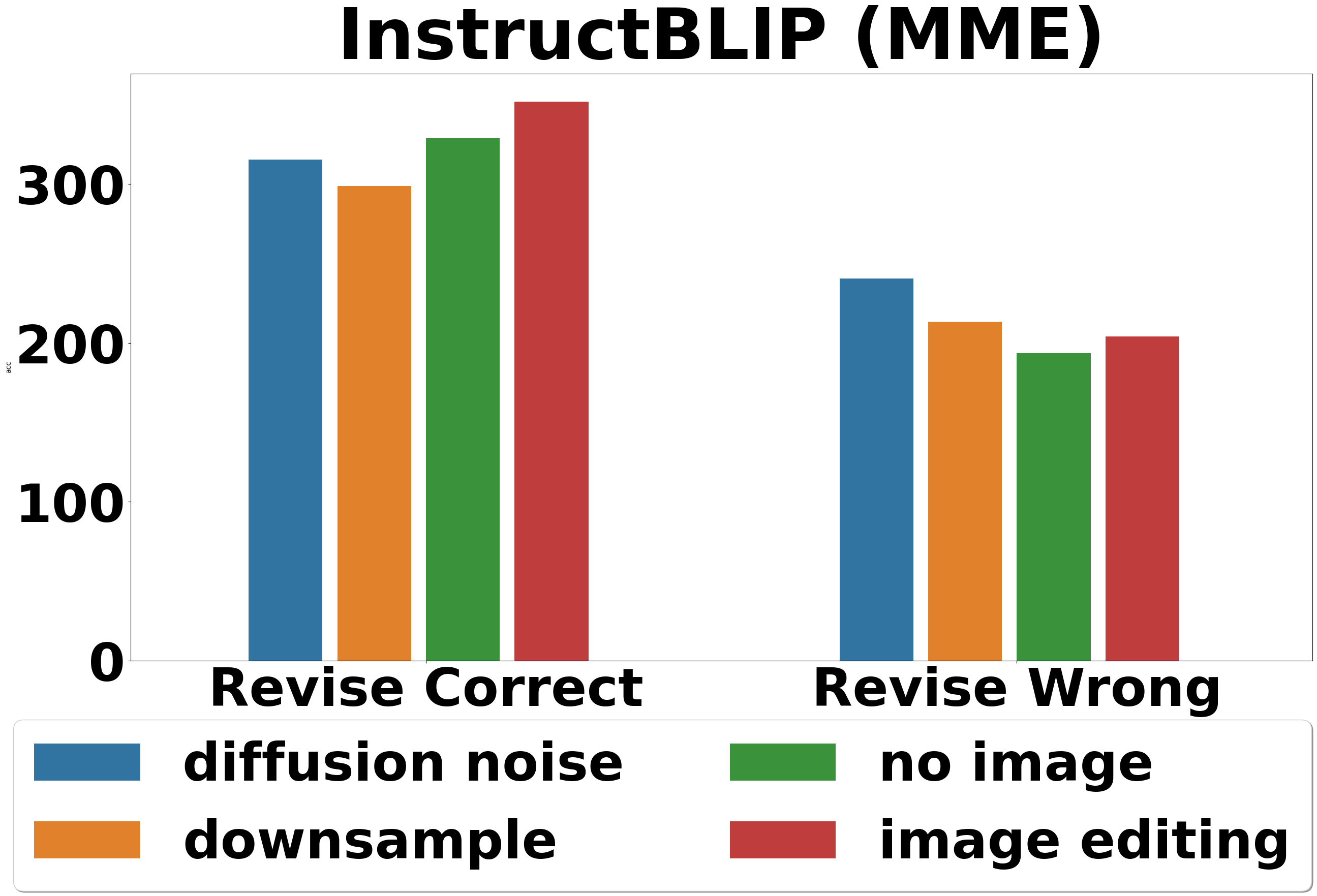}\end{tabular}
        &\begin{tabular}{c}\includegraphics[width=0.7\textwidth]{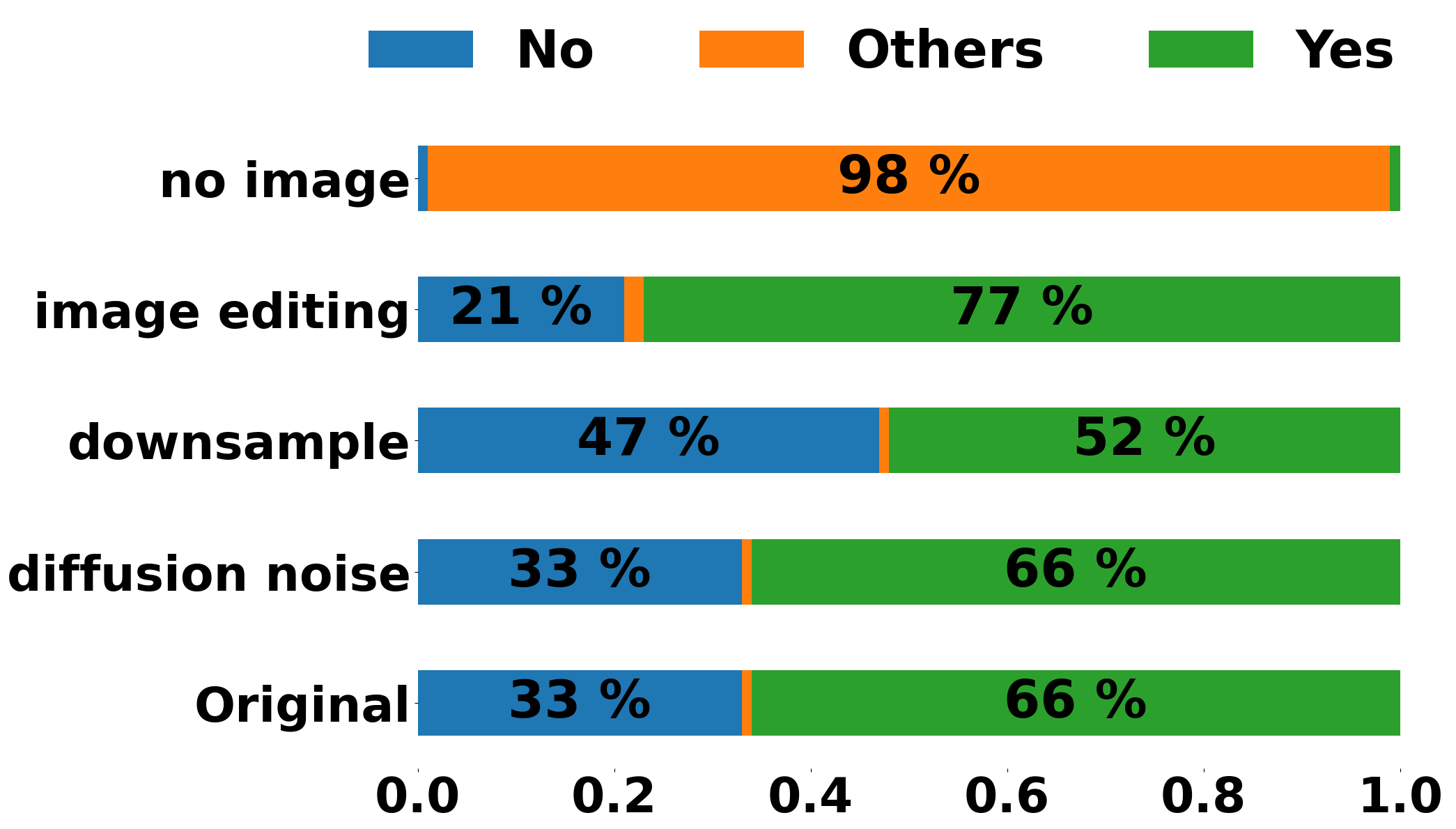}\end{tabular}\\
      
         (a) Revision behaviors on POPE &(b) Answering Tendency on POPE & (c) Revision behaviors on MME & (d) Answering Tendency on MME
        \end{tabular}
    }
    \caption{The revision behaviors and answering tendency of InstructBLIP with different visually changed samples.
    }
    \label{fig:behaviors}
\end{figure*}

\subsection{Analysis with Probability Distribution Distance}
\label{sec:PDD}
In addition to the entropy, we further measure the probability distribution distance (PDD) between the original and CD predictions, which is also investigated in ~\cite{favero2024m3id}. Here, we mainly use the Hellinger distance to estimate the distance between probability distributions, defined as $H(p,q)=2^{1/2}\sqrt{\sum_{i=1}^k(\sqrt{p_i} -\sqrt{p_j})^2}$, where $p=(p_i)_{i \in  \left|k \right|}$ and $q=(q_i)_{i \in \left|k \right|}$ are discrete probability distributions. Intuitively, when the distance increases, the prediction distribution of the original visual input and visually changed sample differs more, leading to a high risk of obtaining hallucination results from visually changed samples. 
According to the results shown in Figure~\ref{fig:probdist}, we draw several observations:\\
\indent 1) \textbf{No image} method presents distinct distributions of PDD with InstructBLLIP and LLaVA-1.5. This phenomenon indicates that the language priors learned from LVLMs are different. \\
\indent 2) \textbf{Diffusion noise} method has PDD distribution centralized at lower PDD values, demonstrating that even the uncertainty of visual inputs is enhanced with added noises, the prediction distribution is still similar to the original input. \textbf{Downsample} method also has lower PDD values, where the overall distribution is smoother, meaning that parts of downsampled samples change their predictions due to the lost of high-frequency details. \\
\indent 3) Different from others, \textbf{image editing} method obtains the PDD distribution with higher values on POPE but lower values on MME. As mentioned previously, editing a specific object on images is easier than editing fine-grained contents.
Therefore, edited images with non-existent objects results in a positive response in LVLMs for answering the corresponding question that queries non-existent objects.
% , \alan{which differs from the negative or ambiguous response predicted from the original visual input.}
On the contrary, editing methods may fail to edit the image with fine-grained contents, leading to similar visual contents, thus having a closer prediction distribution to the original one. \\

% \begin{figure*}[!t]
%     \centering
%     % \LARGE
%     \resizebox{0.8\linewidth}!{
%         \begin{tabular}{cc}
%         % \multirow{3}{*}{(a)} 
%         \begin{tabular}{c}\includegraphics[width=\textwidth]{figures/overlap/instructblip_revise_correct_pope_a.png}\end{tabular}
%         &\begin{tabular}{c}\includegraphics[width=\textwidth]{figures/overlap/instructblip_revise_correct_mme_all.png}\end{tabular}
%         \end{tabular}
%     }
%     \caption{The pairwise overlap of revising-correct samples. CD types consist of (A) diffusion noise, (B) downsample, (C) no image, and (D) image editing.
%     }
%     \label{fig:overlapping}
%     \vspace{-3mm}
% \end{figure*}

\subsection{Revision Behaviors}
To investigate the behavior of each contrastive sample, we focus on three metrics: revise-correct samples, revise-wrong samples, and the predicting tendency on the POPE and MME benchmarks. Revise-correct samples indicate that the contrastive decoding revises the original response with the correct answers, while revise-wrong samples present the opposite case. Predicting tendency is the tendency of LVLMs to answer Yes-No Questions with or without contrastive decoding, including yes, no, and others. The results are shown in Figure~\ref{fig:behaviors}.
We draw several observations:\\
\indent 1) As shown in Figure~\ref{fig:behaviors} (a) and (c), the results present that the contrastive decoding with different visually changed samples adjusts both correct and incorrect answers, where the revise-correct samples are more than revise-wrong ones. This phenomenon reveals that: although these CD methods appear to improve the overall performance, it is because that the number of revise-correct samples outweighs the number of revise-wrong samples. However, in practical applications, the revise-wrong case also emerges as a risk for misleading users.
\\
\indent 2) \textbf{downsample} and \textbf{diffusion noise} methods have more revise-wrong samples than the other two methods, which is consistent with the observation on PDD. The visually changed samples with lower PDD values present similar output prediction as the original input, leading to a high risk of subtracting the probability of correct answers. In contrast, the \textbf{image editing} method has PDD distribution with large values, indicating the dissimilar output prediction to the original prediction. Therefore, when the original prediction is correct, contrastive decoding with image editing samples may not affect the output distribution, resulting in fewer revise-wrong samples.\\
\indent 3) We show the answering tendency of each visually changed sample in Figure~\ref{fig:behaviors} (b) and (d). We find that samples with loss of information (i.e., downsample and diffusion noise methods) tend to predict yes answers more, while samples with dramatic changes like no image or inserting queried specific objects/contents have nearly unified answers, e.g., almost ``yes'' for image editing and ``others'' for no image. Such varying prediction tendencies suggest that the output prediction affected by these CD methods are different, and there are likely existing complementary properties among these visually changed samples. \\

% Key observations:
% \begin{itemize}
%     \item provides the answering tendency, including the ratio of answering Yes, No, and None of both.
%     \item image editing tends to response with Yes, no image tends to response with system tokens (or LLM priors)
%     \item the amount of revising correct and wrong varies across different visually changed samples
%     \item M3ID influence the least, while image editing has the most revision samples.
%     \item consist with answering tendency, image editing 
% \end{itemize}

\subsection{Pairwise Overlap of Rectified Answers}
\label{sec:overlap}
Based on the above observation of revision behaviors, we investigate the complementary properties among different CD methods.
Specifically, we calculate the pairwise overlap of rectified answers among different visually changed samples. That is, the Jaccard similarity coefficient (known as IoU) is computed in the revise-correct set using any of the two CD methods. As shown in Figure~\ref{fig:overlapping}, a lower score means a higher complement between two CD methods.
% investigate the pairwise overlap of rectified samples to validate the relationship among output among different visually changed samples.
% We calculate the Jaccard similarity coefficient (known as IoU) between the revise-correct sets of any two visually changed samples, which can be derived from the following:
% \begin{equation}
% \label{equ:iou}
% % \large
% \begin{aligned}
% overlap = \frac{\left|A \cap B\right|}{\left| A \cup B \right|}
% \end{aligned}
% \end{equation}
% The heatmap of pairwise overlap is shown in Figure~\ref{fig:overlapping}.
Interestingly, results show that most pairwise scores are low, demonstrating the distinct effect of each CD method to rectify answers.
% This indicates that different visually changed samples may have different impacts on decoding results, where such differences are not predictable yet complementary among datasets and LVLMs based on previous analyses.
This evidence motivates us to design a combined approach that leverages the strengths of each contrastive sample.
% suggests that different visually changed samples have their own strengths in different scenarios, motivating us to design a combined approach that leverages the strengths of each contrastive sample.

\begin{figure}[!t]
    \centering
    \includegraphics[width=1\linewidth]{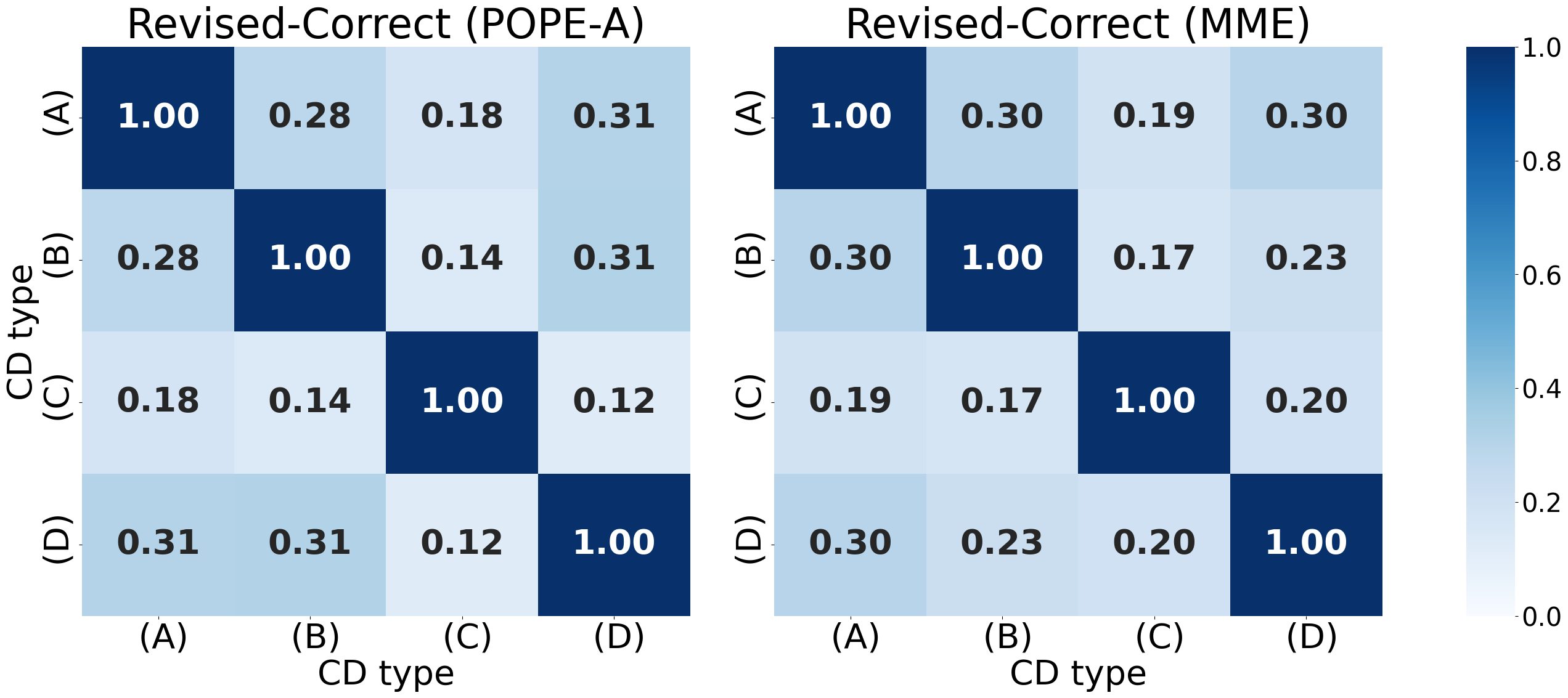}
    \caption{The pairwise overlap of revise-correct samples. CD types consist of (A) diffusion noise, (B) downsample, (C) no image, and (D) image editing.}
    \label{fig:overlapping}
    \vspace{-3mm}
\end{figure}

\begin{table*}[t]
\resizebox{\linewidth}!{
    \begin{tabular}{c|c|c|c|c|c|c|c|c|c|c|c|c|c}
    VQA (Y/N)                & \multicolumn{4}{c|}{InstructBLIP} & \multicolumn{4}{c|}{LLaVA-1.5} & \multicolumn{4}{c|}{QwenVL}    & Overall \\ \hline
    Benchmark (Acc)            & P-R    & P-P    & P-A    & MME   & P-R   & P-P   & P-A   & MME   & P-R   & P-P   & P-A   & MME   & All     \\ \hline
    Original                   & 0.825  & 0.733  & 0.752  & 0.688 & 0.832 & 0.817 & 0.788 & 0.705 & 0.847 & 0.851 & 0.822 & 0.720 & 0.785   \\ \hline
    % ICD~\cite{wang2024icd}     & 0.884  & 0.817  & 0.792  & 0.726 &       &       &       &       &       &       &       &       &         \\
    diffusion noise (VCD~\cite{leng2024vcd})     & 0.858  & 0.805  & 0.779  & 0.720 & 0.863 & 0.842 & 0.805 & 0.733 & 0.865 & 0.865 & 0.832 & 0.747 & 0.810   \\
    no image (M3ID~\cite{favero2024m3id}) & \second{0.880}  & 0.835  & 0.802  & 0.742 & 0.871 & \best{0.856} & \best{0.827} & 0.732 & 0.868 & \second{0.871} & 0.837 & 0.733 & 0.821   \\
    downsample                 & 0.878  & 0.824  & 0.801  & 0.724 & \best{0.879} & \best{0.856} & 0.811 & \second{0.765} & \best{0.875} & \second{0.871} & \second{0.842} & \second{0.764} & 0.824   \\
    image editing                  & 0.860  & 0.839  & 0.829  & 0.747 & 0.832 & 0.826 & 0.811 & 0.748 & 0.845 & 0.838 & 0.824 & 0.763 & 0.814   \\ \hline
    naive fusion              & 0.864  & 0.840  & \best{0.832}  & \best{0.774} & 0.862 & 0.847 & \best{0.827} & \best{0.779} & 0.865 & 0.860 & 0.839 & \best{0.768} & \second{0.830}   \\
    entropy-weighted fusion    & \best{0.890}  & \best{0.857}  & 0.828  & \best{0.774} & 0.872 & \best{0.856} & 0.824 & 0.764 & \second{0.872} & \best{0.873} & \best{0.844} & 0.749 & \best{0.833}   \\
    PDD-weighted fusion   & 0.876  & \second{0.849}  & \second{0.830}  & \second{0.760} & \second{0.872} & \second{0.849} & \second{0.825} & 0.755 & 0.866 & 0.861 & 0.838 & 0.762 & 0.829  
\end{tabular}
}
    \caption{The results on the Yes-No question tasks, including POPE and MME. Please refer to Sec. \ref{sec:naive_fuse} and \ref{sec:entropy_fuse} for the formulation of fusion methods. P-R denotes the \textit{random} task of POPE, P-P denotes the \textit{popular} task of POPE, and P-A denotes the \textit{adversarial} task of POPE. The best performance within each set is marked \textbf{bold}, and the second highest performance is \underline{underlined}.
    }
    \label{table:quantitative}
\end{table*}

\begin{table}[t]
\resizebox{0.7\linewidth}!{
    \begin{tabular}{c|c}
    VQA (Y/N)                 & Overall \\ \hline
    Benchmark (Acc)             & All     \\ \hline
    Original                    & 0.785   \\ \hline
    % ICD~\cite{wang2024icd}     & 0.884  & 0.817  & 0.792  & 0.726 &       &       &       &       &       &       &       &       &         \\
    diffusion noise (VCD~\cite{leng2024vcd}) (A)    & 0.810   \\
    no image (M3ID~\cite{favero2024m3id}) (B) & 0.821   \\
    downsample (C)                & 0.824   \\
    image editing (D)                 & 0.814   \\ \hline \hline
    % A+B              & 0.  & 0.  & 0.  & 0. & 0. & 0. & 0. & 0. & 0. & 0. & 0. & 0. & 0.      \\ 
    % A+B+C              & 0.  & 0.  & 0.  & 0. & 0. & 0. & 0. & 0. & 0. & 0. & 0. & 0. & 0.    \\   
    % A+B+C+D              & 0.  & 0.  & 0.  & 0. & 0. & 0. & 0. & 0. & 0. & 0. & 0. & 0. & 0.     \\  
    % simple fusion              & 0.864  & 0.840  & 0.832  & 0.774 & 0.862 & 0.847 & 0.827 & 0.779 & 0.865 & 0.860 & 0.839 & 0.768 & 0.830   \\ \hline \hline
    A+B               & 0.826    \\   
    A+B+C               & \second{0.829}      \\ 
    A+B+C+D    & \best{0.833}     \\  
    % entropy-weighted fusion    & 0.890  & 0.857  & 0.828  & 0.774 & 0.871 & 0.856 & 0.824 & 0.764 & 0.871 & 0.872 & 0.844 & 0.749 & 0.833   \\

\end{tabular}
}
    \caption{The results of overall performance on the Yes-No question tasks, including POPE and MME. The results below are entropy-weighted fusion methods with different combinations of visually changed samples. The best performance within each set is marked \textbf{bold}, and the second highest performance is \underline{underlined}.
    }
    \label{table:combinations}
\end{table}

\section{Fusion of Visual Contrastive Decoding}
% refer to the observations above, we find that the entropy is more representative on contrastive decoding behaviors of different visually changed samples. 
According to the observations studied above, we find that one visually changed sample may exhibit better performance under some settings due to its distinct behaviors and output distributions.
Instead of manually selecting the best contrastive decoding method when encountering new tasks or datasets, we propose to apply all the visually changed samples and fuse their influence, which can be a more general approach to mitigate hallucinations in practice.

\subsection{Naive Fusion}
\label{sec:naive_fuse}
An intuitive approach to fuse the effect of different visually changed samples is to combine them with the same weight. According to \eqref{equ:contrastive_decoding}, the decoding with fusion of several visually changed samples can be derived as:
\begin{equation}
\label{equ:direct_fuse}
\begin{aligned}
p_{cd}(y_t|v, v_{cd}, x, y_{<t})&=\text{softmax}[(1+\alpha)logit_\theta(y_t|v,x,y_{<t})\\
&-\alpha\sum_{i=1}^{k}logit_\theta(y_t|v_{cd_i},x,y_{<t})],
\end{aligned}
\end{equation}
where the $\{v_{cd_1},v_{cd_2},...,v_{cd_k}\}$ is the set of different visually changed samples.

\subsection{Entropy-weighted Fusion}
\label{sec:entropy_fuse}
As studied above, when the entropy is higher, the LVLM is less confident in its response. Visually changed samples with a higher entropy can be more likely to yield the hallucinated response, which is more suitable for serving as a contrastive sample. Therefore, we fuse the different types of visually changed samples with the corresponding entropy of outputs as the weight. We then reformulate \eqref{equ:contrastive_decoding} as:
\begin{equation}
\label{equ:entropy_fuse}
\begin{aligned}
&p_{cd}(y_t|v, v_{cd}, x, y_{<t})=\text{softmax}[logit_\theta(y_t|v,x,y_{<t})\\
&-\sum_{i=1}^{k}e_i(logit_\theta(y_t|v,x,y_{<t})-logit_\theta(y_t|v_{cd_i},x,y_{<t}))],
\end{aligned}
\end{equation}
where $e_i$ is the entropy of output distribution for a visually changed sample $v_{cd_i}$. Note that, a similar formulation can be also derived for the case of using probability distribution distance, in which we will include details in the supplementary material.

% \subsection{Fuse Based on Probability Distribution Difference}

% Our motivations
% \begin{itemize}
%     \item Different CDs have different behaviors, and so do the backbone MLLMs and datasets/tasks

%     \item The optimal CDs for alleviating hallucination of given inputs are different and not predictable since their behavior is not predictable

%     \item based on observations, we categorize the CDs and
% \end{itemize}

% \section{Experiments}

\subsection{Quantitative Results}
We compare our fusion methods with other contrastive decoding methods with the single visually changed sample in Table~\ref{table:quantitative}. We demonstrate the accuracy under several settings with different LVLMs (i.e., InstructBLIP~\cite{instructblip}, LLaVA-1.5~\cite{llava15}, and Qwen-VL~\cite{bai2023qwen}) and benchmarks (POPE~\cite{li2023pope} and MME~\cite{fu2023mme}), where we highlight the averaged number over all settings in the last column as the overall accuracy. 
In every single setting (i.e., each column indicates a specific task and LVLM), applying fusion of all visually changed samples (i.e., methods in the bottom of the table) has the best or second-best performance compared to other single-sample methods, validating our motivations to combine different samples with complementary properties to mitigate hallucinations. In addition, we find that our downsample method reaches the best overall performance among all single-sample methods, showcasing a more effective CD strategy than other existing methods.
% We also observe that image editing reaches the best performance on MME but the worst performance on POPE-R and POPE-P, demonstrating th

Moreover, we compare different fusion methods, including naive fusion, entropy-weighted fusion, and PDD-weighted fusion, where the naive fusion directly combines all the visually changed samples while the latter two fusions are weighted according to the entropy and probability distribution distance of outputs. As described in Sec.~\ref{sec:entropy} and \ref{sec:PDD}, one of the main differences between these two metrics is the distribution of the image-editing method on POPE, which has lower values of entropy but higher values of PDD. Considering the performance in Table~\ref{table:quantitative} and the above observations, we find that entropy-weighted fusion can alleviate the negative influence of image editing on the POPE benchmark via a smaller value of entropy and keep its strength on the MME benchmark with a higher value of entropy. This evidence validates our expectation that entropy represents the uncertainty of outputs correlating with the likelihood of producing hallucinations, which is more suitable as the fusion weights on different visually changed samples.

% Interestingly, fusion with different weights , 
% Concretely, with different metrics serving as fusion weights, the fusion method strikes a balance on the priority of

\subsection{Combinations of Fusion Methods}
We have shown that entropy-weighted fusion with all visually changed samples is more robust to hallucinations than solely applying a specific visually changed sample; however, one question may emerge: ``Do we need to fuse all designed visually changed samples?''.
% As investigated in Sec.~\ref{sec:overlap}, we observe the similarity of rectified answers (i.e., in the revise-correct set) among any pair of visually changed samples is not obvious, varying efficacy for hallucination mitigation, and every piece matters.
To further study this, we conduct an ablation study of different combinations of fusion methods in Table~\ref{table:combinations}, where the combination ranges from fusing two to four visually changed samples. We find that fusion with more visually changed samples yields better overall performance, demonstrating that it is more general method than selecting specific combinations to be deployed for practical applications on novel data and tasks. More results are provided in the supplementary materials.

% \paragraph{Controllable reductions}

% \subsection{Open-ended Caption Generation}
% \paragraph{CHIARs}
% \paragraph{Qualitative results}
\section{Conclusions}
In this paper, we delve into visual contrastive decoding with various visually changed samples to mitigate hallucinations for large vision-language models. We first explore new visually changed samples from different aspects, including downsampling and image editing. We conduct thorough prediction-level analysis to investigate the prediction behaviors of visually changed samples, concluding that different visually changed images exhibit significantly varied suitability across different LVLMs and datasets. To this end, we propose an entropy-weighted fusion method to leverage the complementary property from visually changed samples based on the entropy of predictions, which is proportional to the likelihood of hallucinations. Experimental results showcase the efficacy of our fusion method to mitigate hallucinations of LVLMs.

{
    \small
    \bibliographystyle{ieeenat_fullname}
    \bibliography{main}

\begin{thebibliography}{32}
\providecommand{\natexlab}[1]{#1}
\providecommand{\url}[1]{\texttt{#1}}
\expandafter\ifx\csname urlstyle\endcsname\relax
  \providecommand{\doi}[1]{doi: #1}\else
  \providecommand{\doi}{doi: \begingroup \urlstyle{rm}\Url}\fi

\bibitem[Bai et~al.(2023{\natexlab{a}})Bai, Bai, Chu, Cui, Dang, Deng, Fan, Ge, Han, Huang, Hui, Ji, Li, Lin, Lin, Liu, Liu, Lu, Lu, Ma, Men, Ren, Ren, Tan, Tan, Tu, Wang, Wang, Wang, Wu, Xu, Xu, Yang, Yang, Yang, Yang, Yao, Yu, Yuan, Yuan, Zhang, Zhang, Zhang, Zhang, Zhou, Zhou, Zhou, and Zhu]{qwen}
Jinze Bai, Shuai Bai, Yunfei Chu, Zeyu Cui, Kai Dang, Xiaodong Deng, Yang Fan, Wenbin Ge, Yu Han, Fei Huang, Binyuan Hui, Luo Ji, Mei Li, Junyang Lin, Runji Lin, Dayiheng Liu, Gao Liu, Chengqiang Lu, Keming Lu, Jianxin Ma, Rui Men, Xingzhang Ren, Xuancheng Ren, Chuanqi Tan, Sinan Tan, Jianhong Tu, Peng Wang, Shijie Wang, Wei Wang, Shengguang Wu, Benfeng Xu, Jin Xu, An Yang, Hao Yang, Jian Yang, Shusheng Yang, Yang Yao, Bowen Yu, Hongyi Yuan, Zheng Yuan, Jianwei Zhang, Xingxuan Zhang, Yichang Zhang, Zhenru Zhang, Chang Zhou, Jingren Zhou, Xiaohuan Zhou, and Tianhang Zhu.
\newblock Qwen technical report.
\newblock \emph{arXiv preprint arXiv:2309.16609}, 2023{\natexlab{a}}.

\bibitem[Bai et~al.(2023{\natexlab{b}})Bai, Bai, Yang, Wang, Tan, Wang, Lin, Zhou, and Zhou]{bai2023qwen}
Jinze Bai, Shuai Bai, Shusheng Yang, Shijie Wang, Sinan Tan, Peng Wang, Junyang Lin, Chang Zhou, and Jingren Zhou.
\newblock Qwen-vl: A frontier large vision-language model with versatile abilities.
\newblock \emph{arXiv preprint arXiv:2308.12966}, 2023{\natexlab{b}}.

\bibitem[Brooks et~al.(2023)Brooks, Holynski, and Efros]{brooks2022instructpix2pix}
Tim Brooks, Aleksander Holynski, and Alexei~A. Efros.
\newblock Instructpix2pix: Learning to follow image editing instructions.
\newblock In \emph{IEEE Conference on Computer Vision and Pattern Recognition (CVPR)}, 2023.

\bibitem[Chen et~al.(2024)Chen, Zhao, Luo, Yao, Li, and Zhou]{chen2024halc}
Zhaorun Chen, Zhuokai Zhao, Hongyin Luo, Huaxiu Yao, Bo Li, and Jiawei Zhou.
\newblock Halc: Object hallucination reduction via adaptive focal-contrast decoding.
\newblock \emph{arXiv preprint arXiv:2403.00425}, 2024.

\bibitem[Dai et~al.(2024)Dai, Li, Li, Tiong, Zhao, Wang, Li, Fung, and Hoi]{instructblip}
Wenliang Dai, Junnan Li, Dongxu Li, Anthony Meng~Huat Tiong, Junqi Zhao, Weisheng Wang, Boyang Li, Pascale Fung, and Steven Hoi.
\newblock Instructblip: Towards general-purpose vision-language models with instruction tuning.
\newblock In \emph{Advances in Neural Information Processing Systems (NeurIPS)}, 2024.

\bibitem[Deng et~al.(2024)Deng, Chen, and Hooi]{deng2024seeing}
Ailin Deng, Zhirui Chen, and Bryan Hooi.
\newblock Seeing is believing: Mitigating hallucination in large vision-language models via clip-guided decoding.
\newblock \emph{arXiv preprint arXiv:2402.15300}, 2024.

\bibitem[Favero et~al.(2024)Favero, Zancato, Trager, Choudhary, Perera, Achille, Swaminathan, and Soatto]{favero2024m3id}
Alessandro Favero, Luca Zancato, Matthew Trager, Siddharth Choudhary, Pramuditha Perera, Alessandro Achille, Ashwin Swaminathan, and Stefano Soatto.
\newblock Multi-modal hallucination control by visual information grounding.
\newblock In \emph{IEEE Conference on Computer Vision and Pattern Recognition (CVPR)}, 2024.

\bibitem[Fu et~al.(2023)Fu, Chen, Shen, Qin, Zhang, Lin, Yang, Zheng, Li, Sun, et~al.]{fu2023mme}
Chaoyou Fu, Peixian Chen, Yunhang Shen, Yulei Qin, Mengdan Zhang, Xu Lin, Jinrui Yang, Xiawu Zheng, Ke Li, Xing Sun, et~al.
\newblock Mme: A comprehensive evaluation benchmark for multimodal large language models.
\newblock \emph{arXiv preprint arXiv:2306.13394}, 2023.

\bibitem[Geng et~al.(2024)Geng, Yang, Hang, Li, Gu, Zhang, Bao, Zhang, Li, Hu, et~al.]{geng2024instructdiffusion}
Zigang Geng, Binxin Yang, Tiankai Hang, Chen Li, Shuyang Gu, Ting Zhang, Jianmin Bao, Zheng Zhang, Houqiang Li, Han Hu, et~al.
\newblock Instructdiffusion: A generalist modeling interface for vision tasks.
\newblock In \emph{IEEE Conference on Computer Vision and Pattern Recognition (CVPR)}, 2024.

\bibitem[Gunjal et~al.(2024)Gunjal, Yin, and Bas]{gunjal2024detecting}
Anisha Gunjal, Jihan Yin, and Erhan Bas.
\newblock Detecting and preventing hallucinations in large vision language models.
\newblock In \emph{AAAI Conference on Artificial Intelligence (AAAI)}, 2024.

\bibitem[Ho et~al.(2020)Ho, Jain, and Abbeel]{ho2020denoising}
Jonathan Ho, Ajay Jain, and Pieter Abbeel.
\newblock Denoising diffusion probabilistic models.
\newblock \emph{Advances in neural information processing systems}, 33:\penalty0 6840--6851, 2020.

\bibitem[Huang et~al.(2024)Huang, Dong, Zhang, Wang, He, Wang, Lin, Zhang, and Yu]{huang2024opera}
Qidong Huang, Xiaoyi Dong, Pan Zhang, Bin Wang, Conghui He, Jiaqi Wang, Dahua Lin, Weiming Zhang, and Nenghai Yu.
\newblock Opera: Alleviating hallucination in multi-modal large language models via over-trust penalty and retrospection-allocation.
\newblock In \emph{IEEE Conference on Computer Vision and Pattern Recognition (CVPR)}, 2024.

\bibitem[Leng et~al.(2024)Leng, Zhang, Chen, Li, Lu, Miao, and Bing]{leng2024vcd}
Sicong Leng, Hang Zhang, Guanzheng Chen, Xin Li, Shijian Lu, Chunyan Miao, and Lidong Bing.
\newblock Mitigating object hallucinations in large vision-language models through visual contrastive decoding.
\newblock In \emph{IEEE Conference on Computer Vision and Pattern Recognition (CVPR)}, 2024.

\bibitem[Li et~al.(2023{\natexlab{a}})Li, Holtzman, Fried, Liang, Eisner, Hashimoto, Zettlemoyer, and Lewis]{li2023cd}
Xiang~Lisa Li, Ari Holtzman, Daniel Fried, Percy Liang, Jason Eisner, Tatsunori Hashimoto, Luke Zettlemoyer, and Mike Lewis.
\newblock Contrastive decoding: Open-ended text generation as optimization.
\newblock In \emph{The 61st Annual Meeting Of The Association For Computational Linguistics}, 2023{\natexlab{a}}.

\bibitem[Li et~al.(2023{\natexlab{b}})Li, Du, Zhou, Wang, Zhao, and Wen]{li2023pope}
Yifan Li, Yifan Du, Kun Zhou, Jinpeng Wang, Wayne~Xin Zhao, and Ji-Rong Wen.
\newblock Evaluating object hallucination in large vision-language models.
\newblock In \emph{Proceedings of the 2023 Conference on Empirical Methods in Natural Language Processing}, 2023{\natexlab{b}}.

\bibitem[Lin et~al.(2024)Lin, Chen, Tsai, Jiang, and Yang]{lin2024text}
Yuanze Lin, Yi-Wen Chen, Yi-Hsuan Tsai, Lu Jiang, and Ming-Hsuan Yang.
\newblock Text-driven image editing via learnable regions.
\newblock In \emph{IEEE Conference on Computer Vision and Pattern Recognition (CVPR)}, 2024.

\bibitem[Liu et~al.(2024{\natexlab{a}})Liu, Lin, Li, Wang, Yacoob, and Wang]{liu2024lrvinstruct}
Fuxiao Liu, Kevin Lin, Linjie Li, Jianfeng Wang, Yaser Yacoob, and Lijuan Wang.
\newblock Mitigating hallucination in large multi-modal models via robust instruction tuning.
\newblock In \emph{International Conference on Learning Representations (ICLR)}, 2024{\natexlab{a}}.

\bibitem[Liu et~al.(2024{\natexlab{b}})Liu, Li, Li, and Lee]{llava15}
Haotian Liu, Chunyuan Li, Yuheng Li, and Yong~Jae Lee.
\newblock Improved baselines with visual instruction tuning.
\newblock In \emph{IEEE Conference on Computer Vision and Pattern Recognition (CVPR)}, 2024{\natexlab{b}}.

\bibitem[Ouali et~al.(2024)Ouali, Bulat, Martinez, and Tzimiropoulos]{ouali2024clipdpo}
Yassine Ouali, Adrian Bulat, Brais Martinez, and Georgios Tzimiropoulos.
\newblock Clip-dpo: Vision-language models as a source of preference for fixing hallucinations in lvlms.
\newblock In \emph{European Conference on Computer Vision (ECCV)}, 2024.

\bibitem[Radford et~al.(2021)Radford, Kim, Hallacy, Ramesh, Goh, Agarwal, Sastry, Askell, Mishkin, Clark, et~al.]{radford2021clip}
Alec Radford, Jong~Wook Kim, Chris Hallacy, Aditya Ramesh, Gabriel Goh, Sandhini Agarwal, Girish Sastry, Amanda Askell, Pamela Mishkin, Jack Clark, et~al.
\newblock Learning transferable visual models from natural language supervision.
\newblock In \emph{International Conference on Machine Learning (ICML)}, 2021.

\bibitem[Rafailov et~al.(2024)Rafailov, Sharma, Mitchell, Manning, Ermon, and Finn]{rafailov2024dpo}
Rafael Rafailov, Archit Sharma, Eric Mitchell, Christopher~D Manning, Stefano Ermon, and Chelsea Finn.
\newblock Direct preference optimization: Your language model is secretly a reward model.
\newblock \emph{Advances in Neural Information Processing Systems (NeurIPS)}, 2024.

\bibitem[Sun et~al.(2023)Sun, Shen, Cao, Liu, Li, Shen, Gan, Gui, Wang, Yang, et~al.]{sun2023rlhf}
Zhiqing Sun, Sheng Shen, Shengcao Cao, Haotian Liu, Chunyuan Li, Yikang Shen, Chuang Gan, Liang-Yan Gui, Yu-Xiong Wang, Yiming Yang, et~al.
\newblock Aligning large multimodal models with factually augmented rlhf.
\newblock \emph{arXiv preprint arXiv:2309.14525}, 2023.

\bibitem[Touvron et~al.(2023)Touvron, Lavril, Izacard, Martinet, Lachaux, Lacroix, Rozi{\`e}re, Goyal, Hambro, Azhar, et~al.]{touvron2023llama}
Hugo Touvron, Thibaut Lavril, Gautier Izacard, Xavier Martinet, Marie-Anne Lachaux, Timoth{\'e}e Lacroix, Baptiste Rozi{\`e}re, Naman Goyal, Eric Hambro, Faisal Azhar, et~al.
\newblock Llama: Open and efficient foundation language models.
\newblock \emph{arXiv preprint arXiv:2302.13971}, 2023.

\bibitem[Wang et~al.(2024)Wang, Pan, Ding, and Biemann]{wang2024icd}
Xintong Wang, Jingheng Pan, Liang Ding, and Chris Biemann.
\newblock Mitigating hallucinations in large vision-language models with instruction contrastive decoding.
\newblock \emph{arXiv preprint arXiv:2403.18715}, 2024.

\bibitem[Whitehead et~al.(2024)Whitehead, Phillips, and Hendryx]{whitehead2024pre}
Spencer Whitehead, Jacob Phillips, and Sean Hendryx.
\newblock Pre-training multimodal hallucination detectors with corrupted grounding data.
\newblock \emph{arXiv preprint arXiv:2409.00238}, 2024.

\bibitem[Ye et~al.(2024)Ye, Xu, Ye, Yan, Hu, Liu, Qian, Zhang, and Huang]{ye2024mplug2}
Qinghao Ye, Haiyang Xu, Jiabo Ye, Ming Yan, Anwen Hu, Haowei Liu, Qi Qian, Ji Zhang, and Fei Huang.
\newblock mplug-owl2: Revolutionizing multi-modal large language model with modality collaboration.
\newblock In \emph{IEEE Conference on Computer Vision and Pattern Recognition (CVPR)}, pages 13040--13051, 2024.

\bibitem[Yin et~al.(2023)Yin, Fu, Zhao, Xu, Wang, Sui, Shen, Li, Sun, and Chen]{yin2023woodpecker}
Shukang Yin, Chaoyou Fu, Sirui Zhao, Tong Xu, Hao Wang, Dianbo Sui, Yunhang Shen, Ke Li, Xing Sun, and Enhong Chen.
\newblock Woodpecker: Hallucination correction for multimodal large language models.
\newblock \emph{arXiv preprint arXiv:2310.16045}, 2023.

\bibitem[Zhao et~al.(2023)Zhao, Wang, Ouyang, Dong, Wang, and He]{zhao2023hadpo}
Zhiyuan Zhao, Bin Wang, Linke Ouyang, Xiaoyi Dong, Jiaqi Wang, and Conghui He.
\newblock Beyond hallucinations: Enhancing lvlms through hallucination-aware direct preference optimization.
\newblock \emph{arXiv preprint arXiv:2311.16839}, 2023.

\bibitem[Zheng et~al.(2023)Zheng, Chiang, Sheng, Zhuang, Wu, Zhuang, Lin, Li, Li, Xing, et~al.]{zheng2023vicuna}
Lianmin Zheng, Wei-Lin Chiang, Ying Sheng, Siyuan Zhuang, Zhanghao Wu, Yonghao Zhuang, Zi Lin, Zhuohan Li, Dacheng Li, Eric Xing, et~al.
\newblock Judging llm-as-a-judge with mt-bench and chatbot arena.
\newblock \emph{Advances in Neural Information Processing Systems (NeurIPS)}, 2023.

\bibitem[Zhou et~al.(2024)Zhou, Cui, Yoon, Zhang, Deng, Finn, Bansal, and Yao]{zhou2023lure}
Yiyang Zhou, Chenhang Cui, Jaehong Yoon, Linjun Zhang, Zhun Deng, Chelsea Finn, Mohit Bansal, and Huaxiu Yao.
\newblock Analyzing and mitigating object hallucination in large vision-language models.
\newblock In \emph{International Conference on Learning Representations (ICLR)}, 2024.

\bibitem[Zhu et~al.(2023)Zhu, Chen, Shen, Li, and Elhoseiny]{zhu2023minigpt}
Deyao Zhu, Jun Chen, Xiaoqian Shen, Xiang Li, and Mohamed Elhoseiny.
\newblock Minigpt-4: Enhancing vision-language understanding with advanced large language models.
\newblock \emph{arXiv preprint arXiv:2304.10592}, 2023.

\bibitem[Zhu et~al.(2024)Zhu, Ji, Chen, Xu, Ye, and Liu]{zhu2024ibd}
Lanyun Zhu, Deyi Ji, Tianrun Chen, Peng Xu, Jieping Ye, and Jun Liu.
\newblock Ibd: Alleviating hallucinations in large vision-language models via image-biased decoding.
\newblock \emph{arXiv preprint arXiv:2402.18476}, 2024.

\end{thebibliography}
}

\clearpage
\maketitlesupplementary

\section{Details of Visually Contrastive Decoding}
\subsection{Plausibility Constraints}
Based on the contrastive decoding process from (\textcolor{blue}{2}) in the main paper, it may penalize the entire LVLM's responses via contrast samples. However, either original or visually changed samples may still generate words related to linguistic standards and common sense. Therefore, inaccurate punishment of these words may cause an unpredictable and even failed response. To tackle this issue, we follow previous works~\cite{li2023cd, leng2024vcd} to introduce the plausibility constraint, which avoids the indiscriminate penalization caused by directly applying contrastive decoding to entire responses. The plausibility constraint is triggered by the confidence related to candidate tokens of original visual inputs, which can be formulated as:
\begin{equation}
\label{equ:plausibility_constraint}
% \large
\begin{aligned}
&V_{cond}(y_{<t}) = \{y_t \in V: \\
&p_\theta(y_t|v, x, y_{<t}) \geq \beta~\underset{\omega}{max}~p_\theta(\omega|v, x, y_{<t})\},  \\
&p_{cd}(y_t|v, v_{cd}, x, y_{<t}) = 0 ~~\text{if}~~ y_t \notin V_{cond}(y_{<t}),
\end{aligned}
\end{equation}
where $V_{cond}$ is the output of LVLMs and $\beta$ in [0, 1] is a hyperparameter that controls how aggressive to keep only high-probability tokens (e.g., a higher value indicates a more aggressive manner). Therefore, the probability of other candidate tokens below the constraints is replaced with 0. With the plausibility constraint, contrastive decoding influences the next token predicting behaviors only when the LVLMs are not confident with their prediction. We set $\beta=0.2$ by default.

% \subsection{General Case of Image Editing}
% \alan{plan not to add due to distraction from our analysis}
% As introduced in the main paper, we leverage the queried objects or contents from the questions (which is the input textual query $x$ for LVLMs) as the editing textual instruction $t_{edit}$ for the question-answering tasks. However, in a general case where LVLMs are required to generate detailed responses, such as image captioning, it could be hard to get the editing textual instruction from the input textual query (e.g., ``please describe the image in detail'' for image captioning). To this end, we propose to leverage the responses prone to hallucination as the editing textual query to generate hallucination-targeted images for contrastive decoding.
% We first generate the queried responses and detect the hallucination sentences according to the CLIP Scores. We select sentences with the top-k lowest CLIP Scores as the editing textual instructions and thus generate edited images with these victim instructions (responses) of hallucinations. Then, we regard these edited images as contrastive samples to perform contrastive decoding to regenerate responses.

\section{More Analysis on LVLMs}
We provide more analysis of visually changed samples on LVLMs, including more LVLMs (e.g., Qwen-VL~\cite{bai2023qwen}), more metrics (e.g., confidence of predictions), and details of POPE benchmarks (e.g., random, popular tasks). In this section, we mainly investigate the behaviors of visually changed samples from the perspective of different LVLMs, in which the conclusions are drawn similarly to the main paper.
% Based on investigations from the perspective of different LVLMs and benchmarks, we draw several observations:

% \indent 1) The behavior of visually changed samples is different across LVLMs and benchmarks. Within the same benchmark (i.e., POPE~\cite{li2023pope}, which focuses on object hallucination and has three tasks of different difficulties), the distributions of entropy, confidence, and probability distribution distance (PDD) are similar. 

% \indent 2)  With similar concepts of changing images (i.e., reducing visual information), the downsample and diffusion noise methods have similar behaviors regarding entropy, confidence, and probability distribution distance (PDD) on the LLaVA-1.5 and Qwen-VL while obtaining distinct behaviors on InstructBLIP. The reason could be the different architecture of visual-textual interfaces in LVLMs, where InstructBLIP leverages Q-Former to extract visual features while LLaVA-1.5 and Qwen-VL directly project the image embedding to text embedding via multi-layer perception (MLP) and cross-attention layer, respectively. 

% \indent 3) 

\subsection{Analysis with Entropy and Confidence}
\begin{figure*}[!t]
    \centering
    % \LARGE
    \resizebox{0.95\linewidth}!{
        \begin{tabular}{c}
        % \multirow{3}{*}{(a)} 
        \begin{tabular}{c}\includegraphics[width=\textwidth]{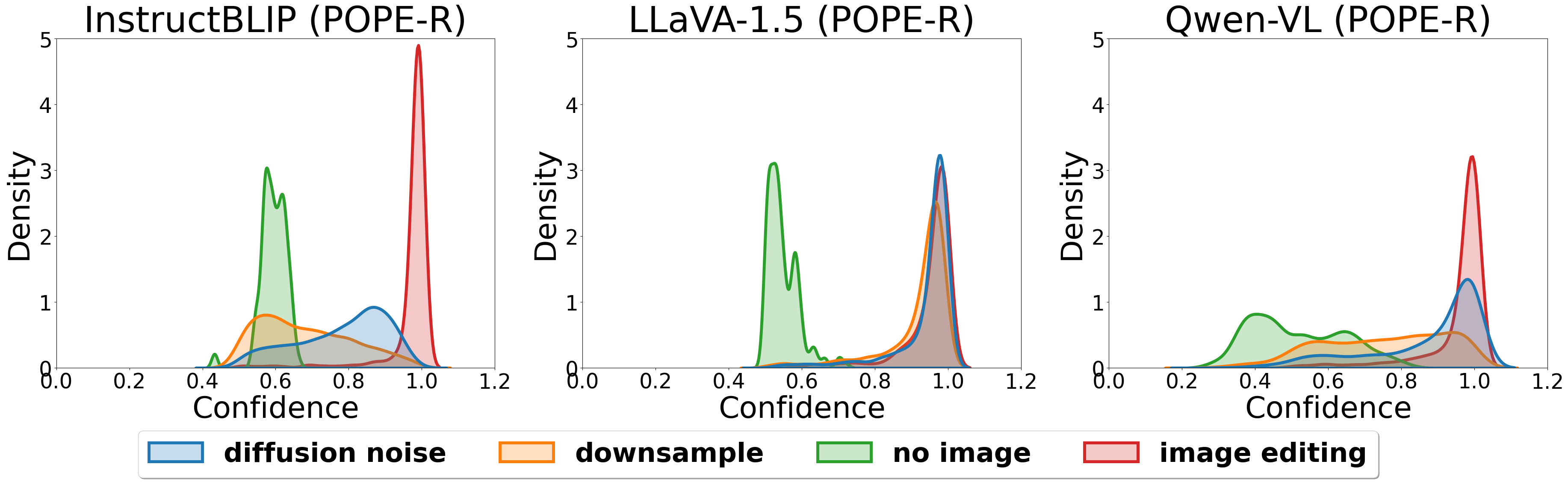}\end{tabular}\\
        (a) POPE \textit{random}\\ \\
        \begin{tabular}{c}\includegraphics[width=\textwidth]{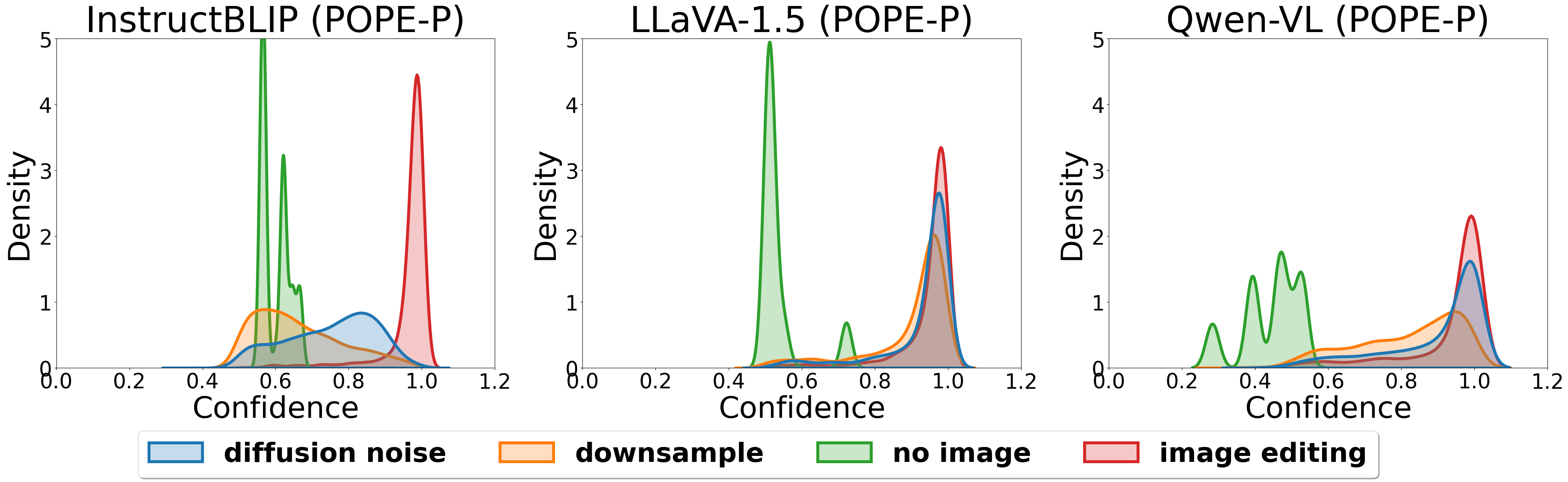}\end{tabular}
        \\
        (b) POPE \textit{popular}\\ \\
        \begin{tabular}{c}\includegraphics[width=\textwidth]{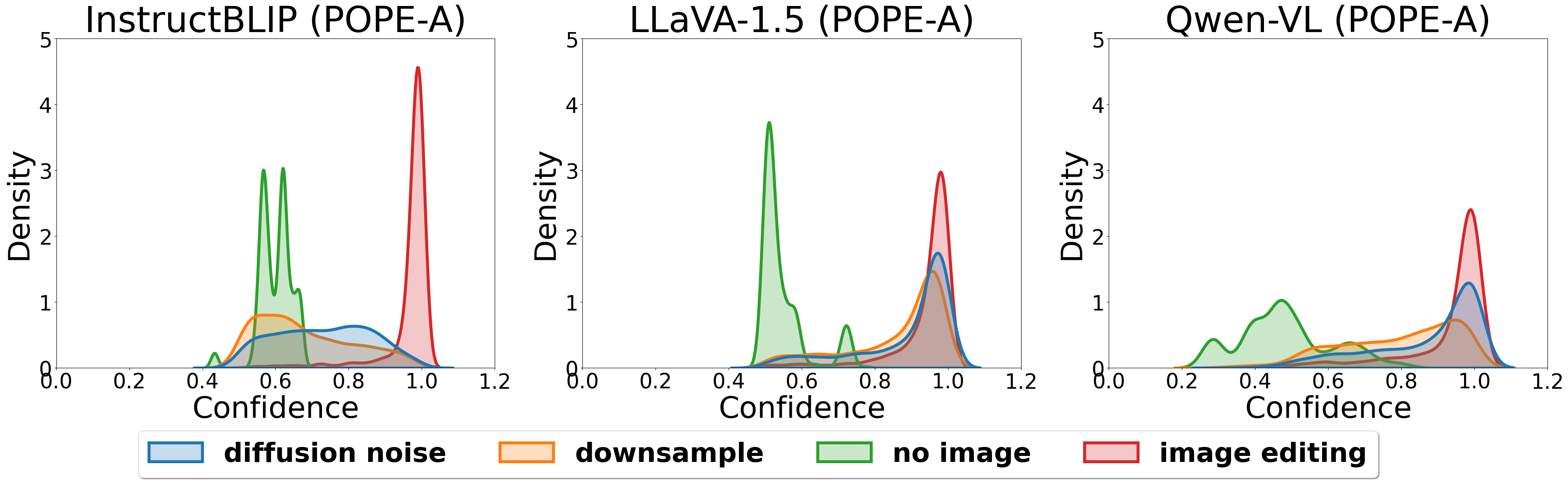}\end{tabular}
        \\
        (c) POPE \textit{adversarial}\\ \\
        \begin{tabular}{c}\includegraphics[width=\textwidth]{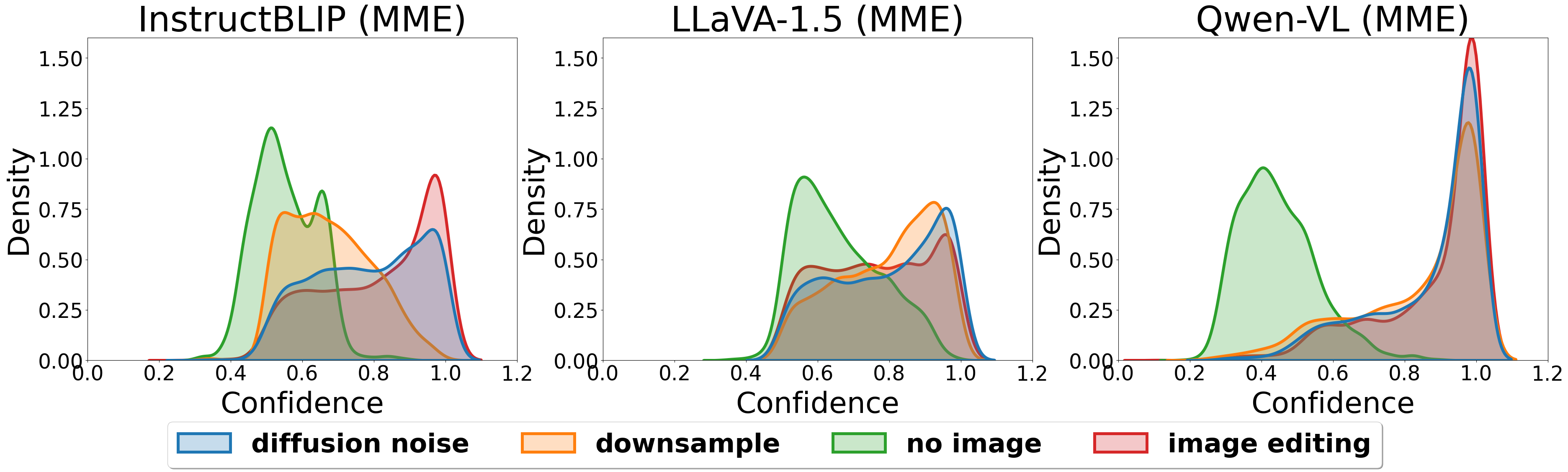}\end{tabular}
        \\
        (d) MME     
        \end{tabular}
    }
    \caption{The analysis of confidence on the POPE and MME benchmarks.
    }
    \label{fig:confidence}
\end{figure*}

We have discussed the investigation of entropy with InstructBLIP~\cite{instructblip} and LLaVA-1.5~\cite{llava15} in the main paper, which presents the different behaviors of visually changed samples. Here we provide the comparison of entropy with another LVLM, Qwen-VL, as shown in Figure~\ref{fig:entropy}. We also visualize the confidence distribution in Figure~\ref{fig:confidence}, where the confidence is estimated from the highest probability of the next-token prediction. The confidence is correlated with the entropy, where the lower confidence induces the higher entropy. We draw several observations:

\indent 1) Within the same benchmark (i.e., POPE~\cite{li2023pope}, which focuses on object hallucination and has tasks of three difficulties), the distributions of entropy and confidence are similar across different LVLMs. 
% However, the behavior of visually changed samples is different across LVLMs and benchmarks.

\indent 2)  With similar concepts of changing images (i.e., reducing visual information), the downsample and diffusion noise methods have similar behaviors regarding entropy and confidence on the LLaVA-1.5 and Qwen-VL, while obtaining distinct behaviors on InstructBLIP. The reason could be the different architecture of visual-textual interfaces in LVLMs, where InstructBLIP leverages Q-Former to extract visual features while LLaVA-1.5 and Qwen-VL directly project the image embedding to text embedding via multi-layer perception (MLP) and cross-attention layer, respectively. 

Based on the above observations, we argue that the behaviors of visually changed samples vary across LVLMs and benchmarks. 

\subsection{Analysis with Probability Distribution Distance}
We investigate the probability distribution distance (PDD) of visually changed samples across different POPE tasks and LVLMs, which present the similarity of output predictions to original ones, as shown in Figure~\ref{fig:probdist}. We find that different visually changed samples have varying PDD distributions across different LVLMs: 

\indent 1) For InstructBLIP, the influence of each visually changed sample is distinct, where the prediction of image editing and no image methods are centralized at high PDD value, while downsample and diffusion noise methods have the opposite distributions.

\indent 2) For LLaVA-1.5, the downsample and diffusion noise methods have similar distributions centralized at lower PDD values, while the image editing method has a distribution with higher PDD values. Different from InstructBLIP, the no image method has a much lower PDD value, which could result from the difference in answering tendency (as shown in Figure~\ref{fig:pope_behaviors}).

\indent 3) However, for the Qwen-VL, the PDD distributions of each visually changed sample are much similar and smoother, meaning that the responses produced by visually changed samples are unpredictable. That is, we cannot determine the impact of downsample or diffusion noise methods given arbitrary visually changed inputs.

These observations show that LVLMs react with visually changed samples in different ways, which can also have varying impacts on the outcomes of contrastive decoding.

\subsection{Revision Behaviors}
\begin{figure*}[!t]
    \centering
    \Huge
    \resizebox{0.95\linewidth}!{
        \begin{tabular}{c}
        % \multirow{3}{*}{(a)} 
        \begin{tabular}{ccc}
        \includegraphics[width=\textwidth]{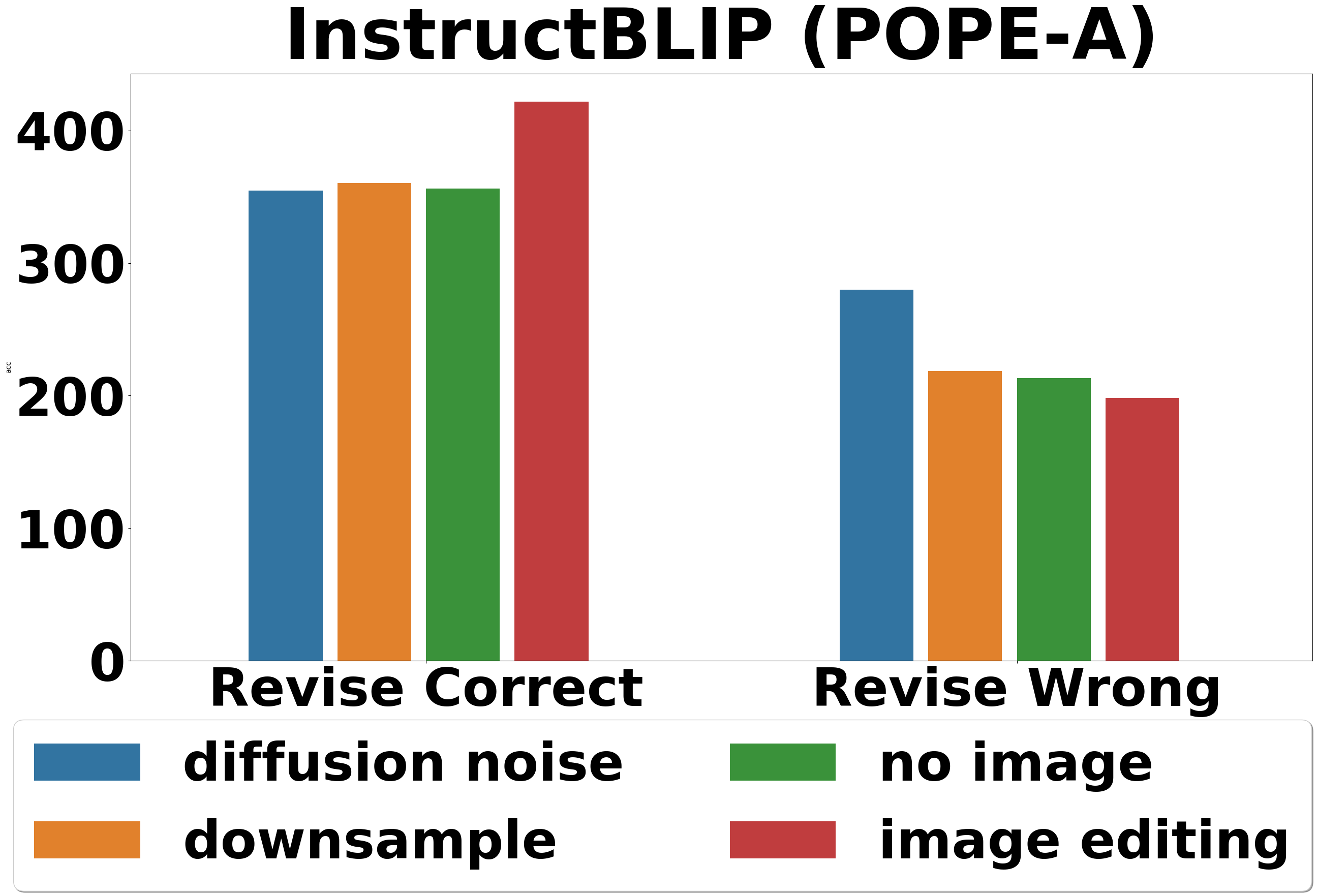} 
        \includegraphics[width=\textwidth]{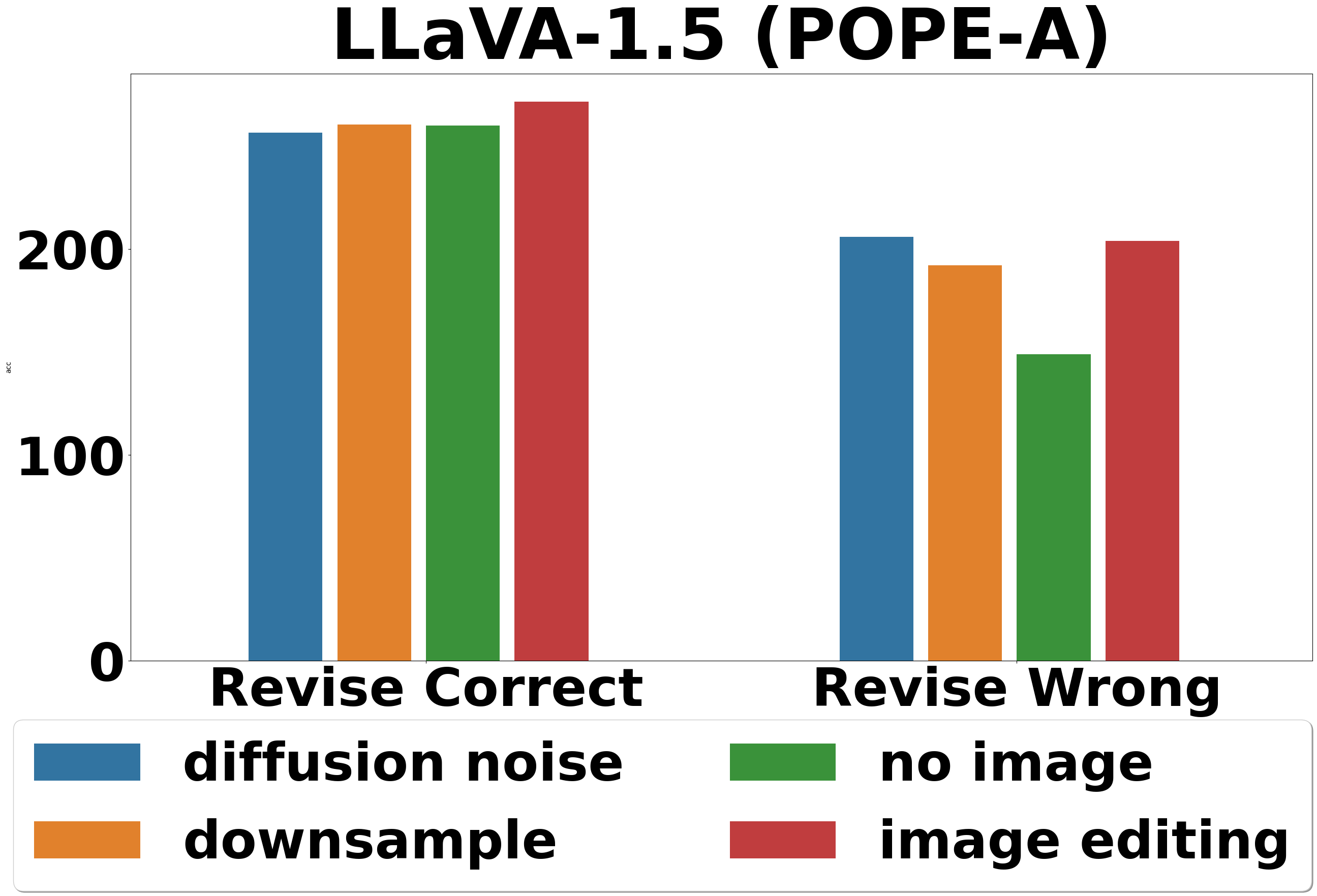} 
        \includegraphics[width=\textwidth]{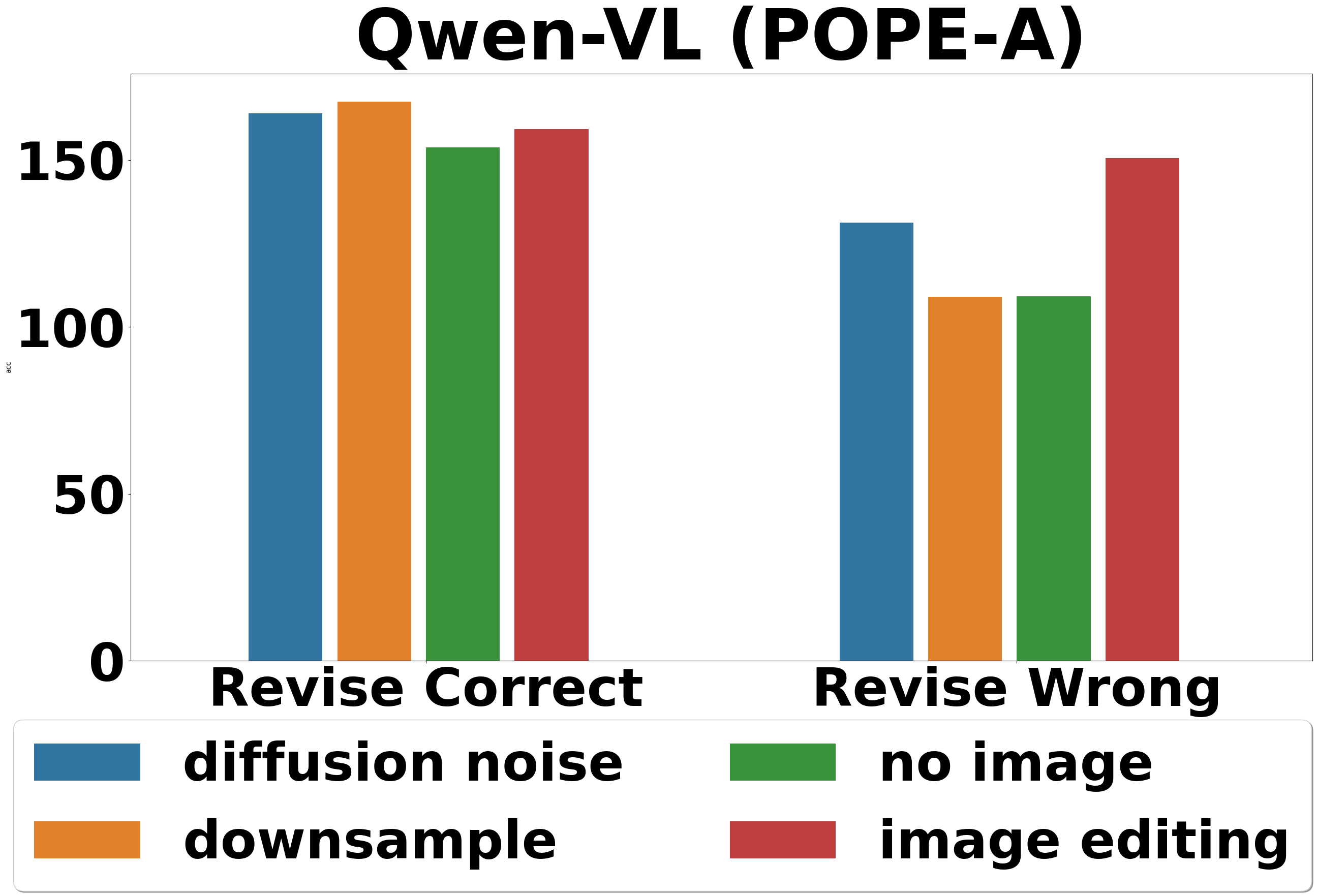} 
        \end{tabular}\\
        (a) Revision behaviors on POPE  \\ \\
        \begin{tabular}{ccc}
        \includegraphics[width=\textwidth]{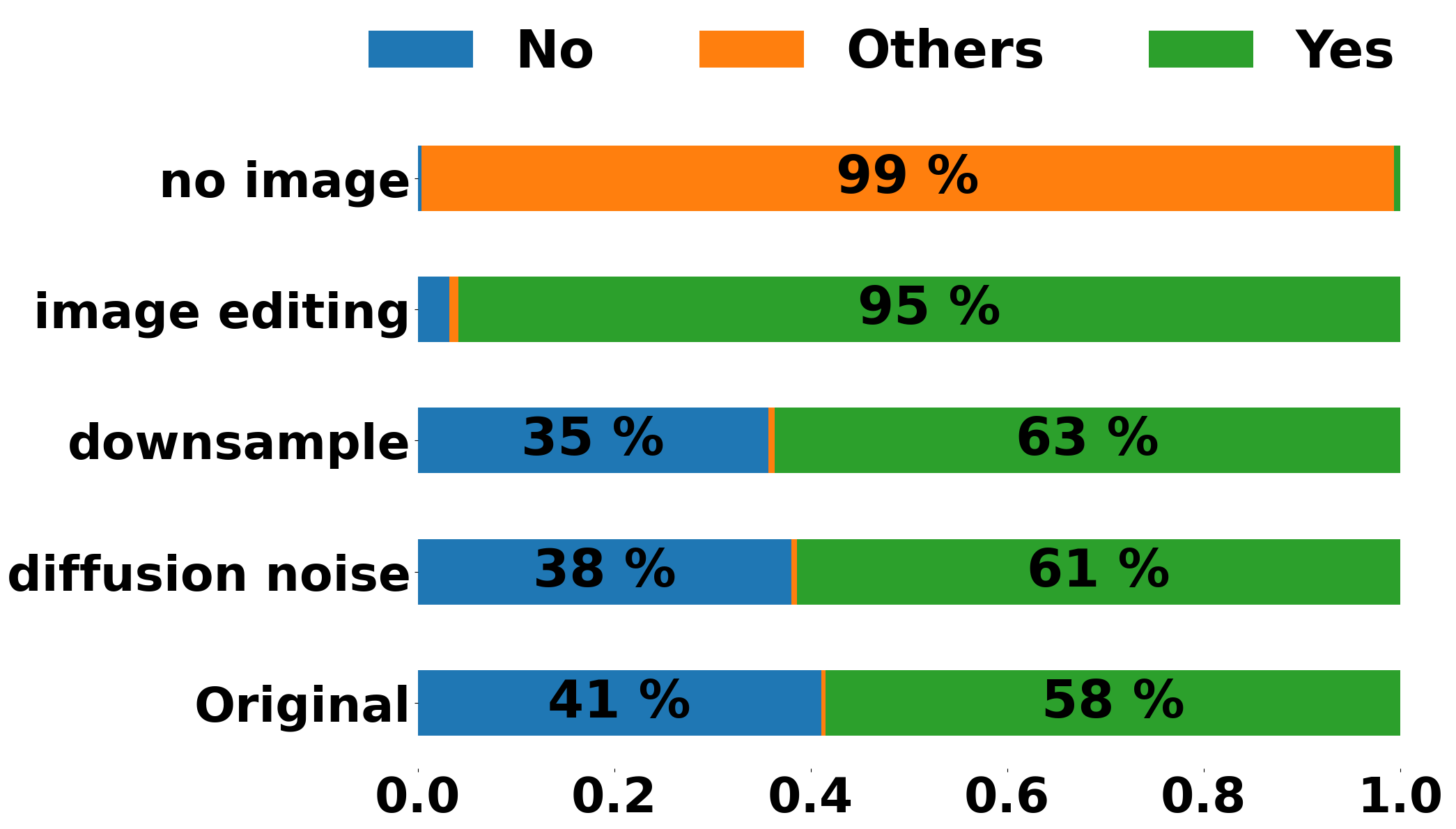}
        \includegraphics[width=\textwidth]{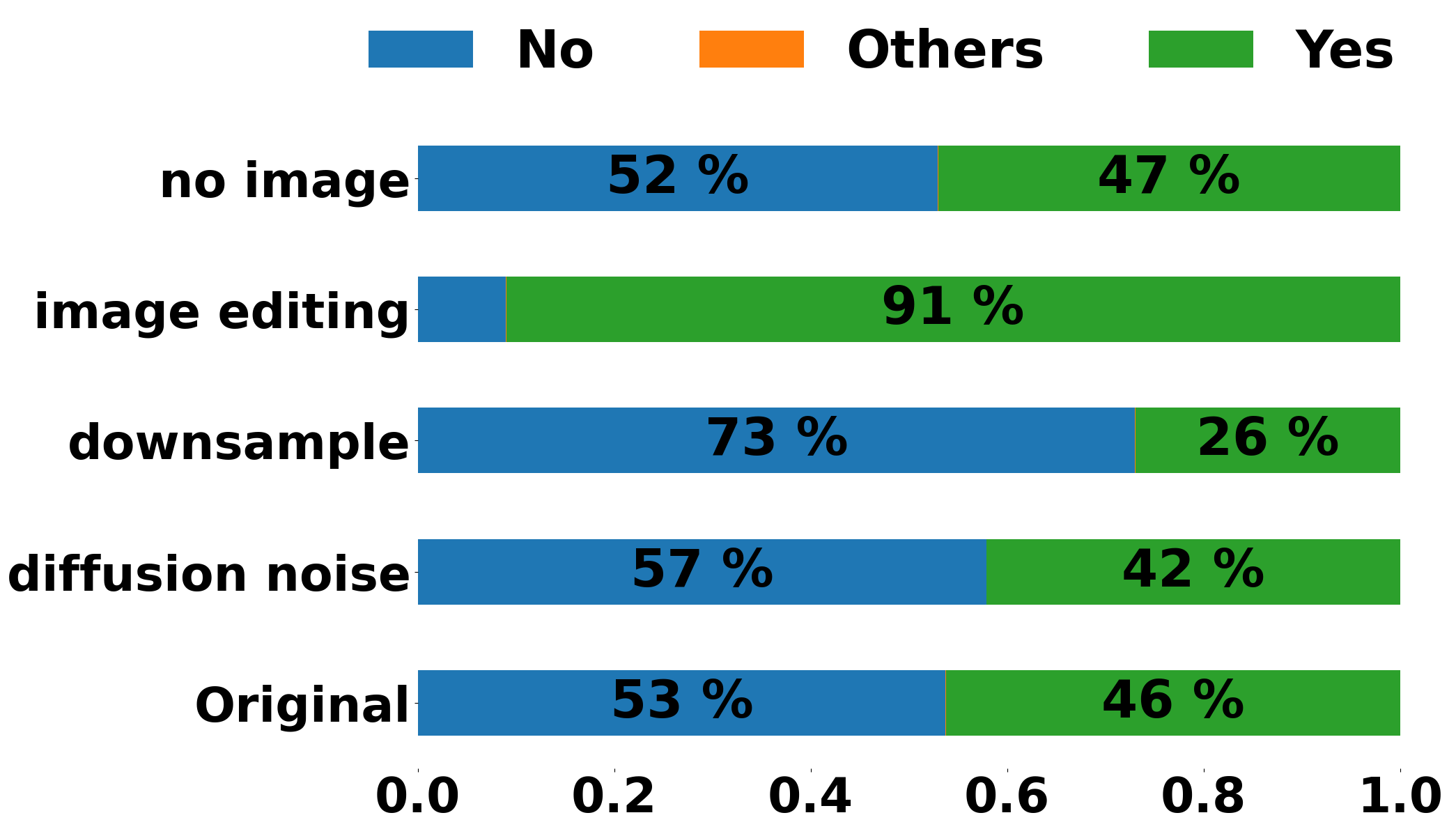}
        \includegraphics[width=\textwidth]{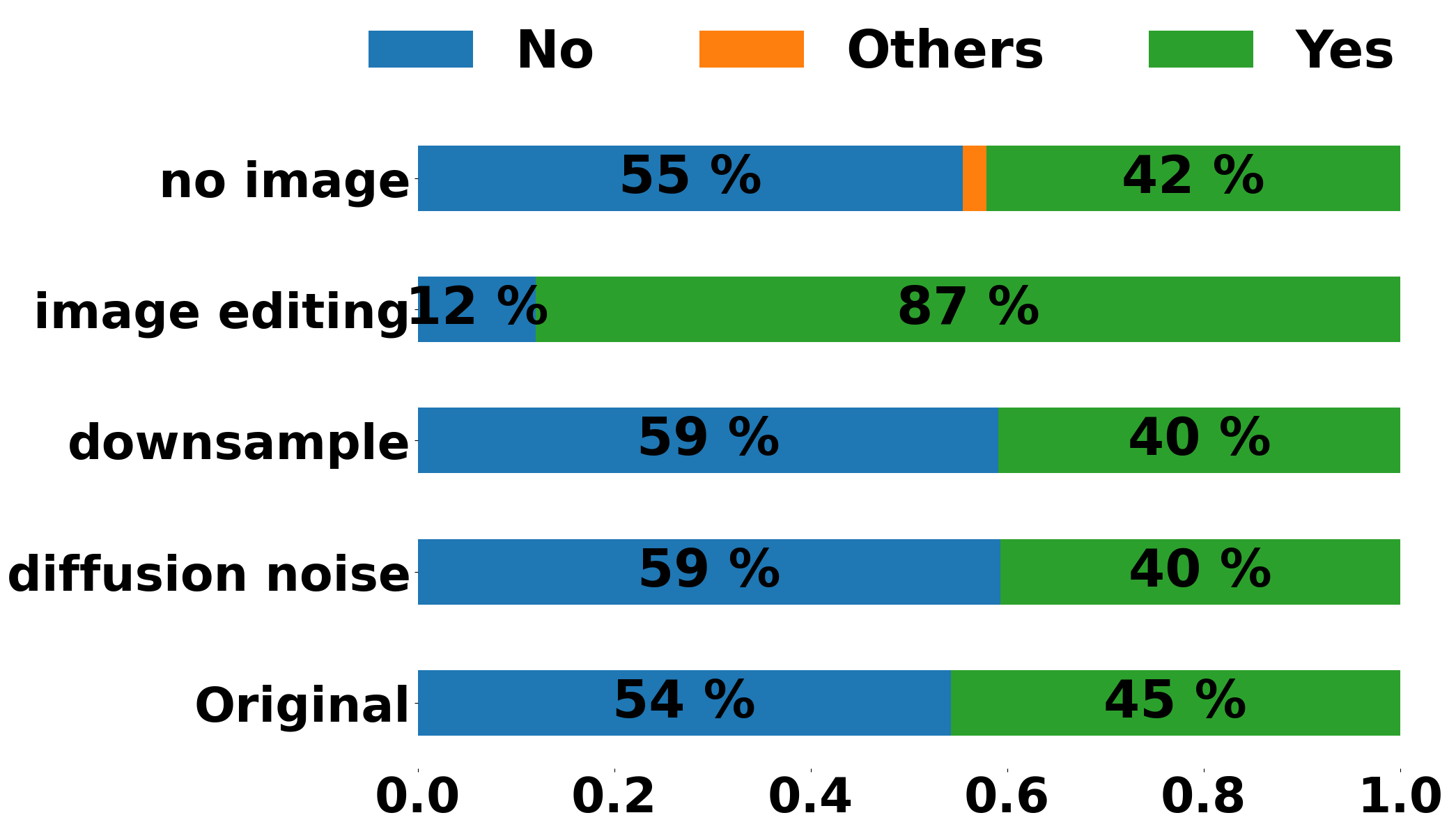}\end{tabular}
        \\
        (b) Answering Tendency on POPE     
        \end{tabular}
    }
    \caption{The revision behaviors and answering tendency on POPE benchmark (adversarial task) with different visually changed samples and LVLMs.
    }
    \label{fig:pope_behaviors}
\end{figure*}

\begin{figure*}[!t]
    \centering
    \Huge
    \resizebox{0.95\linewidth}!{
        \begin{tabular}{c}
        % \multirow{3}{*}{(a)} 
        \begin{tabular}{ccc}
        \includegraphics[width=\textwidth]{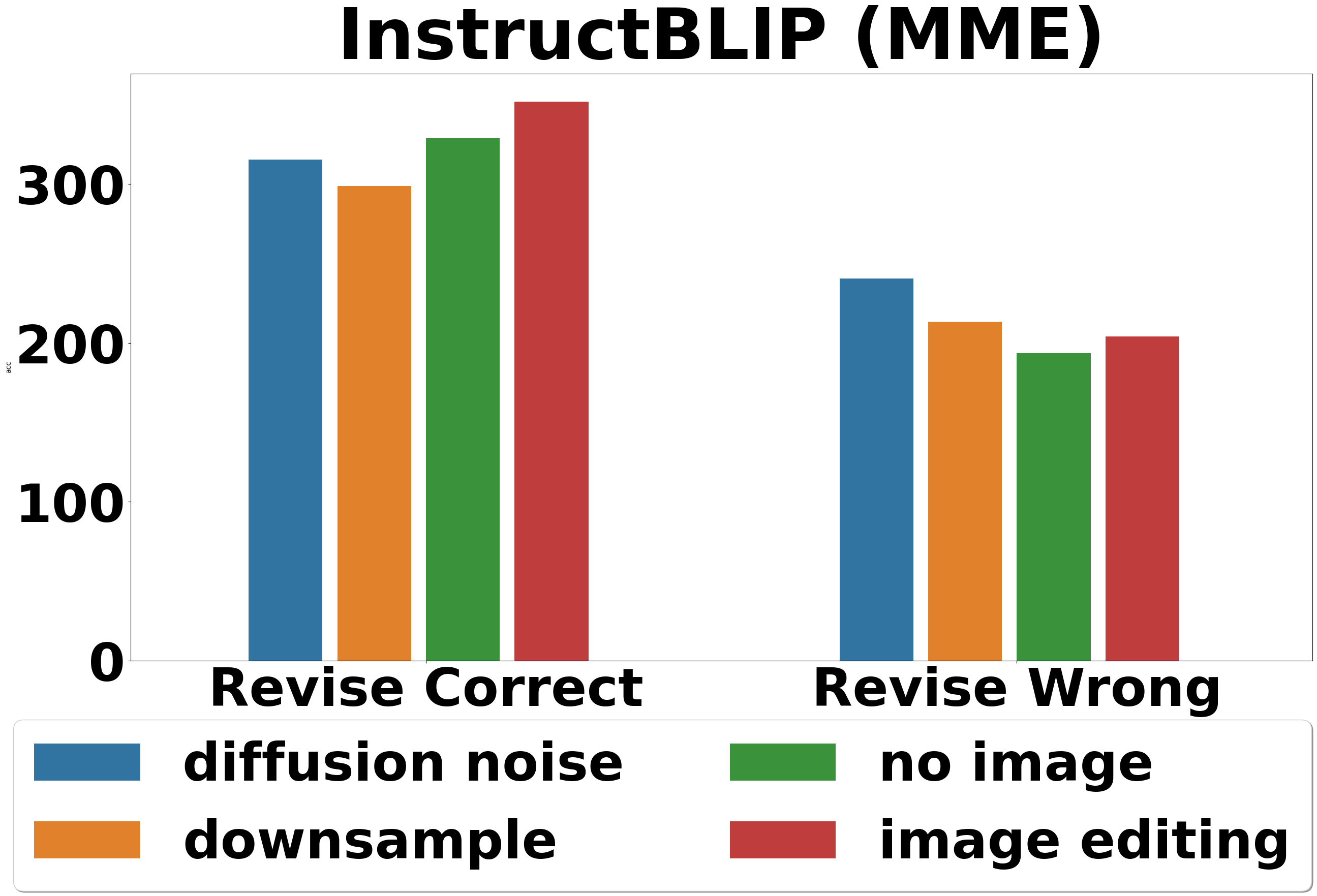} 
        \includegraphics[width=\textwidth]{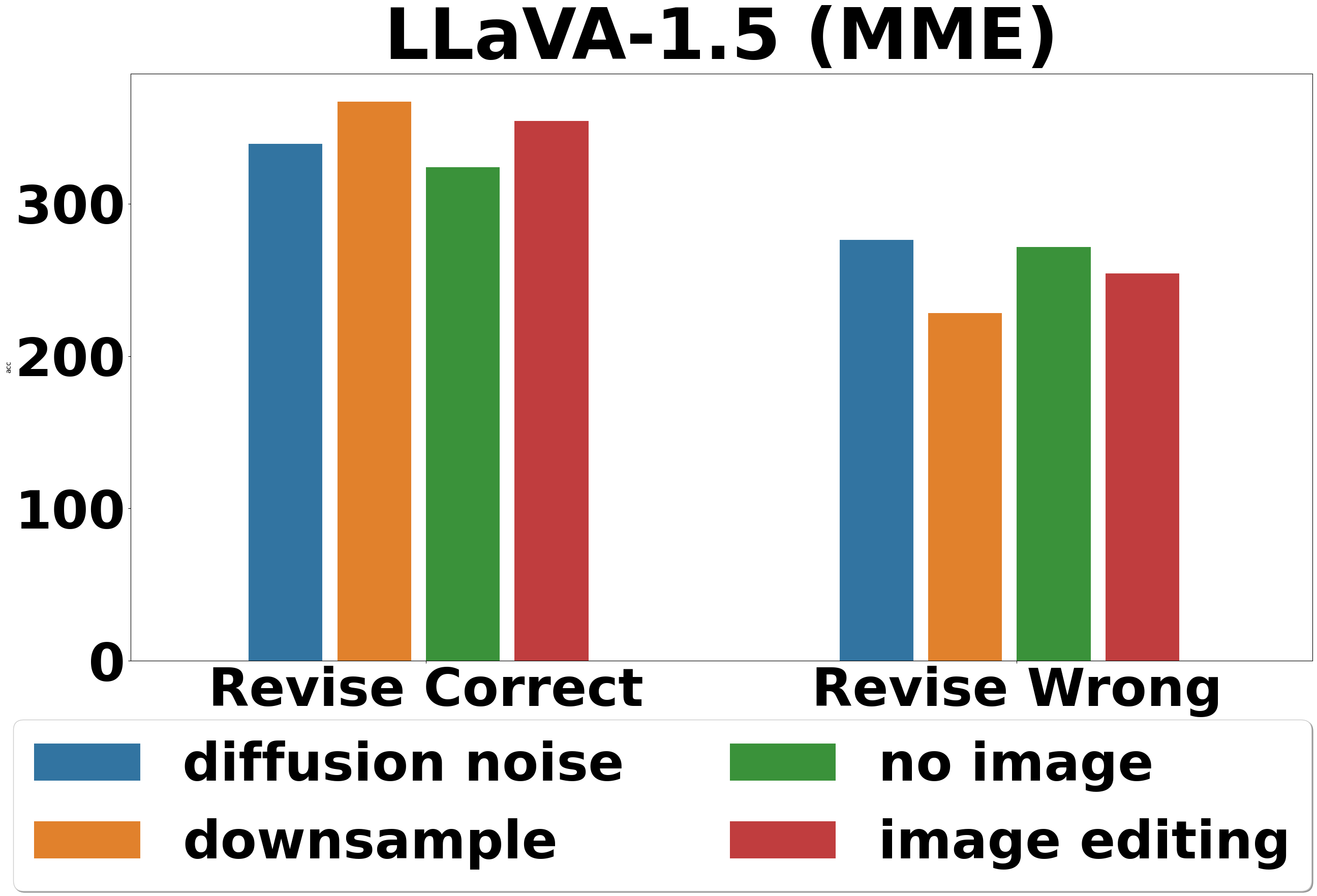} 
        \includegraphics[width=\textwidth]{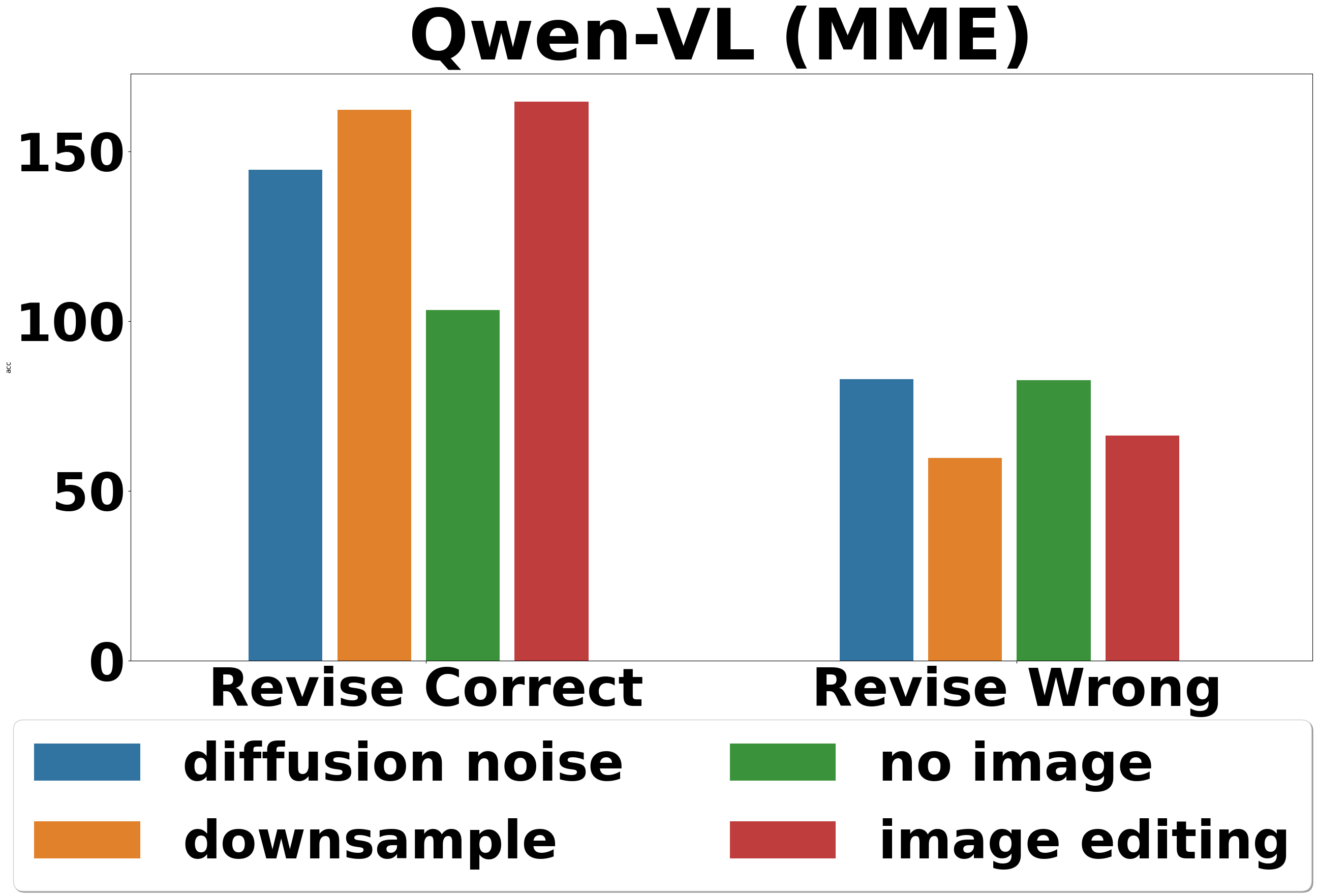} 
        \end{tabular}\\
        (a) Revision behaviors on MME  \\ \\
        \begin{tabular}{ccc}
        \includegraphics[width=\textwidth]{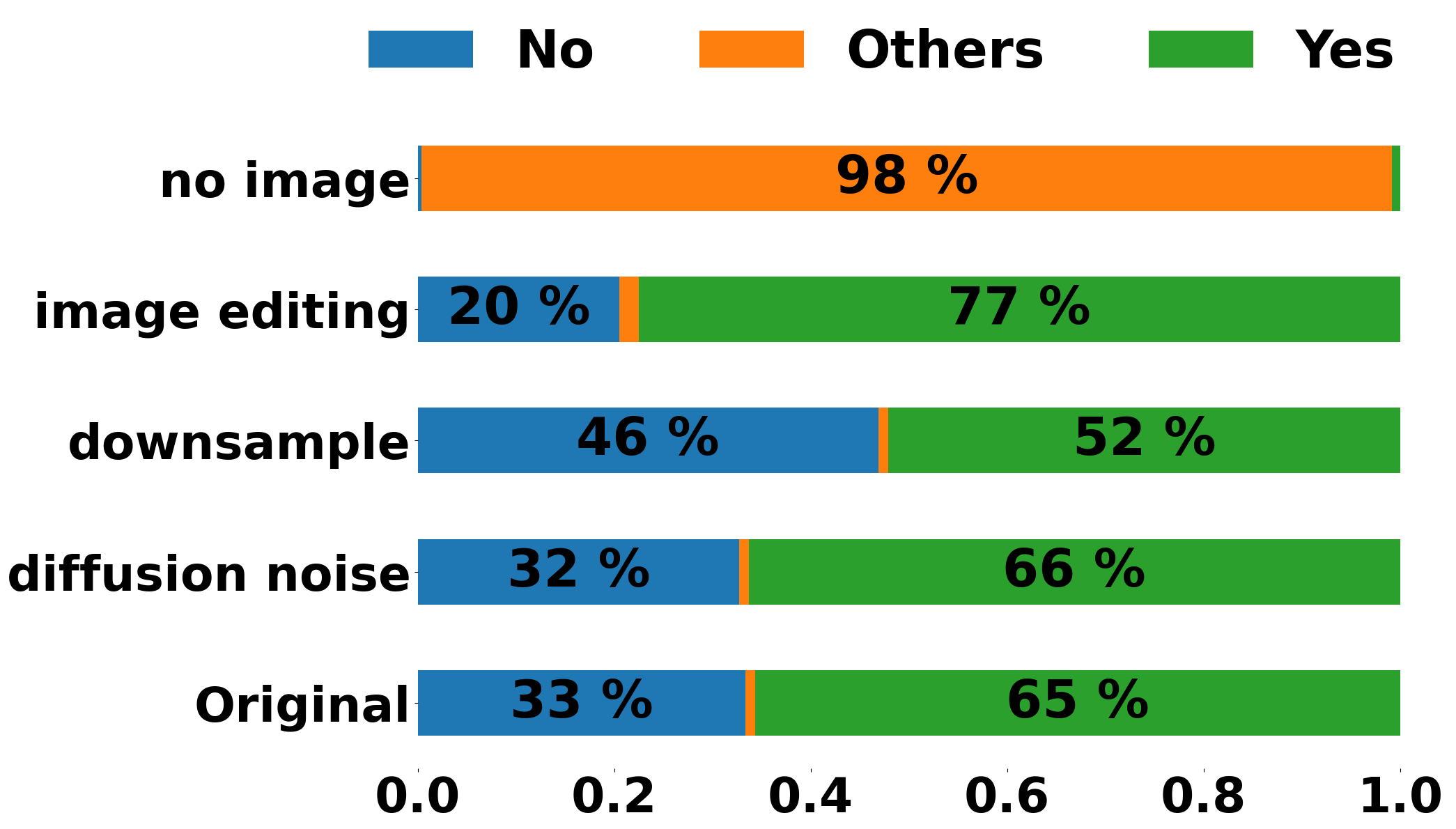}
        \includegraphics[width=\textwidth]{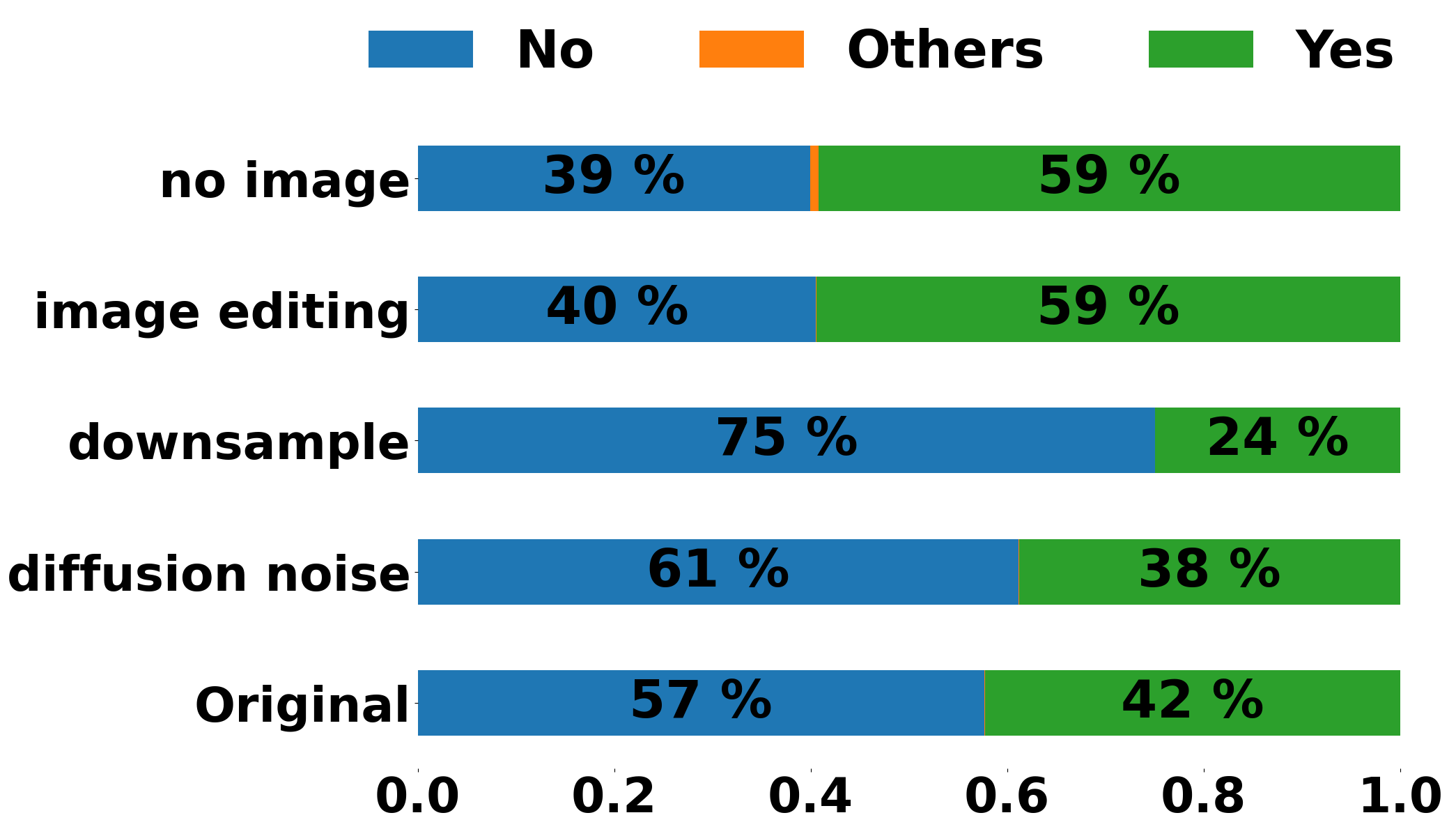}
        \includegraphics[width=\textwidth]{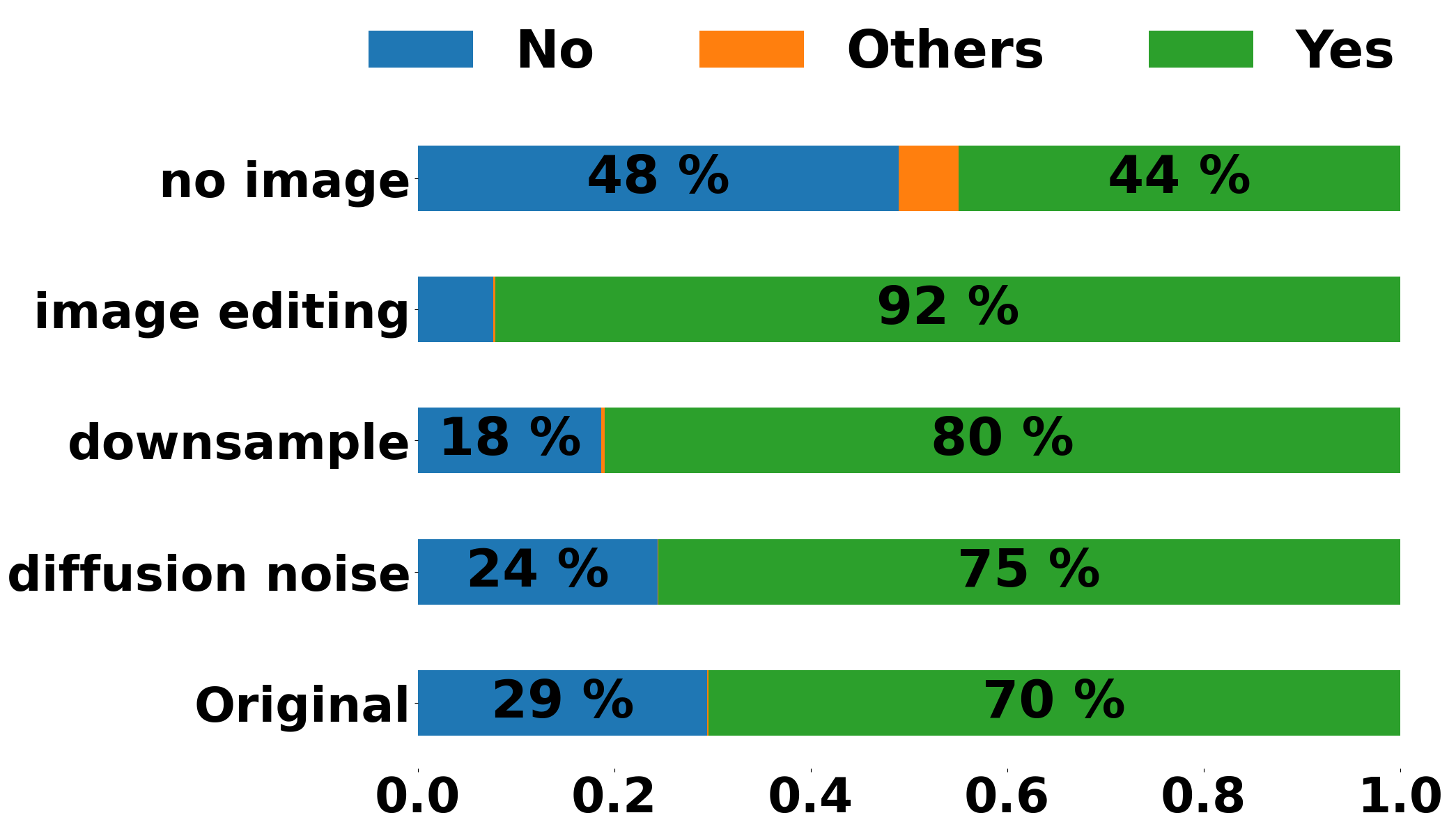}\end{tabular}
        \\
        (b) Answering Tendency on MME     
        \end{tabular}
    }
    \caption{The revision behaviors and answering tendency on MME benchmark with different visually changed samples and LVLMs.
    }
    \label{fig:mme_behaviors}
\end{figure*}

We provide the statistics of revision behaviors and answering tendencies on different benchmarks, as shown in Figure~\ref{fig:pope_behaviors} and \ref{fig:mme_behaviors}. We find that the answering tendencies of visually changed samples on LVLMs are very different. 
The InstructBLIP tends to answer ``yes'' while LLaVA-1.5 is prone to answering ``no'' for the visually changed samples. Moreover, we also find that no image method obtains valid answers (``yes'' or ``no'') on LLaVA-1.5 and Qwen-VL, which differs from the behavior on InstructBLIP. 
This evidence demonstrates the answering tendency varies across different LVLMs, resulting in different behaviors of contrastive decoding with visually changed samples. Such a conclusion can also be obtained from the observation of revision behaviors, which shows that different visually changed samples present different trends in revising answers of different LVLMs (e.g., image editing on POPE-A achieves the least revise-wrong on InstructBLIP but the highest one on Qwen-VL).

\subsection{Pairwise Overlap of Rectified Answers}
We visualize the pairwise overlap of rectified answers of LVLMs in Figure~\ref{fig:overlapping}, and draw several observations:

\indent 1)  For InstructBLIP and LLaVA-1.5, all pairs of CDs have low IoUs, demonstrating the rectified samples resulting from visually changed samples are different and non-overlapping.
Moreover, the IoU of rectified samples from applying downsample and diffusion noise methods is rather higher than other pairs, indicating a similar type of changes on images could lead to similar outcomes.

\indent 2) For Qwen-VL, the results are different from those on the other two LVLMs. On the POPE benchmark, the overlap between downsample and diffusion noise is higher. Meanwhile, on the MME benchmark, the overlaps among downsample, diffusion noise, and no image methods exhibit significantly higher IoUs. On the contrary, the image editing method shows notably low IoUs with the other methods. The evidence suggests that the reduction of visual information may have a similar influence on predictions of Qwen-VL, whereas  the editing of contents impacts its predictions differently.

From these observations, we conclude that for the different LVLMs, the behaviors of visually changed samples could vary, encouraging us to develop a fusion method to leverage the strengths of different visually changed samples to mitigate hallucinations.

% - Add Qwen-VL (mentioned in main paper)
% - if possible, other LVLMs (MiniGPT-4, MPLG-Owl-2)

\section{More Ablation Studies}
To validate our designs for the fusion method of visually changed samples, we conduct several ablation studies, including the combination of fusion, the choice of fusion weights, and the influence of the extent of visual changes.

\subsection{Combinations of Fusion}
We provide results of exhaustive combinations of entropy-weighted fusion methods ranging from involving two to four visually changed samples, as shown in Table~\ref{table:combinations_supp}. In general, combining more visually changed samples obtains better overall performance.
With careful selection, the combination of two visually changed samples (e.g., no image (B) + downsampling (C)) can reach a comparable performance with the combination of all samples. However, it could be difficult to accurately select the optimal combinations when deploying on real-world AI agents. In contrast, our fusion method leveraging all the strengths of different visually changed samples serves as a more general and effective solution to mitigate hallucinations of LVLMs.

\subsection{Ablation of Metric-Weighted Fusions}
\label{sec:fusion}
As mentioned in Sec.~\textcolor{blue}{5.2} of the main paper, when the entropy is higher, the LVLM is more likely to yield the hallucinated response, which is more suitable for serving as a contrastive sample. Here, we investigate using different probability-level metrics as a weight for the fusion of visually changed samples.
To be more general, we reformulate Eq. (\textcolor{blue}{8}) of the main paper as:
\begin{equation}
\label{equ:entropy_fuse_supp}
\begin{aligned}
&p_{cd}(y_t|v, v_{cd}, x, y_{<t})=\text{softmax}[logit_\theta(y_t|v,x,y_{<t})\\
&-\sum_{i=1}^{k}w_i(logit_\theta(y_t|v,x,y_{<t})-logit_\theta(y_t|v_{cd_i},x,y_{<t}))],
\end{aligned}
\end{equation}
where $w_i$ is the weight of output distribution for a visually changed sample $v_{cd_i}$, which can be metrics we have discussed before, including entropy $e_i$, confidence $c_i$, unconfidence $u_i =\frac{1}{c_i}$, and probability distribution distance $d_i$.

The results are shown in Table~\ref{table:weighted}. We find that the entropy-weighted fusion reaches the best performance on overall accuracy, verifying our motivation to combine visually changed samples with entropy as weight. In addition, the unconfidence-weighted (denoted as unconf-weighted) fusion also reaches the best performance, which has a higher performance on MME and a lower performance on POPE, demonstrating that different fusion methods have varying preferences for weighting visually changed samples.
% , resulting in differing performance gains across various benchmarks.

\subsection{Ablation of Extent of Visual Changes}
We investigate the influence of the distortion extent of visually changed samples on different benchmarks from the perspective of performance and entropy. Concretely, the diffusion noise method increases the uncertainty along with the increasing value of noise step $N$ of the diffusion process. The downsample method reduces more detailed high-frequency information as the downsample ratio $r$ increases. For the image editing method, the weight of text guidance $cfg_{text}$ controls how the edited images relate to the editing textual instruction. As the $cfg_{text}$ gets higher, the editing results are more likely consistent with the editing textual instruction and even changing the image contents with the queried objects or contents without preserving the similarity to the original image.

\begin{figure*}[!t]
    \centering
    \LARGE
    \resizebox{\linewidth}!{
        \begin{tabular}{cc}
        % \multirow{3}{*}{(a)} 
            \begin{tabular}{c|c|c|c|c}
             \shortstack{downsample \\ ratio $r$}& POPE-R & POPE-P & POPE-A & MME \\ \hline
            $r=32$ & \textbf{0.879}  & 0.817  &  0.788 & 0.713 \\
            $r=16$ & 0.878  & \textbf{0.824}  &  \textbf{0.801} & \textbf{0.724} \\
            $r=8$ & 0.866  & 0.815  &  0.798 & 0.720 \\
            $r=4$ & 0.858  & 0.812  &  0.789 & 0.720 \\
            
            \end{tabular}
        & \begin{tabular}{c}\includegraphics[width=\textwidth]{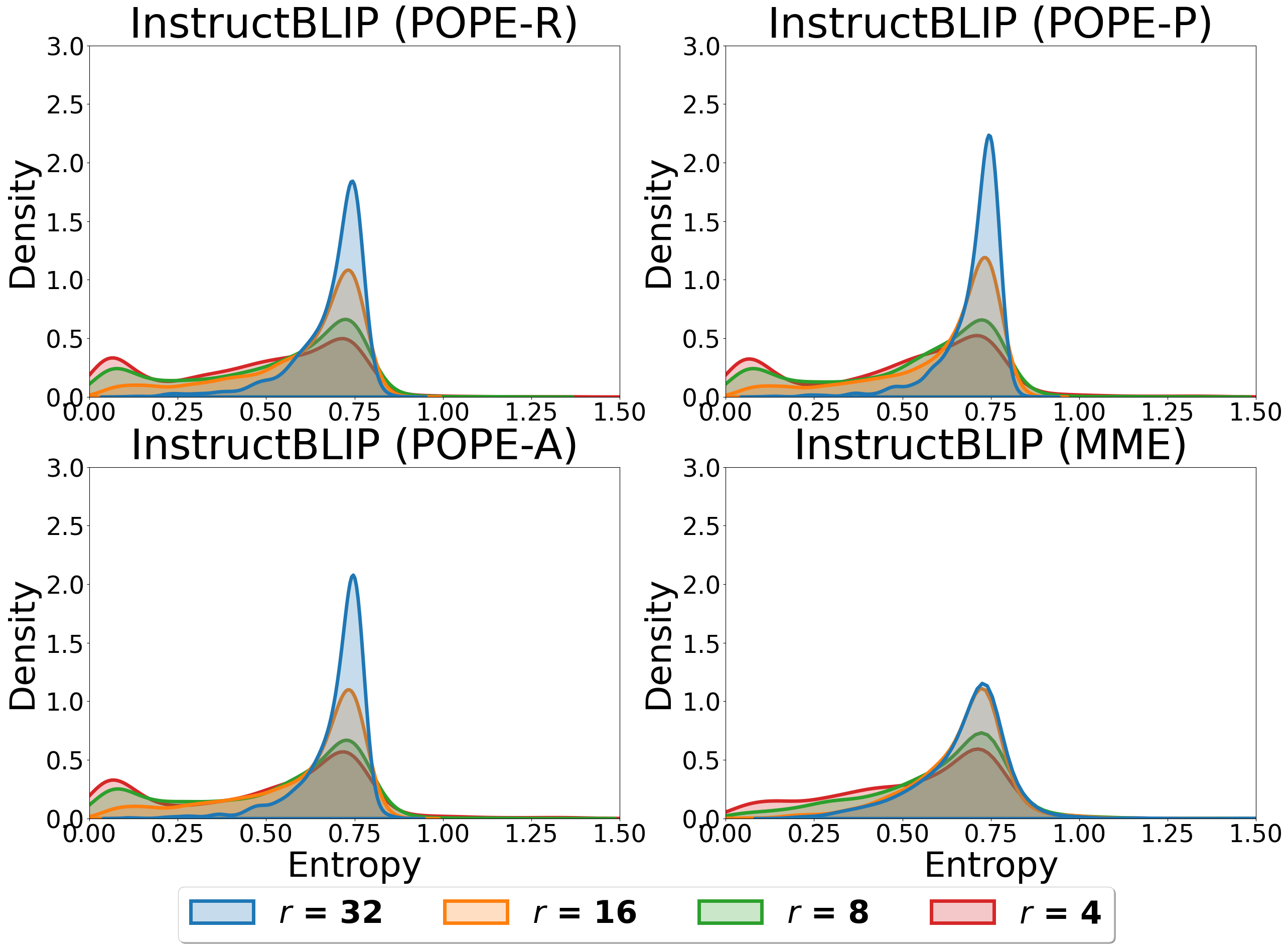}\end{tabular}
        \\
        \end{tabular}
    }
    \caption{The ablation study of different downsample ratios $r$ for the downsample method.
    }
    \label{fig:abl_downsample}
\end{figure*}

\begin{figure*}[!t]
    \centering
    \LARGE
    \resizebox{\linewidth}!{
        \begin{tabular}{cc}
        % \multirow{3}{*}{(a)} 
            \begin{tabular}{c|c|c|c|c}
             \shortstack{noise \\ step $N$}& POPE-R & POPE-P & POPE-A & MME \\ \hline
            $N=750$ & \textbf{0.863}  & \textbf{0.812}  &  \textbf{0.795} & \textbf{0.727} \\
            $N=500$ & 0.858  & 0.805  &  0.779 & 0.720 \\
            $N=250$ & 0.858  & 0.805  &  0.775 & 0.714 \\
            $N=100$ & 0.857  & 0.804  &  0.771 & 0.713 \\
            
            \end{tabular}
        & \begin{tabular}{c}\includegraphics[width=\textwidth]{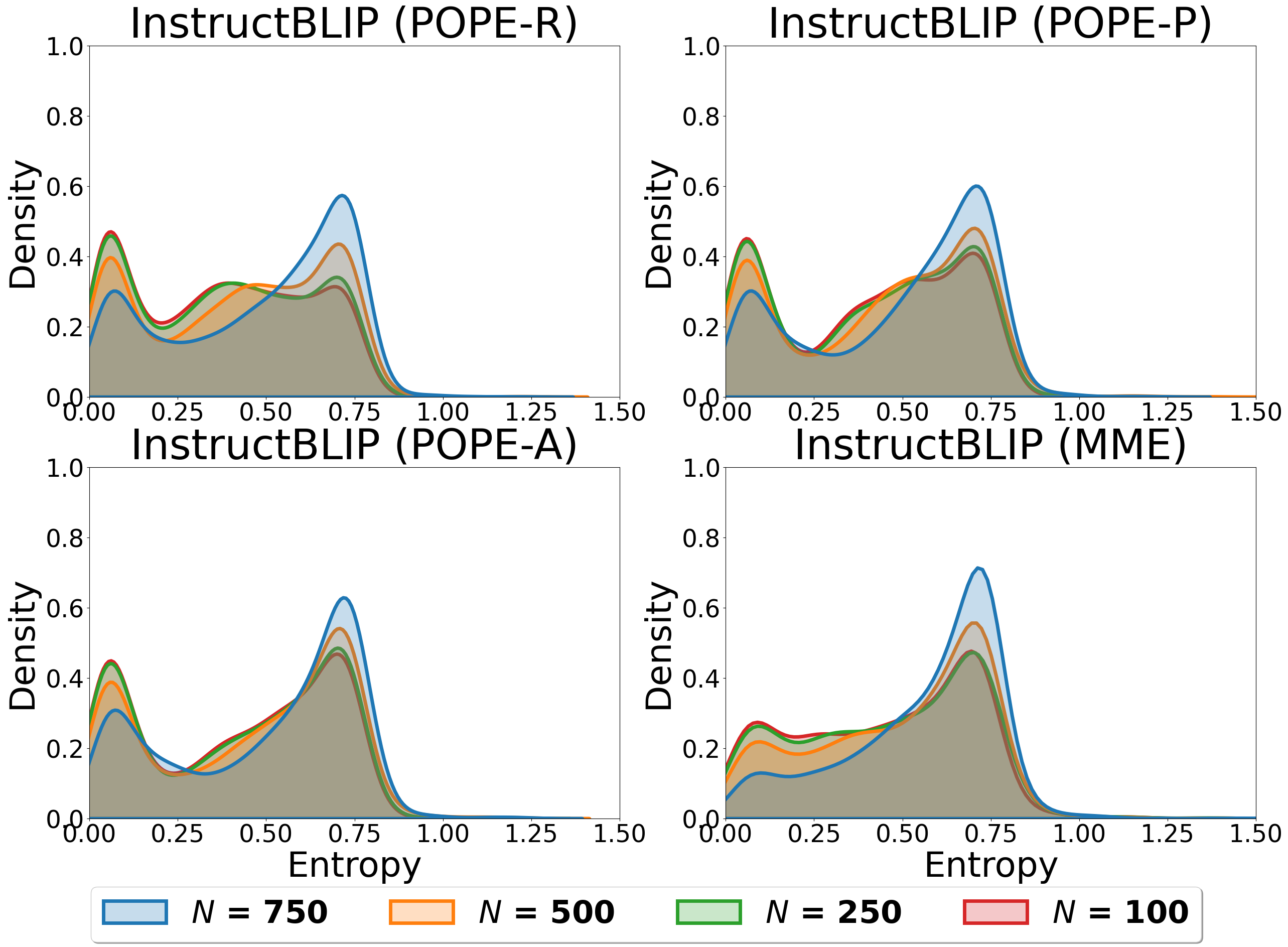}\end{tabular}
        \\
        \end{tabular}
    }
    \caption{The ablation study of different noise steps $N$ for the diffusion noise method.
    }
    \label{fig:abl_diffusion_noise}
\end{figure*}

\begin{figure*}[!t]
    \centering
    \LARGE
    \resizebox{\linewidth}!{
        \begin{tabular}{cc}
        % \multirow{3}{*}{(a)} 
            \begin{tabular}{c|c|c|c|c}
             \shortstack{weight of text \\ guidance $cfg_{text}$}& POPE-R & POPE-P & POPE-A & MME \\ \hline
            $cfg_{text}=20$ & 0.860  & \textbf{0.839}  &  \textbf{0.829} & \textbf{0.747} \\
            $cfg_{text}=15$ & 0.860  & 0.837  &  0.828 & 0.743 \\
            $cfg_{text}=10$ & 0.861  & 0.832  &  0.814 & 0.734 \\
            $cfg_{text}=5$ & \textbf{0.868}  & 0.821  &  0.805 & 0.714 \\
            
            \end{tabular}
        & \begin{tabular}{c}\includegraphics[width=\textwidth]{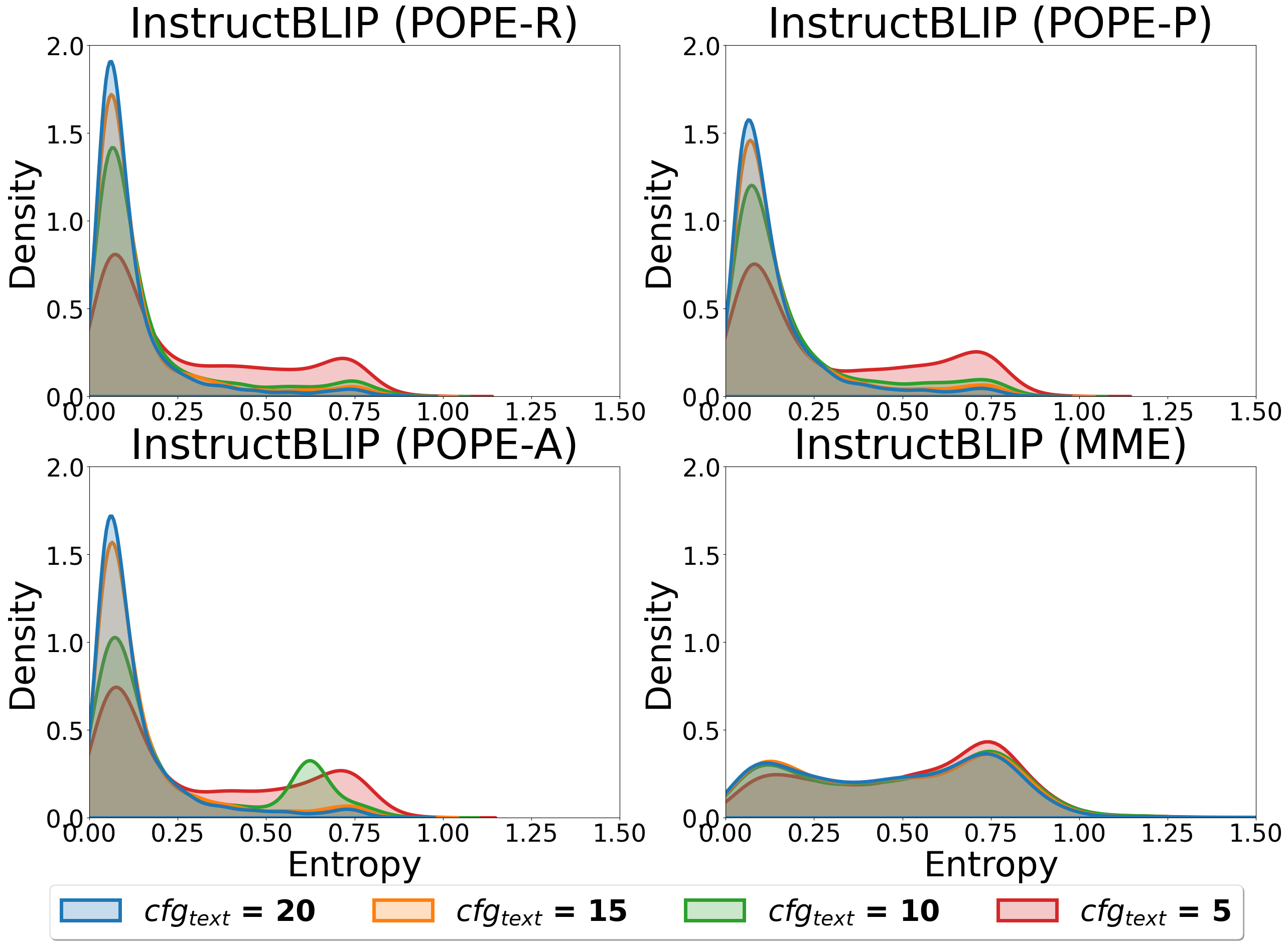}\end{tabular}
        \\
        \end{tabular}
    }
    \caption{The ablation study of different weight of text guidance $cfg_{text}$ for the image editing method.
    }
    \label{fig:abl_edit}
\end{figure*}

We provide the ablation studies of downsample, diffusion noise, and image editing methods in Figure~\ref{fig:abl_downsample}, \ref{fig:abl_diffusion_noise}, and \ref{fig:abl_edit} respectively.
Intuitively, the more severe the distortion, the higher the entropy of CDs, increasing the likelihood of inducing hallucinations, where the downsample and diffusion noise method follow this trend. In contrast, the image editing method that inserts the query objects or contents into images directly presents an opposite trend, where the entropy increases as the weight of text guidance decreases. The general trend is that visually changed samples with more distortion acquire higher performance.

% observations:
% - downsampling: when the ratio increases, the entropy gets higher, and the performance also improves. However, when the ratio is larger than 16, its performance decreases,
% - diffusion noise: with different noise step, the entropy has not obvious different
% - image editing: 

\begin{figure*}[!t]
    \centering
    % \LARGE
    \resizebox{0.95\linewidth}!{
        \begin{tabular}{c}
        % \multirow{3}{*}{(a)} 
        \begin{tabular}{c}\includegraphics[width=\textwidth]{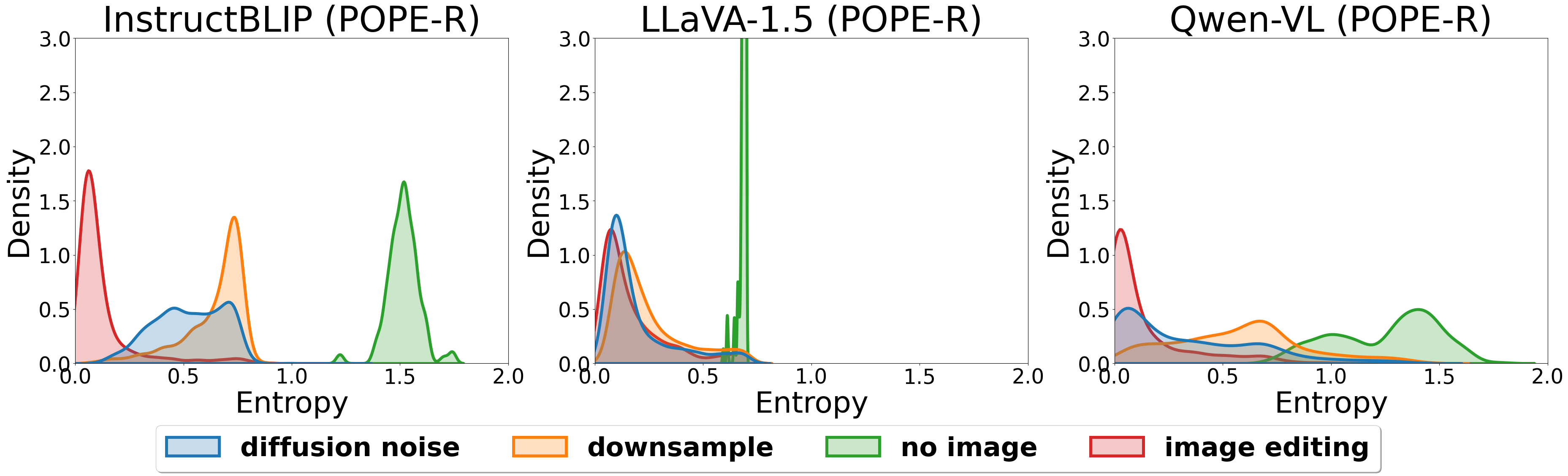}\end{tabular}\\
        (a) POPE \textit{random}\\ \\
        \begin{tabular}{c}\includegraphics[width=\textwidth]{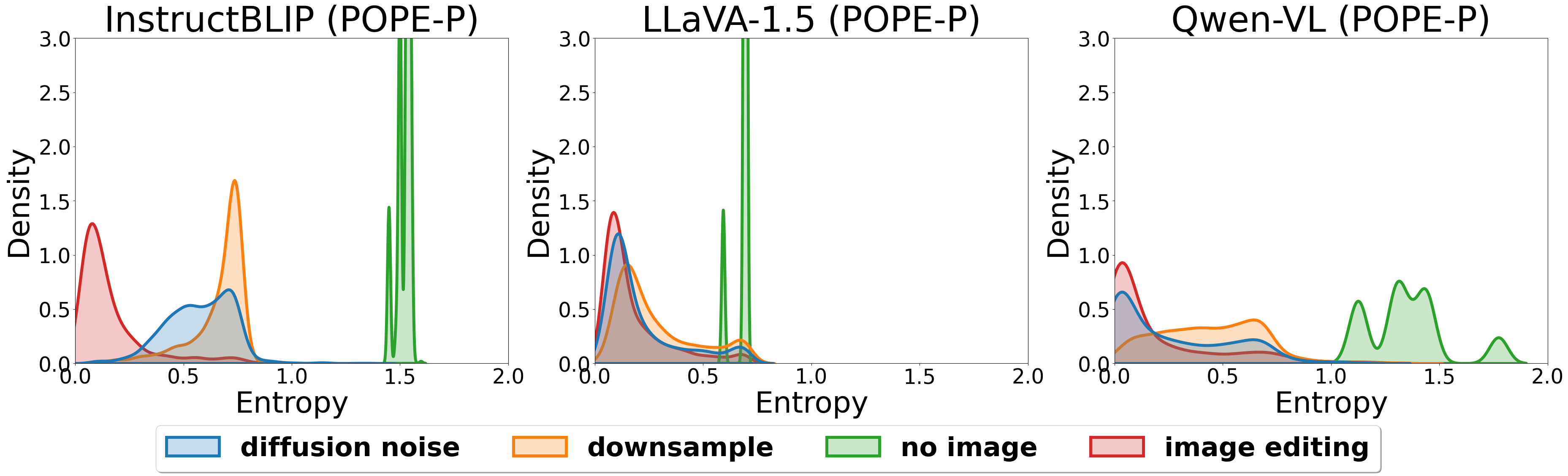}\end{tabular}
        \\
        (b) POPE \textit{popular}\\ \\
        \begin{tabular}{c}\includegraphics[width=\textwidth]{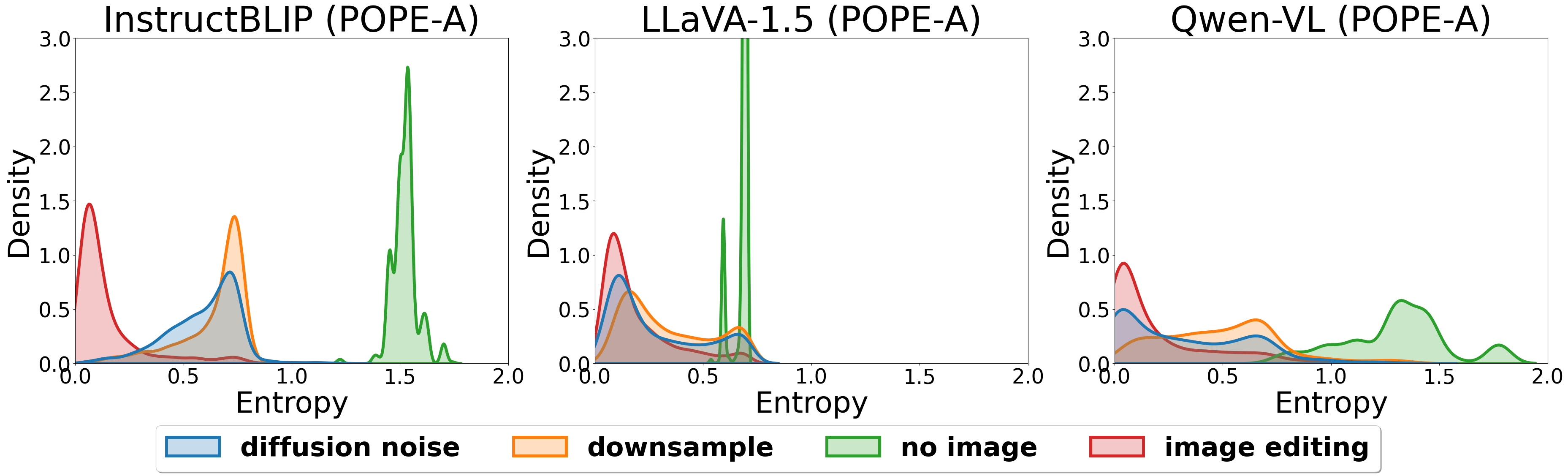}\end{tabular}
        \\
        (c) POPE \textit{adversarial}\\ \\
        \begin{tabular}{c}\includegraphics[width=\textwidth]{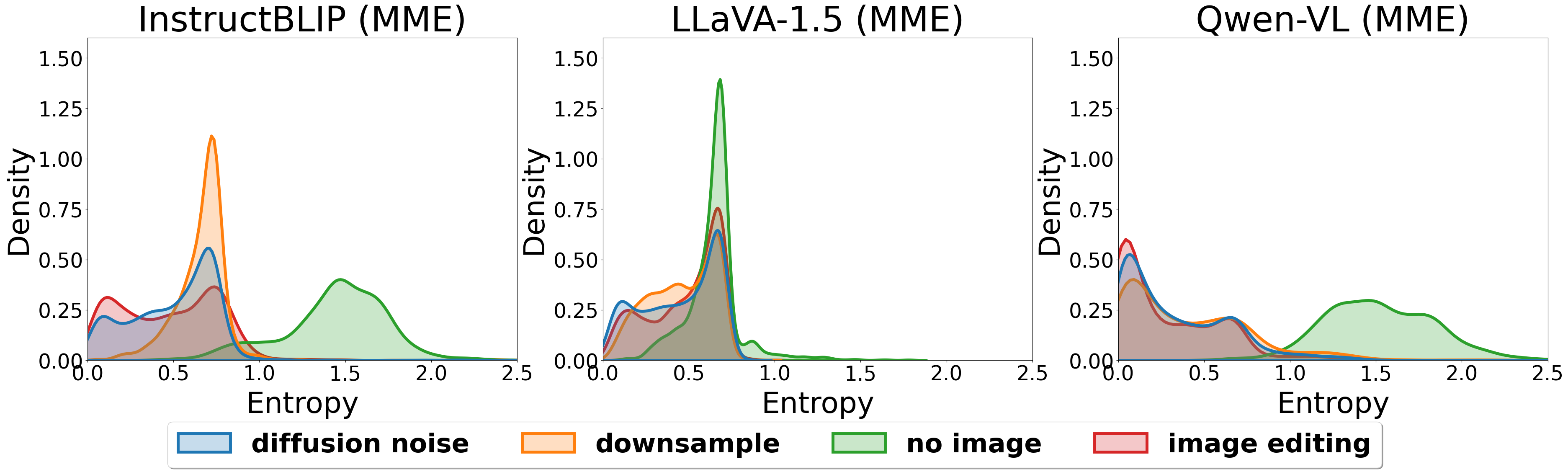}\end{tabular}
        \\
        (d) MME     
        \end{tabular}
    }
    \caption{The analysis of entropy on the POPE and MME benchmarks.
    }
    \label{fig:entropy}
\end{figure*}

\begin{figure*}[!t]
    \centering
    % \LARGE
    \resizebox{0.95\linewidth}!{
        \begin{tabular}{c}
        % \multirow{3}{*}{(a)} 
        \begin{tabular}{c}\includegraphics[width=\textwidth]{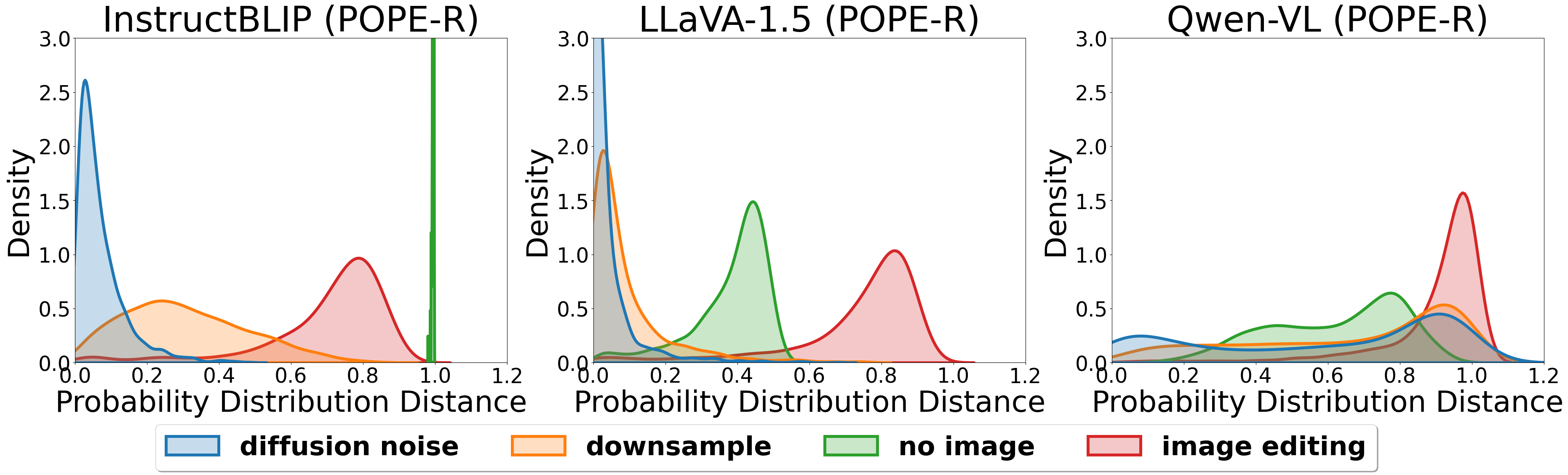}\end{tabular}\\
        (a) POPE \textit{random}\\ \\
        \begin{tabular}{c}\includegraphics[width=\textwidth]{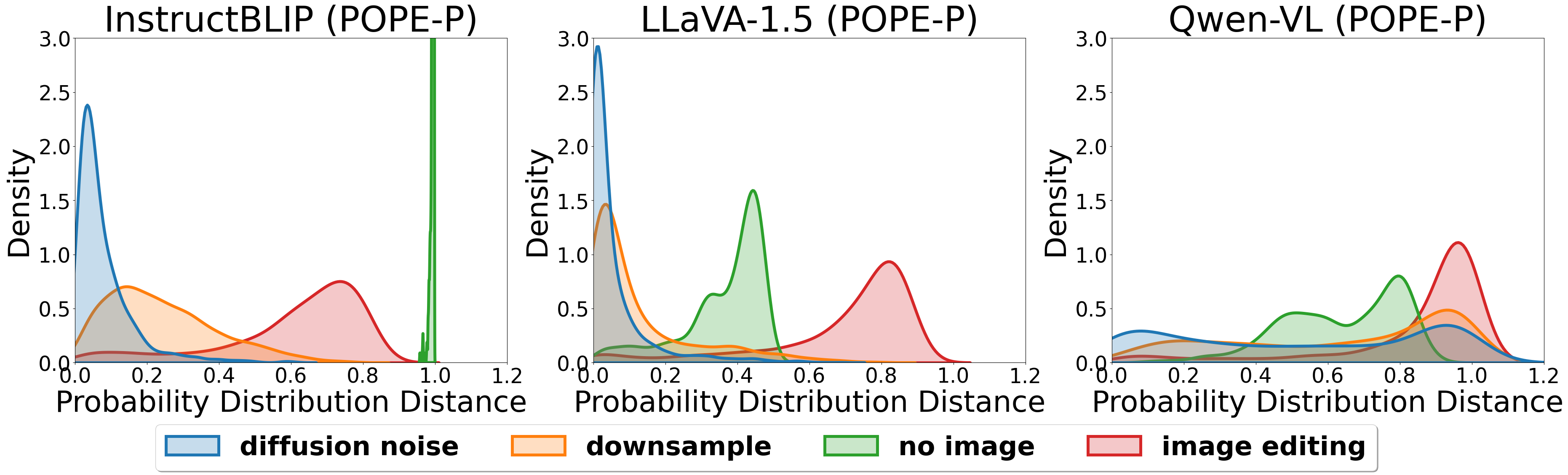}\end{tabular}
        \\
        (b) POPE \textit{popular}\\ \\
        \begin{tabular}{c}\includegraphics[width=\textwidth]{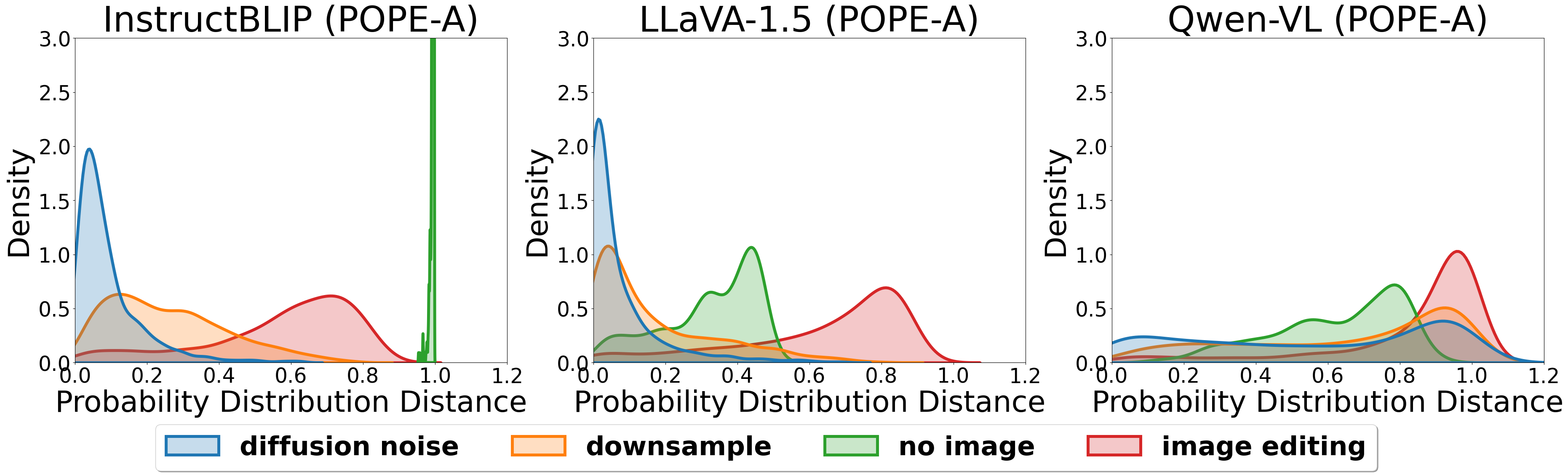}\end{tabular}
        \\
        (c) POPE \textit{adversarial}\\ \\
        \begin{tabular}{c}\includegraphics[width=\textwidth]{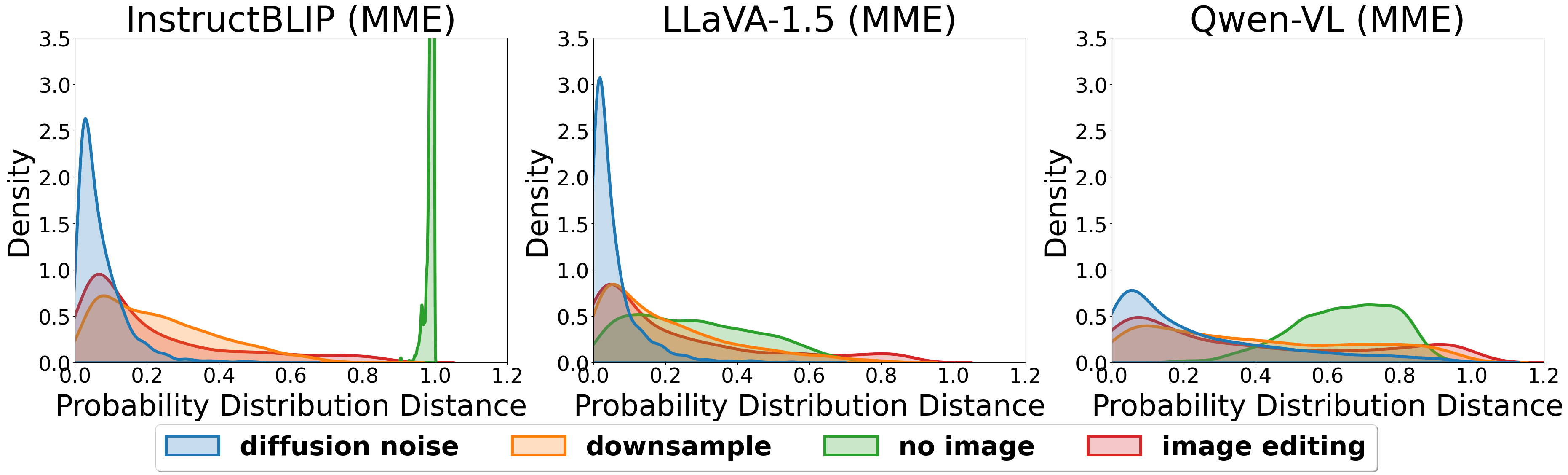}\end{tabular}
        \\
        (d) MME     
        \end{tabular}
    }
    \caption{The analysis of probability distribution distance on the POPE and MME benchmarks.
    }
    \label{fig:probdist}
\end{figure*}

\begin{figure*}[!t]
    \centering
    % \LARGE
    \resizebox{1.0\linewidth}!{
        \begin{tabular}{c}
        % \multirow{3}{*}{(a)} 
        \begin{tabular}{c}\includegraphics[width=\textwidth]{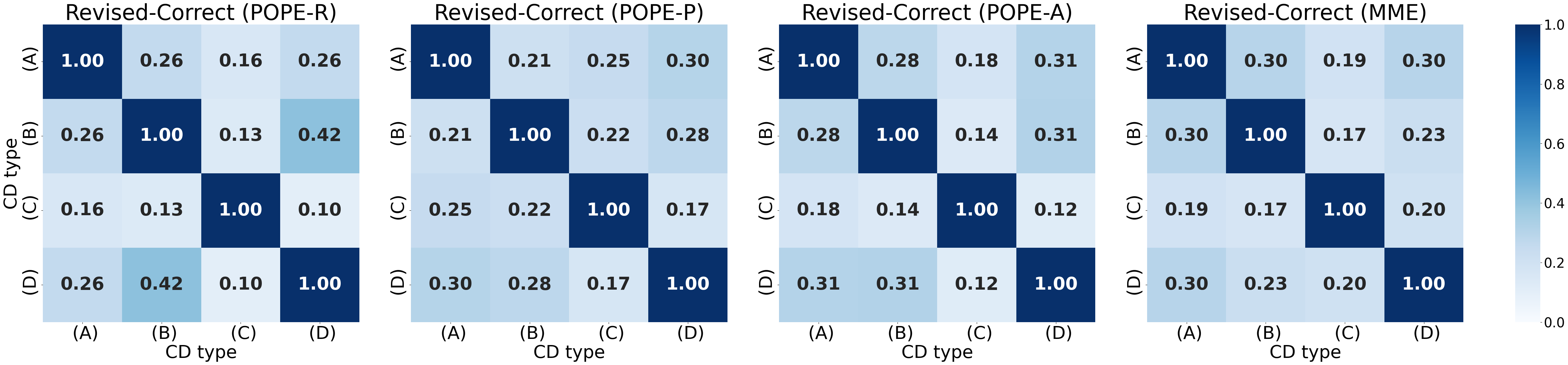}\end{tabular}\\
        (a) InstructBLIP\\ \\
        \begin{tabular}{c}\includegraphics[width=\textwidth]{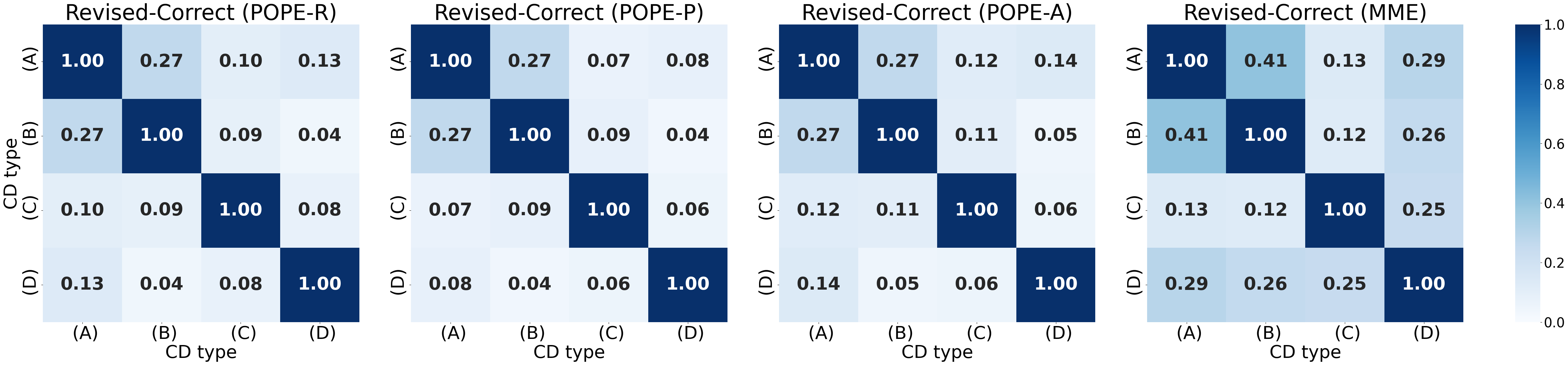}\end{tabular}
        \\
        (b) LLaVA-1.5\\ \\
        \begin{tabular}{c}\includegraphics[width=\textwidth]{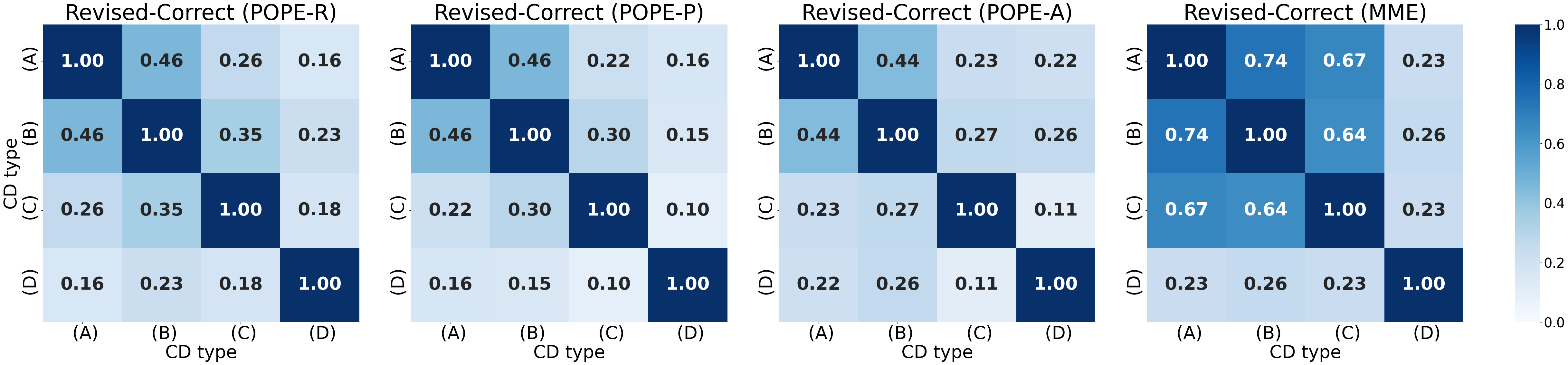}\end{tabular}
        \\
        (c) Qwen-VL
    
        \end{tabular}
    }
    \caption{The pairwise overlap of revise-correct samples. CD types consist of (A) diffusion noise, (B) downsample, (C) no image, and (D) image editing.
    }
    \label{fig:overlapping}
\end{figure*}

\begin{table*}[t]
\resizebox{\linewidth}!{
    \begin{tabular}{c|c|c|c|c|c|c|c|c|c|c|c|c|c}
    VQA (Y/N)                & \multicolumn{4}{c|}{InstructBLIP} & \multicolumn{4}{c|}{LLaVA-1.5} & \multicolumn{4}{c|}{QwenVL}    & Overall \\ \hline
    Benchmark (Acc)            & P-R    & P-P    & P-A    & MME   & P-R   & P-P   & P-A   & MME   & P-R   & P-P   & P-A   & MME   & All     \\ \hline
    Original                   & 0.825  & 0.733  & 0.752  & 0.688 & 0.832 & 0.817 & 0.788 & 0.705 & 0.847 & 0.851 & 0.822 & 0.720 & 0.785   \\ \hline
    % ICD~\cite{wang2024icd}     & 0.884  & 0.817  & 0.792  & 0.726 &       &       &       &       &       &       &       &       &         \\
    diffusion noise (VCD~\cite{leng2024vcd}) (A)    & 0.858  & 0.805  & 0.779  & 0.720 & 0.863 & 0.842 & 0.805 & 0.733 & 0.865 & 0.865 & 0.832 & 0.747 & 0.810   \\
    no image (M3ID~\cite{favero2024m3id}) (B) & 0.880  & 0.835  & 0.802  & 0.742 & 0.871 & \second{0.856} & \best{0.827} & 0.732 & 0.868 & 0.871 & 0.837 & 0.733 & 0.821   \\
    downsample (C)                & 0.878  & 0.824  & 0.801  & 0.724 & \best{0.879} & \second{0.856} & 0.811 & \second{0.765} & \best{0.875} & 0.871 & 0.842 & \best{0.764} & 0.824   \\
    image editing (D)                 & 0.860  & 0.839  & \best{0.829}  & 0.747 & 0.832 & 0.826 & 0.811 & 0.748 & 0.845 & 0.838 & 0.824 & \second{0.763} & 0.814   \\ \hline \hline
    % A+B              & 0.  & 0.  & 0.  & 0. & 0. & 0. & 0. & 0. & 0. & 0. & 0. & 0. & 0.      \\ 
    % A+B+C              & 0.  & 0.  & 0.  & 0. & 0. & 0. & 0. & 0. & 0. & 0. & 0. & 0. & 0.    \\   
    % A+B+C+D              & 0.  & 0.  & 0.  & 0. & 0. & 0. & 0. & 0. & 0. & 0. & 0. & 0. & 0.     \\  
    % simple fusion              & 0.864  & 0.840  & 0.832  & 0.774 & 0.862 & 0.847 & 0.827 & 0.779 & 0.865 & 0.860 & 0.839 & 0.768 & 0.830   \\ \hline \hline
    A+B         & 0.885  & 0.847  & 0.813  & 0.757 & 0.872 & 0.853 & 0.819 & 0.747 & 0.870 & 0.872 & 0.841 & 0.738 & 0.826  \\   
    A+C         & 0.873  & 0.822  & 0.797  & 0.727 & 0.870 & 0.855 & 0.817 & 0.761 & 0.872 & 0.869 & 0.839 & 0.739 & 0.820      \\   
    A+D         & 0.865  & 0.827  & 0.799  & 0.743 & 0.866 & 0.844 & 0.818 & 0.750 & 0.862 & 0.862 & 0.836 & 0.739 & 0.818      \\ 
    B+C         & 0.887  & 0.850  & 0.816  & 0.757 & \second{0.877} & \best{0.863} & 0.824 & 0.758 & \best{0.875} & \second{0.873} & \best{0.844} & 0.741 & 0.830      \\ 
    B+D         & 0.883  & 0.851  & 0.815  & 0.764 & 0.865 & 0.849 & 0.824 & 0.749 & 0.869 & 0.870 & 0.838 & 0.744 & 0.827      \\ 
    C+D         & 0.874  & 0.834  & 0.805  & 0.740 & 0.864 & 0.848 & 0.817 & 0.754 & 0.866 & 0.867 & 0.839 & 0.746 & 0.821      \\ 
    A+B+C       & \best{0.892}  & 0.849  & 0.818  & 0.758 & 0.876 & \second{0.856} & 0.820 & 0.762 & \second{0.874} & \best{0.874} & \best{0.844} & 0.742 & 0.829 \\ 
    A+B+D       & 0.888  & 0.852  & 0.820  & 0.765 & 0.865 & 0.852 & \second{0.825} & 0.754 & 0.869 & 0.870 & 0.840 & 0.743 & 0.829     \\      
    A+C+D       & 0.878  & 0.832  & 0.805  & 0.740 & 0.868 & 0.852 & 0.820 & 0.762 & 0.865 & 0.870 & 0.838 & 0.748 & 0.823     \\      
    B+C+D       & 0.889  & \second{0.854}  & 0.822  & \second{0.770} & 0.869 & 0.854 & \second{0.825} & \best{0.770} & 0.873 & 0.871 & \best{0.844} & 0.748 & \second{0.832}     \\      
    A+B+C+D     & \second{0.890}  & \best{0.857}  & \second{0.828}  & \best{0.774} & 0.872 & \second{0.856} & 0.824 & 0.764 & 0.872 & \second{0.873} & \best{0.844} & 0.749 & \best{0.833}     \\  
    % entropy-weighted fusion    & 0.890  & 0.857  & 0.828  & 0.774 & 0.871 & 0.856 & 0.824 & 0.764 & 0.871 & 0.872 & 0.844 & 0.749 & 0.833   \\

\end{tabular}
}
    \caption{The overall results on the Yes-No question tasks, including POPE and MME. The results below are entropy-weighted fusion methods with different combinations of visually changed samples. P-R denotes the \textit{random} task of POPE, P-P denotes the \textit{popular} task of POPE., and P-A denotes the \textit{adversarial} task of POPE. The best performances within each set are marked \textbf{bold}, and the second highest performances are \underline{underlined}.
    }
    \label{table:combinations_supp}
\end{table*}

\begin{table*}[]
\resizebox{\linewidth}!{
    \begin{tabular}{c|c|c|c|c|c|c|c|c|c|c|c|c|c}
    VQA (Y/N)                & \multicolumn{4}{c|}{InstructBLIP} & \multicolumn{4}{c|}{LLaVA-1.5} & \multicolumn{4}{c|}{QwenVL}    & Overall \\ \hline
    Benchmark (Acc)            & P-R    & P-P    & P-A    & MME   & P-R   & P-P   & P-A   & MME   & P-R   & P-P   & P-A   & MME   & All     \\ \hline
    Original                   & 0.825  & 0.733  & 0.752  & 0.688 & 0.832 & 0.817 & 0.788 & 0.705 & 0.847 & 0.851 & 0.822 & 0.720 & 0.785   \\ \hline
    % ICD~\cite{wang2024icd}     & 0.884  & 0.817  & 0.792  & 0.726 &       &       &       &       &       &       &       &       &         \\
    VCD~\cite{leng2024vcd}     & 0.858  & 0.805  & 0.779  & 0.720 & 0.863 & 0.842 & 0.805 & 0.733 & 0.865 & 0.865 & 0.832 & 0.747 & 0.810   \\
    M3ID~\cite{favero2024m3id} & \second{0.880}  & 0.835  & 0.802  & 0.742 & 0.871 & \best{0.856} & 0.827 & 0.732 & 0.868 & \second{0.871} & 0.837 & 0.733 & 0.821   \\
    downsample                 & 0.878  & 0.824  & 0.801  & 0.724 & \best{0.879} & \best{0.856} & 0.811 & 0.765 & \best{0.875} & \second{0.871} & 0.842 & 0.764 & 0.824   \\
    IP2P-edit                  & 0.860  & 0.839  & 0.829  & 0.747 & 0.832 & 0.826 & 0.811 & 0.748 & 0.845 & 0.838 & 0.824 & 0.763 & 0.814   \\ \hline
    naive fusion              & 0.864  & 0.840  & \best{0.832}  & \second{0.774} & 0.862 & 0.847 & \second{0.827} & \second{0.779} & 0.865 & 0.860 & 0.839 & \best{0.768} & \second{0.830}   \\
    entropy-weighted fusion    & \best{0.890}  & \best{0.857}  & 0.828  & \second{0.774} & 0.871 & \best{0.856} & 0.824 & 0.764 & \second{0.871} & \best{0.872} & \best{0.844} & 0.749 & \best{0.833}   \\
    PDD-weighted fusion   & 0.876  & \second{0.849}  & \second{0.830}  & 0.760 & \second{0.872} & 0.849 & 0.825 & 0.755 & 0.866 & 0.861 & 0.838 & 0.762 & 0.829  \\
    conf-weighted fusion       & 0.861  & 0.839  & \best{0.832}  & 0.766 & 0.865 & 0.845 & 0.824 & 0.775 & 0.862 & 0.858 & 0.835 & \best{0.768} & 0.828  \\
    unconf-weighted fusion     & 0.868  & 0.843  & \best{0.832}  & \best{0.778} & 0.871 & \second{0.853} & \best{0.829} & \best{0.780} & \second{0.871} & 0.865 & \second{0.843} & \second{0.767} & \best{0.833}  
\end{tabular}
    \caption{The overall results on the Yes-No question tasks, including POPE and MME. Please refer to Sec.~\ref{sec:fusion} for the formulation of fusion methods. P-R denotes the \textit{random} task of POPE, P-P denotes the \textit{popular} task of POPE., and P-A denotes the \textit{adversarial} task of POPE. The best performances within each set are marked \textbf{bold}, and the second highest performances are \underline{underlined}.
    }
    \label{table:weighted}
}
\end{table*}

\section{Examples of Image Editing}
\begin{figure*}[!t]
    \centering
    \resizebox{\linewidth}!{
        \Huge
        \begin{tabular}{c||cccc}
        % \multirow{3}{*}{(a)} 
        \begin{tabular}{c}\includegraphics[width=0.5\textwidth]{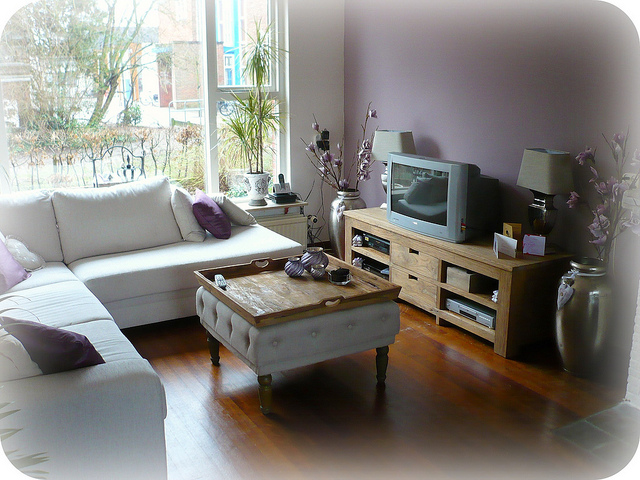}\end{tabular}
        &\begin{tabular}{c}\includegraphics[width=0.5\textwidth]{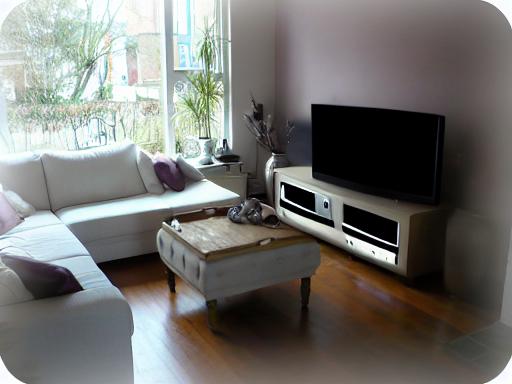}\end{tabular}
        &\begin{tabular}{c}\includegraphics[width=0.5\textwidth]{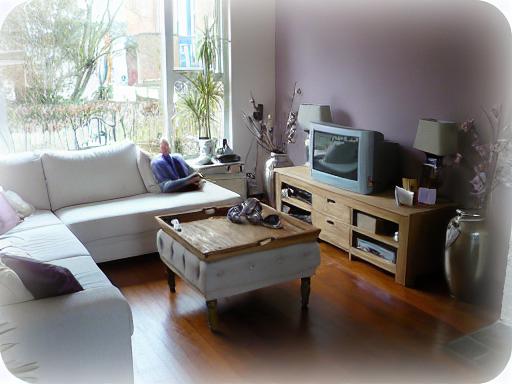}\end{tabular}
        &\begin{tabular}{c}\includegraphics[width=0.5\textwidth]{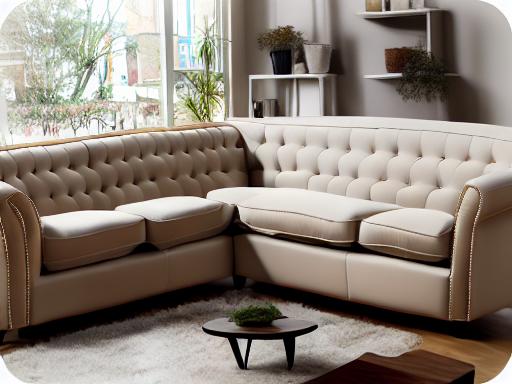}\end{tabular}
        &\begin{tabular}{c}\includegraphics[width=0.5\textwidth]{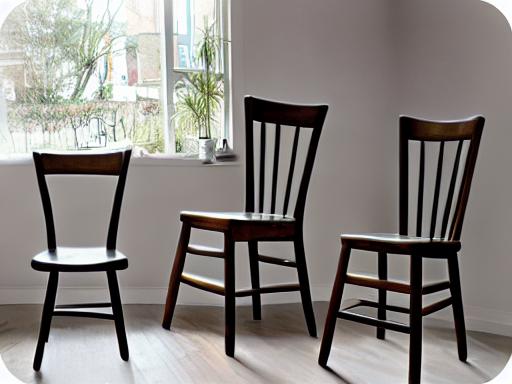}\end{tabular}\\
        \textbf{input} & \red{a tv} & \blue{a person} & \red{a couch} & \blue{a chair} \\
        
        \begin{tabular}{c}\includegraphics[width=0.5\textwidth]{supp_figures/vis_edit/ex2/input.jpg}\end{tabular}
        &\begin{tabular}{c}\includegraphics[width=0.5\textwidth]{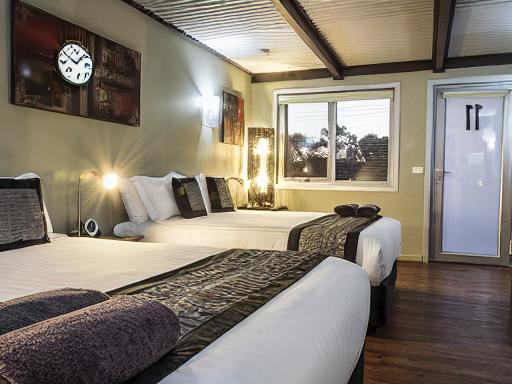}\end{tabular}
        &\begin{tabular}{c}\includegraphics[width=0.5\textwidth]{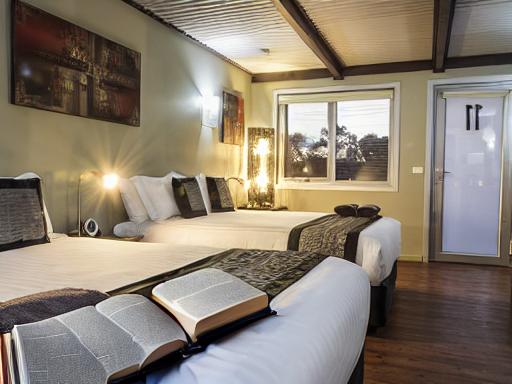}\end{tabular}
        &\begin{tabular}{c}\includegraphics[width=0.5\textwidth]{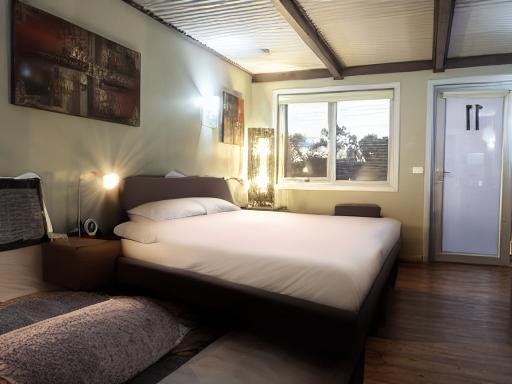}\end{tabular}
        &\begin{tabular}{c}\includegraphics[width=0.5\textwidth]{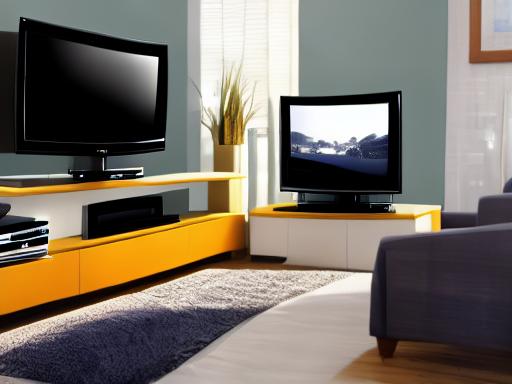}\end{tabular}\\
        \textbf{input} & \red{a clock} & \blue{a book} & \red{a bed} & \blue{a tv} \\
        
        \begin{tabular}{c}\includegraphics[width=0.5\textwidth]{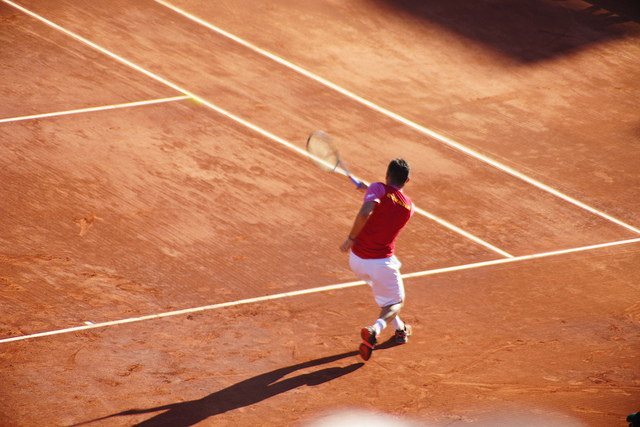}\end{tabular}
        &\begin{tabular}{c}\includegraphics[width=0.5\textwidth]{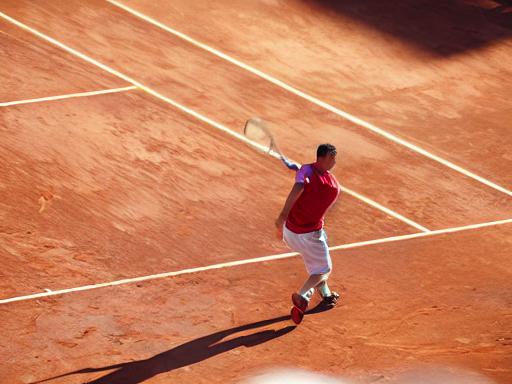}\end{tabular}
        &\begin{tabular}{c}\includegraphics[width=0.5\textwidth]{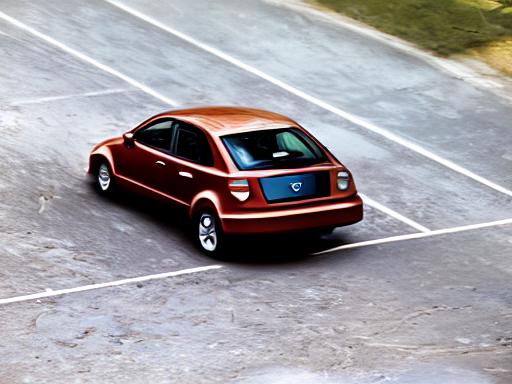}\end{tabular}
        &\begin{tabular}{c}\includegraphics[width=0.5\textwidth]{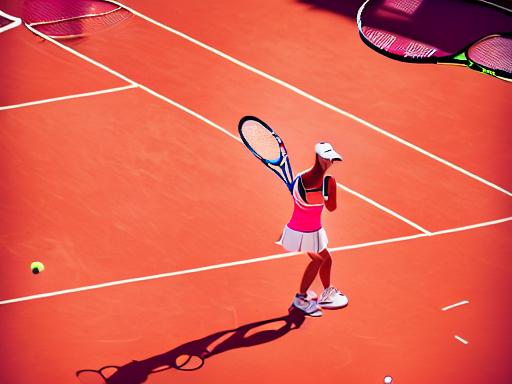}\end{tabular}
        &\begin{tabular}{c}\includegraphics[width=0.5\textwidth]{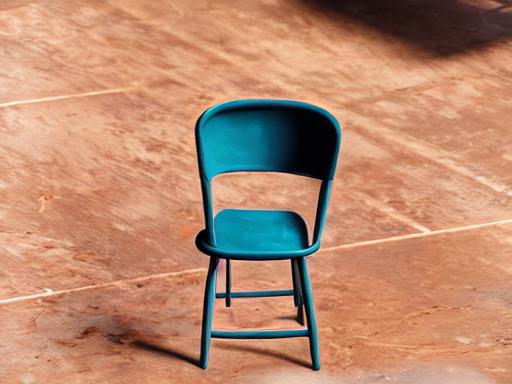}\end{tabular}\\        
        \textbf{input} & \red{a person} & \blue{a car} & \red{a tennis racket} & \blue{a chair} \\

        \begin{tabular}{c}\includegraphics[width=0.5\textwidth]{supp_figures/vis_edit/ex4/input.jpg}\end{tabular}
        &\begin{tabular}{c}\includegraphics[width=0.5\textwidth]{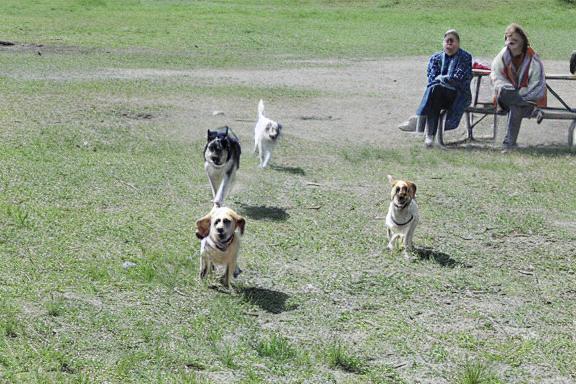}\end{tabular}
        &\begin{tabular}{c}\includegraphics[width=0.5\textwidth]{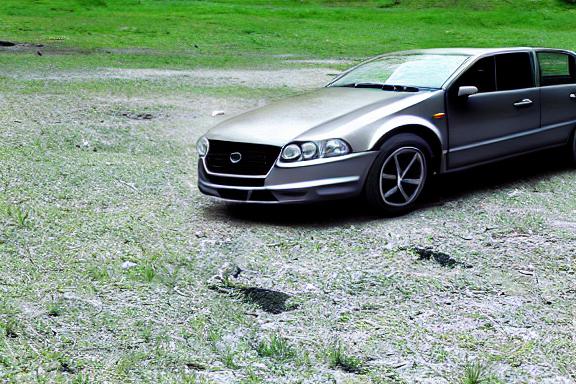}\end{tabular}
        &\begin{tabular}{c}\includegraphics[width=0.5\textwidth]{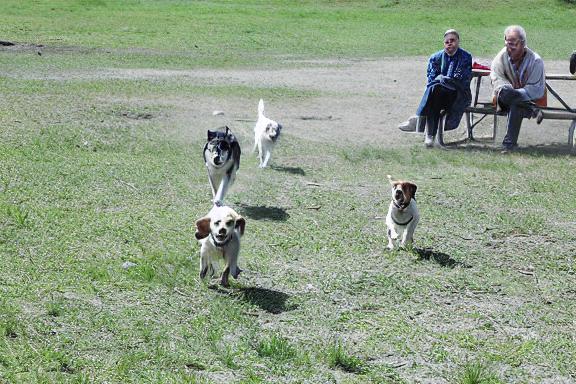}\end{tabular}
        &\begin{tabular}{c}\includegraphics[width=0.5\textwidth]{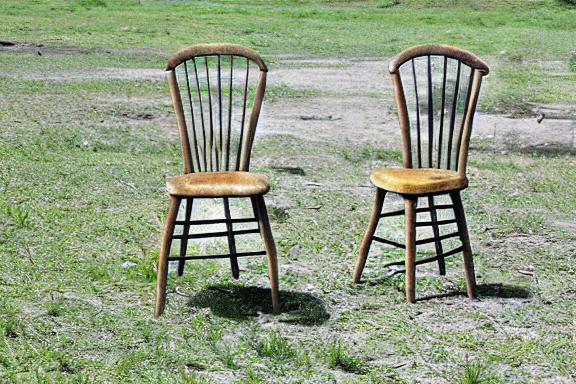}\end{tabular}\\        
        \textbf{input} & \red{a dog} & \blue{a car} & \red{a person} & \blue{a chair} \\

        \begin{tabular}{c}\includegraphics[width=0.5\textwidth]{supp_figures/vis_edit/ex5/input.jpg}\end{tabular}
        &\begin{tabular}{c}\includegraphics[width=0.5\textwidth]{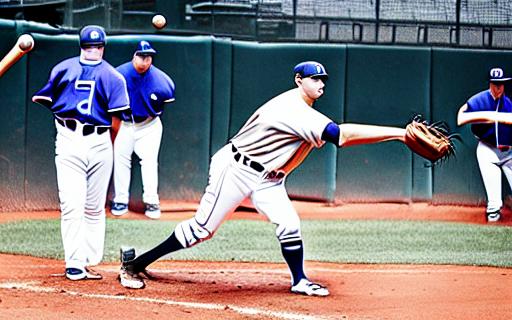}\end{tabular}
        &\begin{tabular}{c}\includegraphics[width=0.5\textwidth]{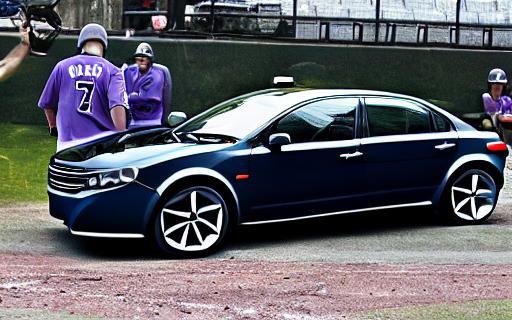}\end{tabular}
        &\begin{tabular}{c}\includegraphics[width=0.5\textwidth]{supp_figures/vis_edit/ex5/Q3.jpg}\end{tabular}
        &\begin{tabular}{c}\includegraphics[width=0.5\textwidth]{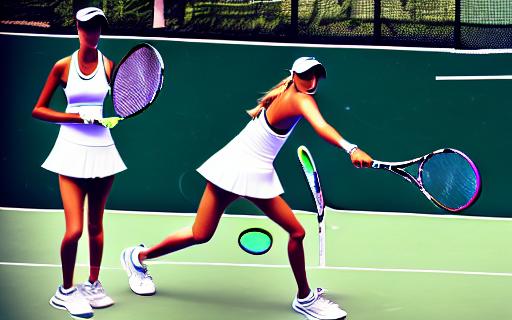}\end{tabular}\\        
        \textbf{input} & \red{a baseball glove} & \blue{a car} & \red{a person} & \blue{a tennis racket} \\

        \begin{tabular}{c}\includegraphics[width=0.5\textwidth]{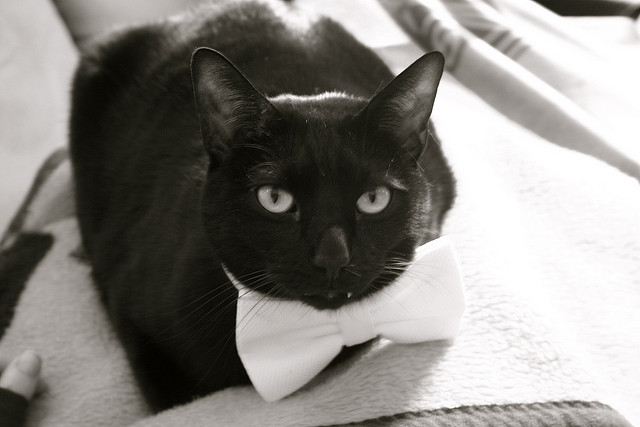}\end{tabular}
        &\begin{tabular}{c}\includegraphics[width=0.5\textwidth]{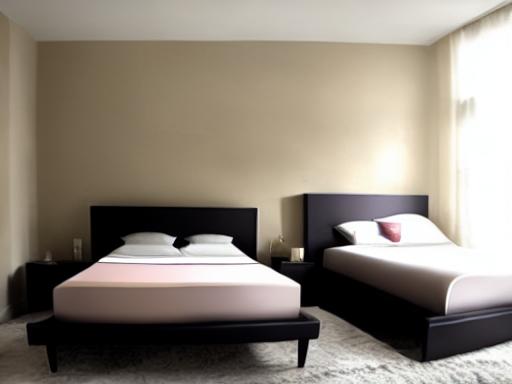}\end{tabular}
        &\begin{tabular}{c}\includegraphics[width=0.5\textwidth]{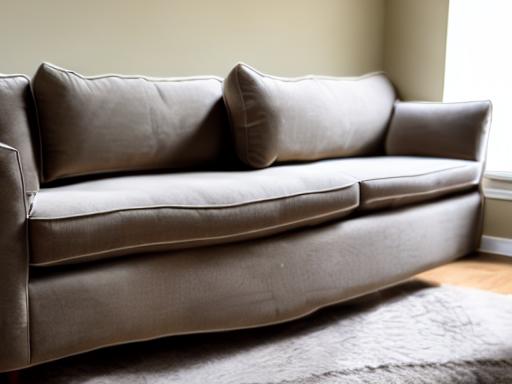}\end{tabular}
        &\begin{tabular}{c}\includegraphics[width=0.5\textwidth]{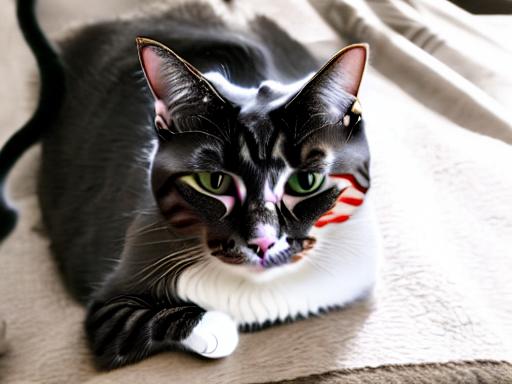}\end{tabular}
        &\begin{tabular}{c}\includegraphics[width=0.5\textwidth]{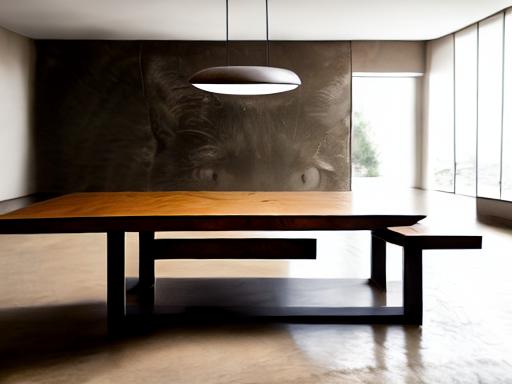}\end{tabular}\\        
        \textbf{input} & \red{a bed} & \blue{a couch} & \red{a cat} & \blue{a dining table} \\        
        
        \end{tabular}
    }
    \caption{Examples of image editing on POPE benchmark.
    }
    \label{fig:vis_edit_pope}
\end{figure*}

\begin{figure*}[!t]
    \centering
    \resizebox{\linewidth}!{
        \Huge
        \begin{tabular}{c||cc}
        % \multirow{3}{*}{(a)} 
        \begin{tabular}{c}\includegraphics[width=0.5\textwidth]{supp_figures/vis_edit/mme_ex1/input.jpg}\\\textbf{input}\end{tabular}
        &\begin{tabular}{c}\includegraphics[width=0.5\textwidth]{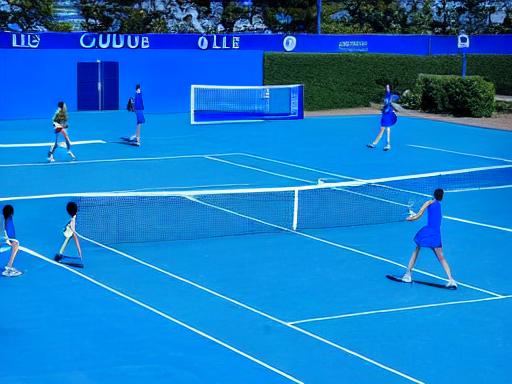}\\\red{a blue court}\end{tabular}
        &\begin{tabular}{c}\includegraphics[width=0.5\textwidth]{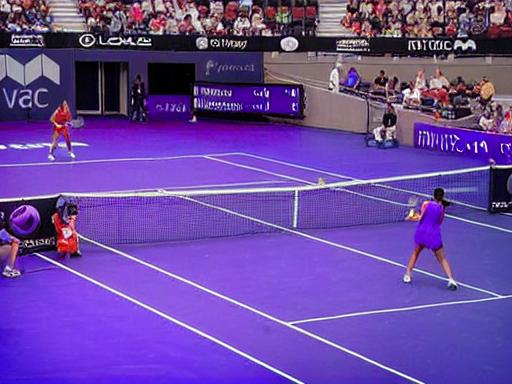}\\\blue{a purple court}\end{tabular}\\\\

        \begin{tabular}{c}\includegraphics[width=0.5\textwidth]{supp_figures/vis_edit/mme_ex2/input.jpg}\\\textbf{input}\end{tabular}
        &\begin{tabular}{c}\includegraphics[width=0.5\textwidth]{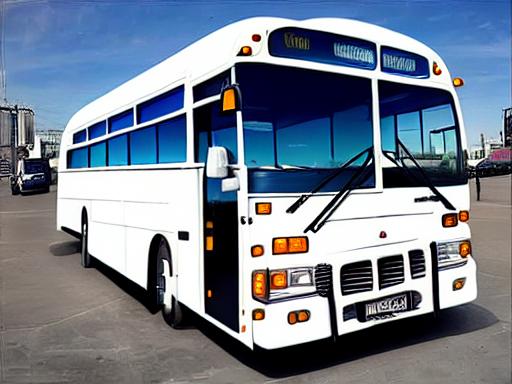}\\\red{a white bus }\end{tabular}
        &\begin{tabular}{c}\includegraphics[width=0.5\textwidth]{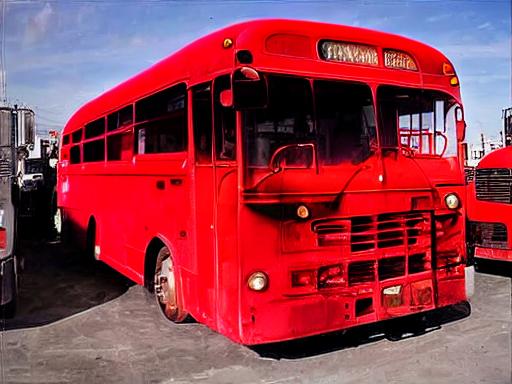}\\\blue{a red bus}\end{tabular}\\\\

        \begin{tabular}{c}\includegraphics[width=0.5\textwidth]{supp_figures/vis_edit/mme_ex3/input.jpg}\\\textbf{input}\end{tabular}
        &\begin{tabular}{c}\includegraphics[width=0.5\textwidth]{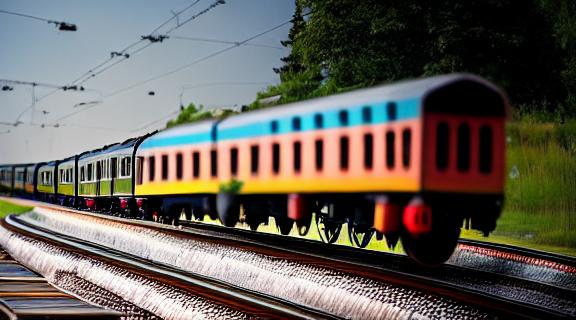}\\\red{a train}\end{tabular}
        &\begin{tabular}{c}\includegraphics[width=0.5\textwidth]{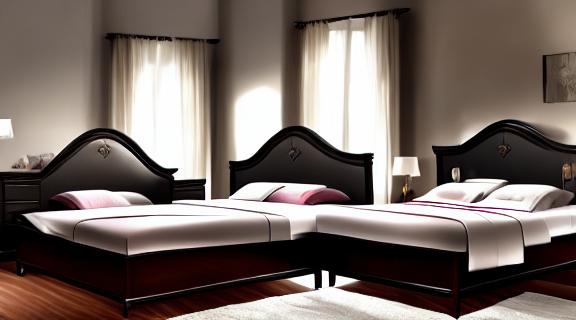}\\\blue{a bed}\end{tabular}\\\\
        
        \begin{tabular}{c}\includegraphics[width=0.5\textwidth]{supp_figures/vis_edit/mme_ex4/input.jpg}\\\textbf{input}\end{tabular}
        &\begin{tabular}{c}\includegraphics[width=0.5\textwidth]{supp_figures/vis_edit/mme_ex4/Q1.jpg}\\\red{a person}\end{tabular}
        &\begin{tabular}{c}\includegraphics[width=0.5\textwidth]{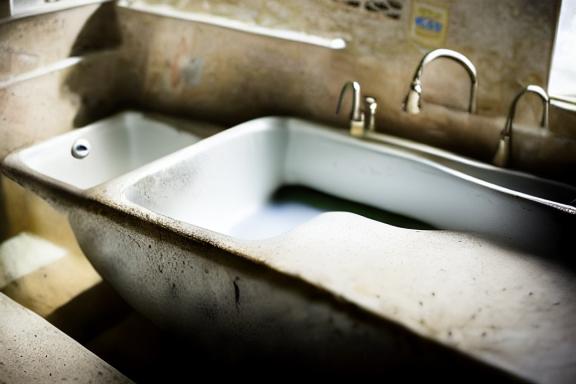}\\\blue{a sink}\end{tabular}\\\\

        \end{tabular}
    }
    \caption{Examples of image editing on coarse-grained perception tasks (color and existence) in MME benchmark.
    }
    \label{fig:vis_edit_mme_c}
\end{figure*}

\begin{figure*}[!t]
    \centering
    \resizebox{\linewidth}!{
        \Huge
        \begin{tabular}{c||cc}
        % \multirow{3}{*}{(a)} 
        \begin{tabular}{c}\includegraphics[width=0.5\textwidth]{supp_figures/vis_edit/mme_ex5/input.jpg}\\\textbf{input}\end{tabular}
        &\begin{tabular}{c}\includegraphics[width=0.5\textwidth]{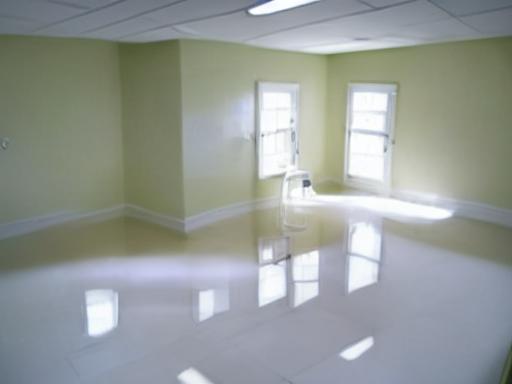}\\\red{a clean room}\end{tabular}
        &\begin{tabular}{c}\includegraphics[width=0.5\textwidth]{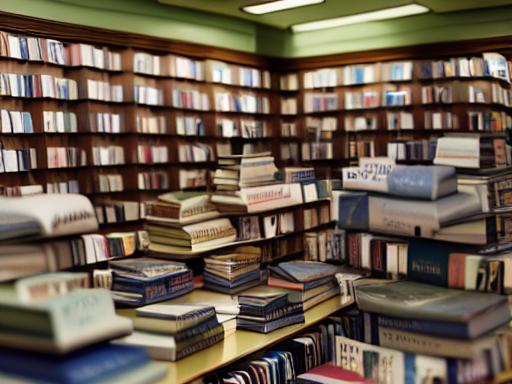}\\\blue{a bookstore}\end{tabular}\\\\
        
        \begin{tabular}{c}\includegraphics[width=0.5\textwidth]{supp_figures/vis_edit/mme_ex6/input.jpg}\\\textbf{input}\end{tabular}
        &\begin{tabular}{c}\includegraphics[width=0.5\textwidth]{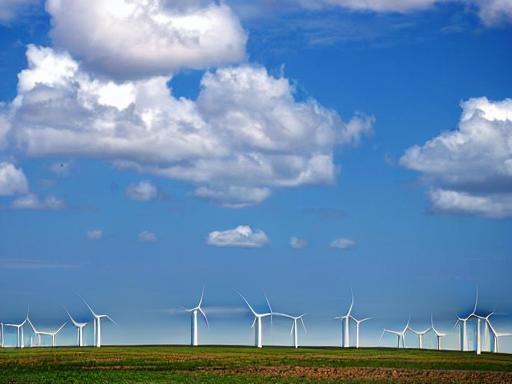}\\\red{a wind farm}\end{tabular}
        &\begin{tabular}{c}\includegraphics[width=0.5\textwidth]{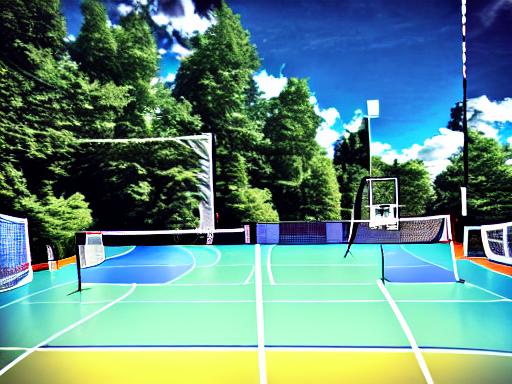}\\\blue{a volleyball court outdoor}\end{tabular}\\\\        
        
        \begin{tabular}{c}\includegraphics[width=0.5\textwidth]{supp_figures/vis_edit/mme_ex7/input.jpg}\\\textbf{input}\end{tabular}
        &\begin{tabular}{c}\includegraphics[width=0.5\textwidth]{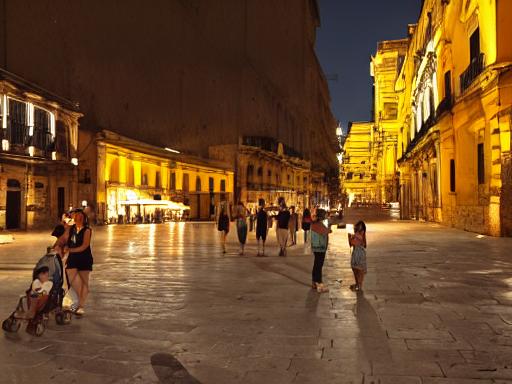}\\\red{Ortigia}\end{tabular}
        &\begin{tabular}{c}\includegraphics[width=0.5\textwidth]{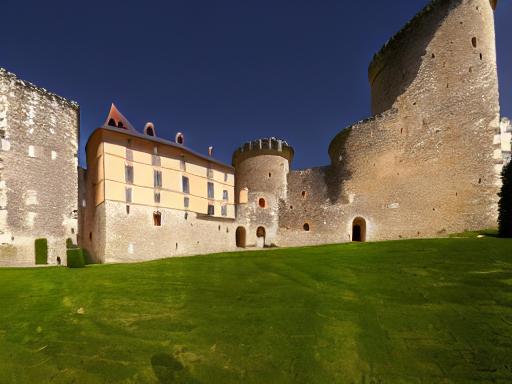}\\\blue{Montmayeur castle}\end{tabular}\\\\
        
        \begin{tabular}{c}\includegraphics[width=0.5\textwidth]{supp_figures/vis_edit/mme_ex8/input.jpg}\\\textbf{input}\end{tabular}
        &\begin{tabular}{c}\includegraphics[width=0.5\textwidth]{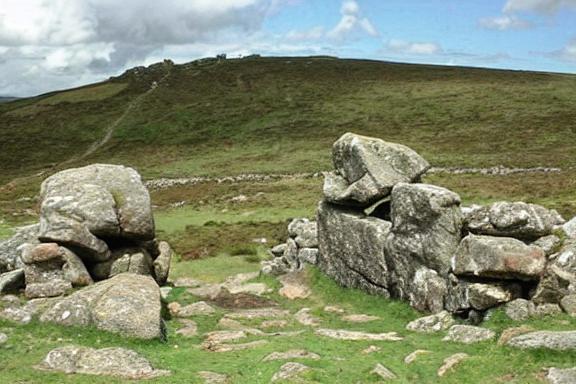}\\\red{Grimspound}\end{tabular}
        &\begin{tabular}{c}\includegraphics[width=0.5\textwidth]{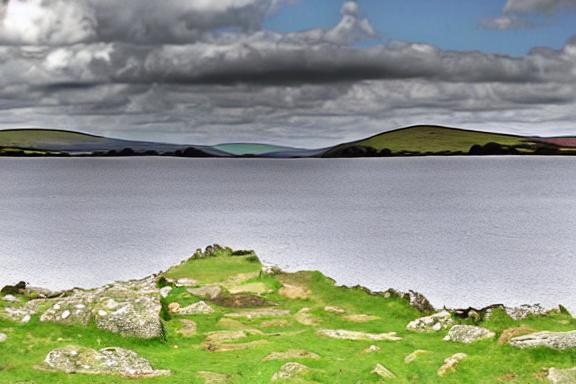}\\\blue{Lough Owel}\end{tabular}\\\\                 
        
        \end{tabular}
    }
    \caption{Examples of image editing on fine-grained perception tasks (scene and landmark) in MME benchmark.
    }
    \label{fig:vis_edit_mme_f}
\end{figure*}

\begin{figure*}[!t]
    \centering
    \resizebox{\linewidth}!{
        \Huge
        \begin{tabular}{c||cc}
        % \multirow{3}{*}{(a)} 
        \begin{tabular}{c}\includegraphics[width=0.5\textwidth]{supp_figures/vis_edit/mme_ex9/input.jpg}\\\textbf{input}\end{tabular}
        &\begin{tabular}{c}\includegraphics[width=0.5\textwidth]{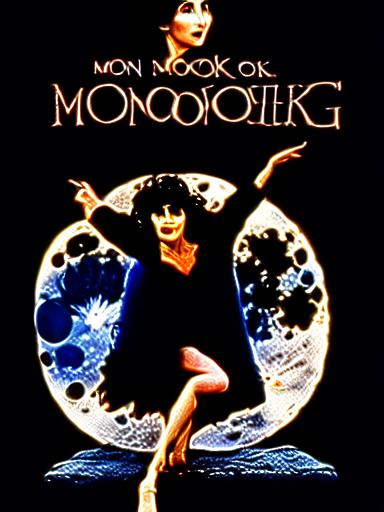}\\\red{\shortstack{This movie is titled \\ moonstruck (1987).}}\end{tabular}
        &\begin{tabular}{c}\includegraphics[width=0.5\textwidth]{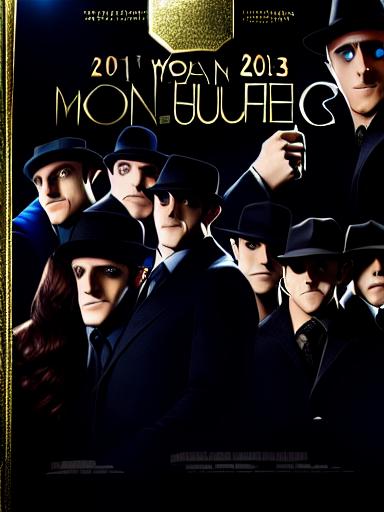}\\\blue{\shortstack{This movie is titled \\ now you see me (2013).}}\end{tabular}\\\\        

        \begin{tabular}{c}\includegraphics[width=0.5\textwidth]{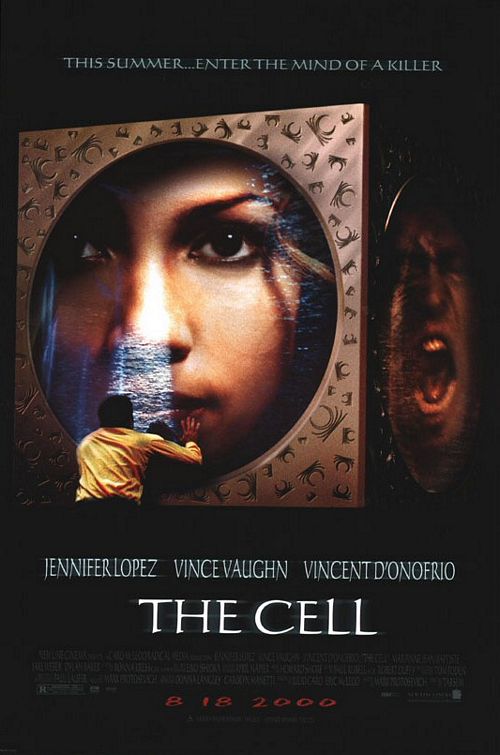}\\\textbf{input}\end{tabular}
        &\begin{tabular}{c}\includegraphics[width=0.5\textwidth]{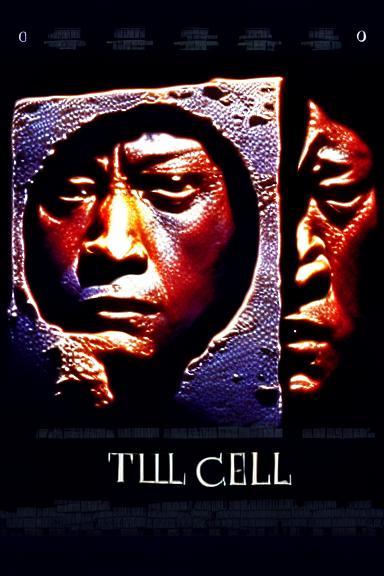}\\\red{\shortstack{This movie is titled \\ the cell (2000).}}\end{tabular}
        &\begin{tabular}{c}\includegraphics[width=0.5\textwidth]{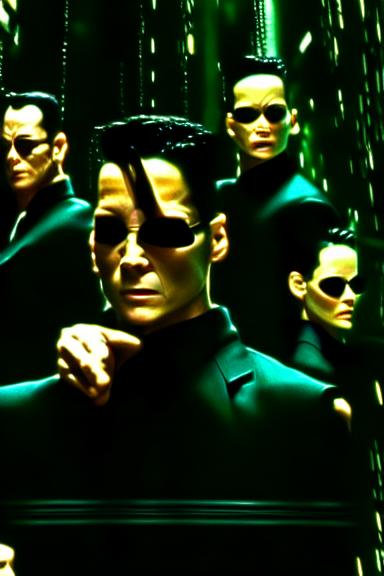}\\\blue{\shortstack{This movie is titled \\ the matrix revolutions (2003).}}\end{tabular}\\\\                
        
        \end{tabular}
    }
    \caption{Examples of image editing on fine-grained perception tasks (posters) in MME benchmark.
    }
    \label{fig:vis_edit_mme_f2}
\end{figure*}

We provide examples of obtaining editing textual instructions from the questions in Table~\ref{table:edit_instruction_example}. We automatically capture the objects or revise the questions to the affirmative sentences as the editing textual instructions.
Figure~\ref{fig:vis_edit_pope}, \ref{fig:vis_edit_mme_c}, \ref{fig:vis_edit_mme_f}, and \ref{fig:vis_edit_mme_f2} are the visual examples of image editing on the POPE and MME benchmarks. The weight of text guidance $cfg_{text}$ is 20 by default. We find that, when editing images with non-existent objects, the resultant images can either successfully add the queried objects without dramatic distortion or just produce the objects without consistency with the original images.
However, when editing images with existent objects, the resultant images may have minor changes in comparison with the original images. This evidence suggests that the image editing method can somewhat reflect the hallucination contents.
In addition, when the tasks are more complicated, where the queried contents are the specific landmarks or posters of movies, the image editing still can edit the queried contents into the images if editing models know the contents.

\begin{table*}[t]
\resizebox{\linewidth}!{
    \begin{tabular}{c|p{3cm}|p{5cm}|p{5cm}}
    Benchmarks & Tasks & Questions & Editing Textual Instructions \\ \hline
    POPE & random, popular, adversarial & Is there a car in the image? & a car \\ \hline
    \multirow{15}{*}{MME} & existence & Is there a elephant in this image? & an elephant \\ \cline{2-4} 
     & count & Are there a total of two trains in the picture? & a total of two trains \\ \cline{2-4} 
     & color & Is there a blue court in the image? & a blue court \\ \cline{2-4} 
     & position & Is the vase on the left of the toothbrush? & the vase on the left of the toothbrush \\ \cline{2-4} 
     & posters & Is this movie titled a beautiful mind (2001)? & This movie is titled a beautiful mind (2001)\\ \cline{2-4} 
     & celebrity & Is the person inside the red bounding box called Sally Field? & the person inside the red bounding box is called Sally Field \\ \cline{2-4} 
     & scene & Does this image describe a place of windmill? & a windmill\\ \cline{2-4} 
     & landmark & Is this a photo of Clearwater Beach, Florida? & Clearwater Beach, Florida  \\ \cline{2-4} 
     & artwork & Does this artwork exist in the form of painting? & This artwork exists in the form of painting \\ \cline{2-4} 
     & OCR & Is the word in the logo "cold drinks"? & the word in the logo "cold drinks" \\ \hline 
\end{tabular}
}
    \caption{ The examples of editing textual instruction obtained from the question. For the cognition tasks in MME, we simply duplicate the question for the instruction.
    }
    \label{table:edit_instruction_example}
\end{table*}

\end{document}